\documentclass[lettersize,journal]{IEEEtran}
\usepackage{amsmath,amsfonts}
\usepackage{algorithmic}
\usepackage{algorithm}
\usepackage{array}

\usepackage{textcomp}
\usepackage{stfloats}
\usepackage{url}
\usepackage{verbatim}
\usepackage{graphicx}
\usepackage{xcolor}
\usepackage{cite}
\usepackage{multirow}
\usepackage{graphicx,subfigure}
\usepackage{caption}
\usepackage{subcaption}
\begin{document}

\title{CCC++: Optimized Color Classified Colorization with Segment Anything Model (SAM) Empowered Object Selective Color Harmonization}

\author{\IEEEauthorblockN{ Mrityunjoy Gain, Avi Deb Raha, and Rameswar Debnath*, \textit{Member, IEEE}}\\
	\IEEEauthorblockA{\textit{Computer Science and Engineering Discipline, Khulna University, Khulna 9208, Bangladesh}\\ 
		E-mail: \{gain1624, dev1611, rdebnath\}@cseku.ac.bd}

{
	\vspace{3mm}
\includegraphics[width=.093\textwidth,height = 1.8cm]{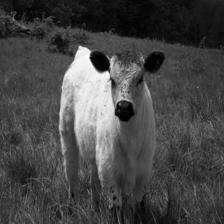}
\includegraphics[width=.093\textwidth,height = 1.8cm]{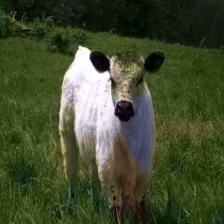}
\includegraphics[width=.093\textwidth,height = 1.8cm]{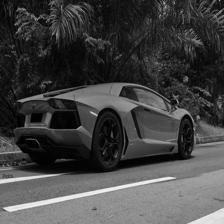}
\includegraphics[width=.093\textwidth,height = 1.8cm]{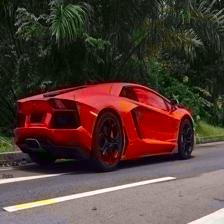}
\includegraphics[width=.093\textwidth,height = 1.8cm]{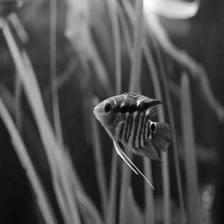}
\includegraphics[width=.093\textwidth,height = 1.8cm]{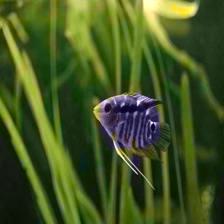}
\includegraphics[width=.093\textwidth,height = 1.8cm]{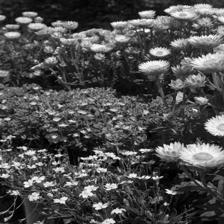}
\includegraphics[width=.093\textwidth,height = 1.8cm]{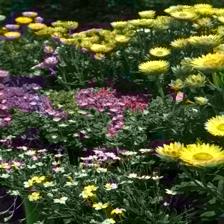}
\includegraphics[width=.093\textwidth,height = 1.8cm]{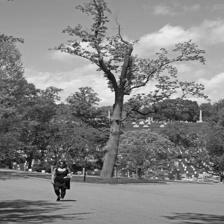}
\includegraphics[width=.093\textwidth,height = 1.8cm]{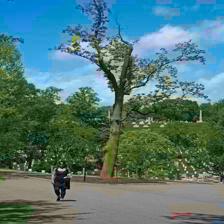}\\
{\small Unveiling the Spectrum: From Monochrome to a Splash of Hues. On the top, grayscale images set the stage, while our colorization marvel, CCC++, transforms them into a vivid symphony of colors on the down.}
\vspace{-10mm}
}
}

\markboth{Journal of \LaTeX\ Class Files,~Vol.~00, No.~00, August~0000}%
{Shell \MakeLowercase{\textit{et al.}}: A Sample Article Using IEEEtran.cls for IEEE Journals}

\maketitle

\begin{abstract}
Automated chromatic transformation of grayscale images involving objects of different colors and sizes is challenging. Both inter and intra-object color variations are often seen where the main objects hold a tiny area in proportion to the extensive background. As a result, a biased model is developed in favor of dominant features, causing the colors of primary objects to merge with the background colors in the resulting image. To tackle this issue, we re-formulate the colorization task as a heterogeneous classification problem and subsequently employ a weighted function to the classes. In this paper, we propose a series of formulas that can convert color values into color classes and vice versa. After conducting experiments using various bin sizes on real-time images, we have determined that our classification task requires 532 distinct color classes. We propose a class-weighted function in each batch where we modify the weights of the predominant classes by decreasing them while increasing the weights of minor groups to reduce the influence of major classes on minor ones. We also introduce a hyper-parameter for ideal balancing between the major and minor classes to prevent the supremacy of minor classes over the major classes while prioritizing the minor classes.  We propose a unique color harmonization technique to improve and enhance the edges of objects using segment anything model (SAM). We propose two novel color image assessment metrics, namely the color class activation ratio (CCAR) and the true activation ratio (TAR), to precisely measure the degree of color component richness. We evaluate our proposed model against eight baselines and SOTA methods using six distinct datasets employing both qualitative and quantitative methodologies. The experimental results demonstrate that our proposed model surpasses previous models in terms of visualization, CNR, CCAR and TAR while maintaining excellent performance in regression, similarity, and generative criteria.

\end{abstract}

\vspace{-2mm}
\section{Introduction}
\IEEEPARstart{H}{uman} vision can perceive thousands of colors. Color is a powerful descriptor that often facilitates object identification from an image. Visual analysis of complex multi-spectral images is easier in color images than in grayscale images. People are still captivated by coloring, whether as a way to revive distant memories or as a means of expressing creative ingenuity. Bringing black and white material into color has a distinctive and profoundly gratifying effect. Moreover, images from antiquity, medicine, industry and astronomy are dull and unable to convey their meanings and expressions accurately. To better comprehend the image's meaning, it is essential to colorize it. Color-coded subject continues to pique the interest of the general public, as seen by the remastered versions of vintage black and white movies, the enduring appeal of colored books for all ages, and the unexpected excitement for a variety of online automatic colorization bots. 

Colorization is the process of assigning color components to intensities of a grayscale image. This process is non-linear and ill-posed. A single gray image can be of many colors. For example, the color of a fruit can be light green, yellow and red. A non-living object can also hold different colors. For example, a vehicle company lunch different colors of a fixed model. During prediction, if the model predicts the red color of a car but the ground truth color of it is black then we cannot say that the prediction is wrong because the red version is also available. The natural colorization means predicting credible color distribution than the ground truth distribution. For this reason, the colorization process is not just predicting the color values of intensities of a gray image like its ground truth image color values.

\begin{figure}[!h]
	\centering
	\begin{subfigure}
		\centering
		{\rotatebox{90}{\small Gray}} 
		\includegraphics[width=.18\linewidth,height = 1.6cm]{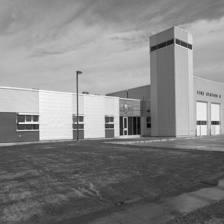}
		\includegraphics[width=.18\linewidth,height = 1.6cm]{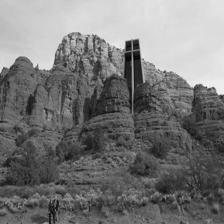}
		\includegraphics[width=.18\linewidth,height = 1.6cm]{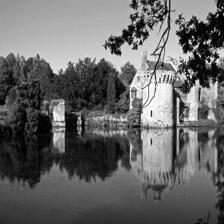}
		\includegraphics[width=.18\linewidth,height = 1.6cm]{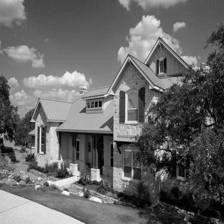}
		\includegraphics[width=.18\linewidth,height = 1.6cm]{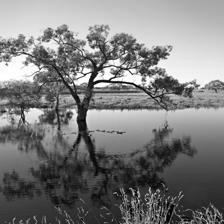}
	\end{subfigure}%
	\\
	\begin{subfigure}
		\centering
		{\rotatebox{90}{\small Colored}} 
		\includegraphics[width=.18\linewidth,height = 1.6cm]{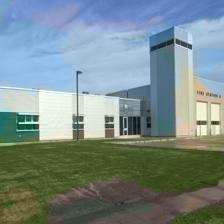}
		\includegraphics[width=.18\linewidth,height = 1.6cm]{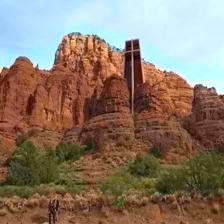}
		\includegraphics[width=.18\linewidth,height = 1.6cm]{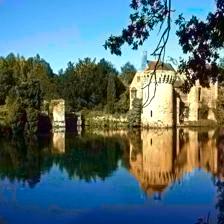}
		\includegraphics[width=.18\linewidth,height = 1.6cm]{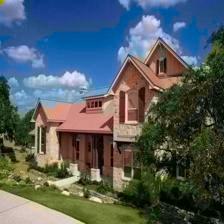}
		\includegraphics[width=.18\linewidth,height = 1.6cm]{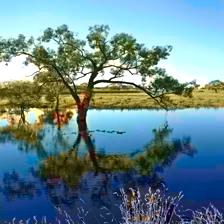}
	\end{subfigure}
	\\
	\begin{subfigure}
		\centering
		{\rotatebox{90}{\small Original}} 
		\includegraphics[width=.18\linewidth,height = 1.6cm]{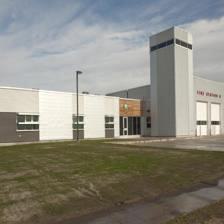}
		\includegraphics[width=.18\linewidth,height = 1.6cm]{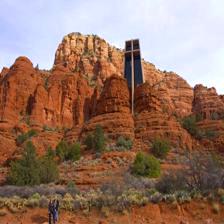}
		\includegraphics[width=.18\linewidth,height = 1.6cm]{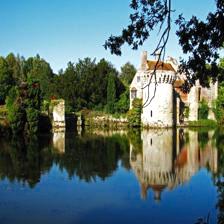}
		\includegraphics[width=.18\linewidth,height = 1.6cm]{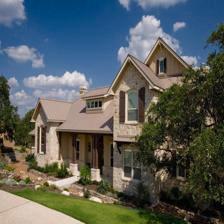}
		\includegraphics[width=.18\linewidth,height = 1.6cm]{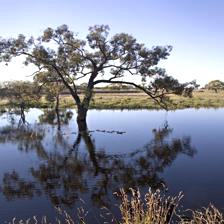}
	\end{subfigure}
	\caption{A visual triad illustrating the transformative power of our model and its fidelity compared to ground truth references. Grayscale inputs on the left, CCC++ colorized outputs in the center, and ground truth images on the right.}
	\label{fig:fig1}
	\vspace{-4mm}
\end{figure}  
Researchers employed a variety of methods for image coloring. User-guided colorization\cite{ Huang, Levin, Yatziv, Qu, Luan, Welsh, Ironi, Tai, Chia, Liu, Sousa, He, Zhang_tog,Charpiat,Gupta,Bugeau} and learning-based colorization\cite{Wu1, Wu2, Wu3, Guo, Bahng, Liang, Zang, Larsson, Gain1, Gain2, An, Iizuka, Su, Xu, Dai, Dahl,Baldassarre,Zhang_eccv,Xia,Hesham,Ozbulak,Kong,Treneska}  are the two primary groups into which existing techniques for coloring grayscale images can be categorized. Too much human interaction is required for traditional user-guided colorization to effectively colorize an image. Scribble-based colorization and example-based colorization are two sub-parts of user-guided colorization. The user-guided approach has lost favor in respect of learning-based strategies, which are easier to implement and require less human labor. Learning-based strategies for image colorization, including classical regression \cite{Gain1,Iizuka,Dahl,Baldassarre,Nguyen-Quynh}, object segmentation \cite{Su,Xu,Xia,Hesham,Kong}, generative approaches \cite{Wu1,Wu2,Wu3,Guo,Bahng,Liang,Zang,Ozbulak,Treneska}, feature-balancing \cite{Larsson,Gain2,An,Zhang_eccv}, and vision transformer and diffusion-based techniques \cite{icolorit, DD, codebook, Lee, lcad, diffusart, colorcontrol}, are gaining popularity. These deep learning methods, particularly regression-based approaches, are favored for their ease of implementation and reduced need for human intervention, enhancing the efficiency and effectiveness of the colorization process.

Deep Neural Networks (DNNs) derive representative features and hidden structural knowledge from data through multiple underlying network layers via training \cite{Gain_facemask, Raha_access, Raha_advancing, Gain_iceeict}. Every epoch in training phase, the loss function generates feedback to refine the model's parameters. The loss function quantifies the difference between the predicted output and the actual output, also known as the error. In every epoch, networks adjust the model's weights proportionally to the error. The backpropagation algorithm assigns equal weight to misclassification errors of data instances belonging to each class. For imbalanced class distributions, the training procedure modifies the classifier in favor of the majority class\cite{DWBL_12}. In an unbalanced distribution, difficult instances from classes with fewer observations result in class probabilities that are lower, as predicted by the models. Incorrect and out-of-distribution instances have lower softmax probabilities than correct instances\cite{Hendrycks_13}. Since the training process with an imbalanced class distribution typically underestimates the class probability of minority class instances, learning models incorrectly classify minority class observations as hard instances\cite{Wallace_14}. Thus, the training process with an imbalanced class distribution does not hinder the model's performance on tasks requiring unambiguous class separation, however it does impact the performance of instances that are inherently more challenging to classify. Therefore, predicted class probabilities are subsequently unreliable in imbalanced class distributions.

In the colorization problem, feature distributions are more important than class distribution. When feature distribution exhibit imbalance, similar imbalanced learning scenarios are encountered during the training process. We observed that desaturated color components are far more prevalent than saturated color components in training images. The training process is therefore biased towards the desaturated color components. It negatively affects the performance of saturated color components, causing the targeted image's saturated color components to be biased toward desaturated color components. The training process is therefore biased towards the larger feature subsets. Therefore, the characteristics of the smaller feature subsets are eliminated by the predicted models. Occasionally, the hues of smaller objects in the image of interest merge with the background. Handling feature imbalance is of utmost significance because the smaller subsets of features are the features of interest for the learning task. 

Typically, the class imbalance is addressed by resampling the dataset to achieve class parity or rescaling the data samples or employing a weighted function to impose a higher weight on the minority class \cite{DWBL_12}. In the training process, a sample's features determine the gradient directions on the loss function. An image is comprised of pieces of varying sizes and hues. However, each unit has the same significance. The input dimension of a learning model is determined by the sample's spatial resolution. The output dimension of colorization models is the same as the input dimension. Still there is no way to resize or reproduce a feature. Defining a set of rules that can transform feature values into class values is a solution to feature imbalance problems. 

\begin{figure}[!h]
	\centering
	\begin{subfigure}
		\centering
		{\rotatebox{90}{\small Gray}} 
		\includegraphics[width=.148\linewidth,height = 1.5cm]{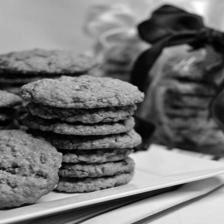}
		\includegraphics[width=.148\linewidth,height = 1.5cm]{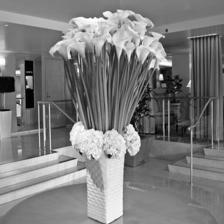}
		\includegraphics[width=.148\linewidth,height = 1.5cm]{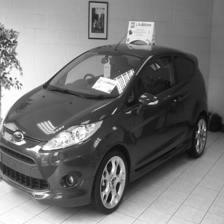}
		\includegraphics[width=.148\linewidth,height = 1.5cm]{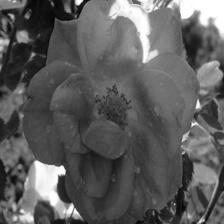}
		\includegraphics[width=.148\linewidth,height = 1.5cm]{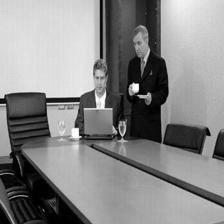}
		\includegraphics[width=.148\linewidth,height = 1.5cm]{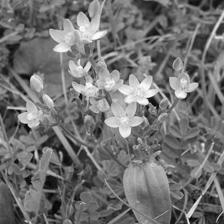}
	\end{subfigure}%
	\hfill
	\begin{subfigure}
		\centering
		{\rotatebox{90}{\small Regression}} 
		\includegraphics[width=.148\linewidth,height = 1.5cm]{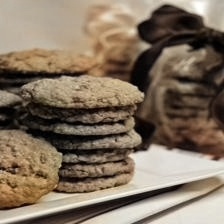}
		\includegraphics[width=.148\linewidth,height = 1.5cm]{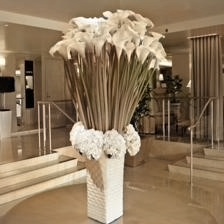}
		\includegraphics[width=.148\linewidth,height = 1.5cm]{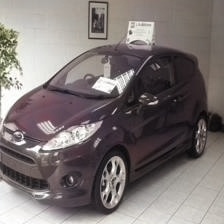}
		\includegraphics[width=.148\linewidth,height = 1.5cm]{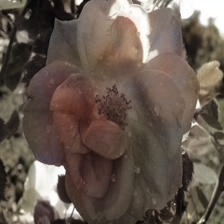}
		\includegraphics[width=.148\linewidth,height = 1.5cm]{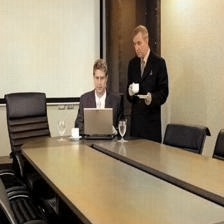}
		\includegraphics[width=.148\linewidth,height = 1.5cm]{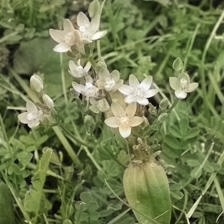}
	\end{subfigure}
	\hfill
	\begin{subfigure}
		\centering
		{\rotatebox{90}{\scriptsize Classification}} 
		\includegraphics[width=.148\linewidth,height = 1.5cm]{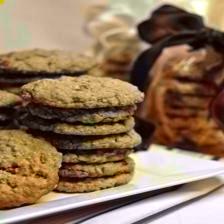}
		\includegraphics[width=.148\linewidth,height = 1.5cm]{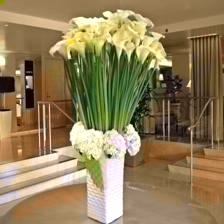}
		\includegraphics[width=.148\linewidth,height = 1.5cm]{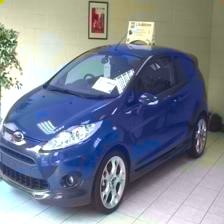}
		\includegraphics[width=.148\linewidth,height = 1.5cm]{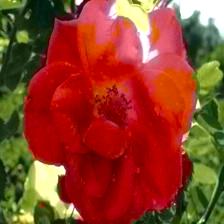}
		\includegraphics[width=.148\linewidth,height = 1.5cm]{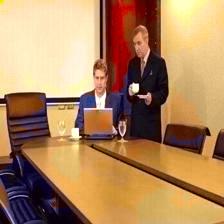}
		\includegraphics[width=.148\linewidth,height = 1.5cm]{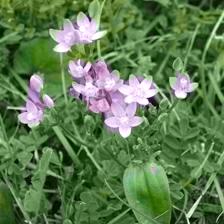}
	\end{subfigure}
	\hfill
	\begin{subfigure}
		\centering
		{\rotatebox{90}{\small Original}} 
		\includegraphics[width=.148\linewidth,height = 1.5cm]{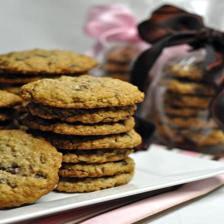}
		\includegraphics[width=.148\linewidth,height = 1.5cm]{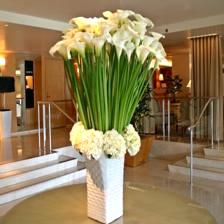}
		\includegraphics[width=.148\linewidth,height = 1.5cm]{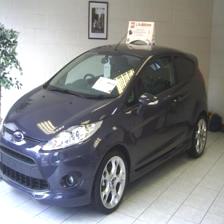}
		\includegraphics[width=.148\linewidth,height = 1.5cm]{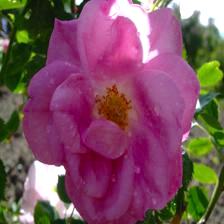}
		\includegraphics[width=.148\linewidth,height = 1.5cm]{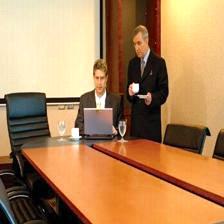}
		\includegraphics[width=.148\linewidth,height = 1.5cm]{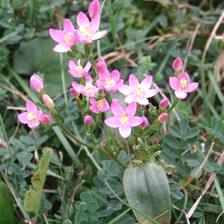}
	\end{subfigure}
	\caption{Results with regression and classification}
	\label{fig:reg_cla}
	\vspace{-2mm}
\end{figure}

\begin{figure}[!h]
	\centering
	\begin{subfigure}
		\centering
		{\rotatebox{90}{\small Gray}} 
		\includegraphics[width=.148\linewidth,height = 1.5cm]{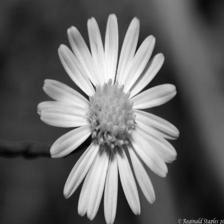}
		\includegraphics[width=.148\linewidth,height = 1.5cm]{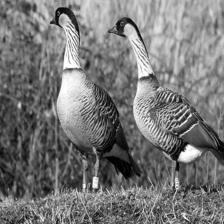}
		\includegraphics[width=.148\linewidth,height = 1.5cm]{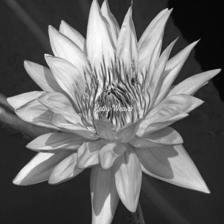}
		\includegraphics[width=.148\linewidth,height = 1.5cm]{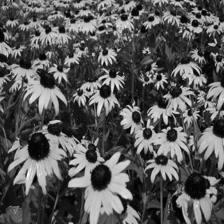}
		\includegraphics[width=.148\linewidth,height = 1.5cm]{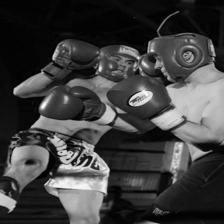}
		\includegraphics[width=.148\linewidth,height = 1.5cm]{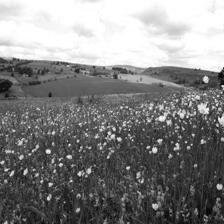}
	\end{subfigure}%
	\hfill
	\begin{subfigure}
		\centering
		{\rotatebox{90}{\small Full Class}} 
		\includegraphics[width=.148\linewidth,height = 1.5cm]{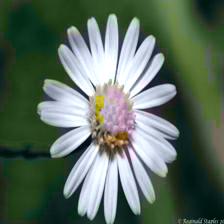}
		\includegraphics[width=.148\linewidth,height = 1.5cm]{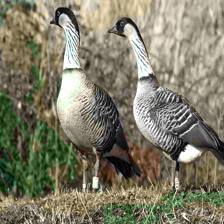}
		\includegraphics[width=.148\linewidth,height = 1.5cm]{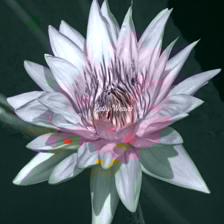}
		\includegraphics[width=.148\linewidth,height = 1.5cm]{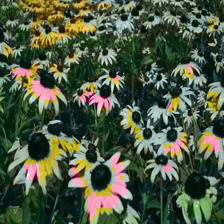}
		\includegraphics[width=.148\linewidth,height = 1.5cm]{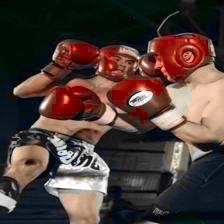}
		\includegraphics[width=.148\linewidth,height = 1.5cm]{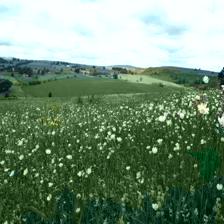}
	\end{subfigure}
	\hfill
	\begin{subfigure}
		\centering
		{\rotatebox{90}{\small Opt. Class}} 
		\includegraphics[width=.148\linewidth,height = 1.5cm]{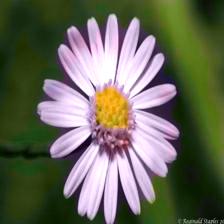}
		\includegraphics[width=.148\linewidth,height = 1.5cm]{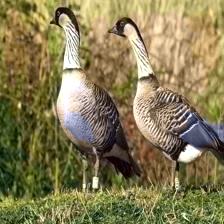}
		\includegraphics[width=.148\linewidth,height = 1.5cm]{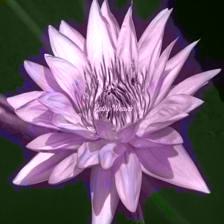}
		\includegraphics[width=.148\linewidth,height = 1.5cm]{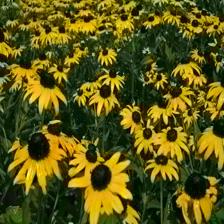}
		\includegraphics[width=.148\linewidth,height = 1.5cm]{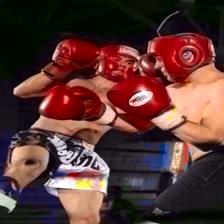}
		\includegraphics[width=.148\linewidth,height = 1.5cm]{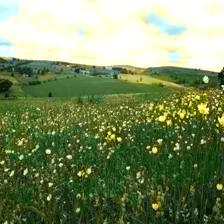}
	\end{subfigure}
	\hfill
	\begin{subfigure}
		\centering
		{\rotatebox{90}{\small Original}} 
		\includegraphics[width=.148\linewidth,height = 1.5cm]{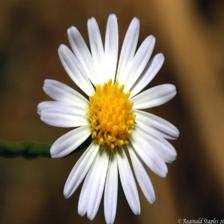}
		\includegraphics[width=.148\linewidth,height = 1.5cm]{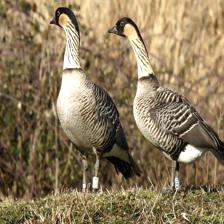}
		\includegraphics[width=.148\linewidth,height = 1.5cm]{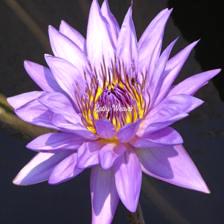}
		\includegraphics[width=.148\linewidth,height = 1.5cm]{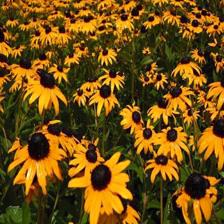}
		\includegraphics[width=.148\linewidth,height = 1.5cm]{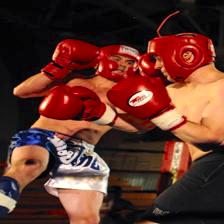}
		\includegraphics[width=.148\linewidth,height = 1.5cm]{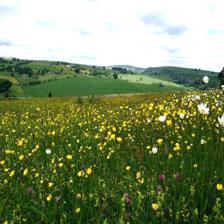}
	\end{subfigure}
	\caption{Results with Complete and Optimized bin}
	\label{fig:com_opti}
	\vspace{-2mm}
\end{figure}

As a solution to the problems, we take motivation from the work CIC\cite{Zhang_eccv}. we define a set of rules to transform continuous color values into discrete color classes and vice versa to predict a distribution of possible colors for each pixel instead of the average pixel. We first transform double-channel a*b* color components to a single channel 1296 color classes, where a* channel and b* channel value range are physiclly in $[-108, 108)$. The basic works which are improvement of CCC++ over the work CIC\cite{Zhang_eccv} are illustrated in TABLE \ref{table1}. In the early version of our work\cite{ccc} we formed 400 classes taking bin size value as 10 and then optimized to 215 classes based on real time appearance. In this work, we have optimized our previous work. we experiment on different bin size and find 6 as optimal for our colorization task. All color classes are not found in real-time color images. We analyze the \textit{Place365 Validation} dataset\cite{Place} and found that 532 out of 1296 color classes are mostly exist. Because redundant classes reduces classification accuracy. We apply weighted cross-entropy loss instead of classical cross-entropy loss. In this paper, we propose a technique for feature balancing. Class weights are determined by analyzing the true class of each batch during training. Rare-appeared classes need high weights than the most appeared classes to remove desaturation and biases toward major features. We re-weighted the classes making a suitable trade-off between major classes and minor classes to remove both desaturation and over-saturation. Moreover, we introduce an innovative color harmonization approach empowered by SAM (Segment Anything Model)\cite{SAM}. This harmonization method is specifically designed for object-selective refinement and enhancement of color representation in images. We propose two new color image asses metrics along with previous metric of our earliest version. 

\begin{table}[ht]
	\centering
	\caption{Comparison of Features between CIC \cite{Zhang_eccv} and CCC++}
	\begin{tabular}{p{2.3cm}p{1.2cm}p{4.5cm}}
		\hline
		\multicolumn{1}{c}{\textbf{Feature}} & \multicolumn{1}{c}{\textbf{CIC \cite{Zhang_eccv}}} & \multicolumn{1}{c}{\textbf{CCC++}} \\
		\hline
		Bidirectional color class conversion formula & Not implemented & Implements conversion formulas between color classes and vice versa, enhancing flexibility in color manipulation \\
		
		Experiment-based optimal bin size determination & Not performed & Uses experimental methods to determine the optimal bin size, ensuring accurate and efficient color class representation \\
		
		Data-driven class points optimization & Not utilized & Employs data-driven techniques to optimize class points, improving the accuracy and relevance of color classifications \\
		
		Task generalization for feature imbalance & Lacks adaptability & Generalizes to similar feature imbalance problems, making the model adaptable to a wide range of scenarios and datasets \\
		
		Data-driven hyperparameter based class weight formulation & Absent & Formulates class weights function using real time data-driven hyperparameter, resulting in more balanced and effective colorization outcomes \\
		
		Learning based hyperparameter tuning for optimal performance & Not available & Conducts hyperparameter tuning during model train to find the perfect trade-off between various performance metrics, enhancing overall model efficacy \\
		
		Segmentation-based edge refinement & Not included & Refines object edges using segmentation techniques, resulting in cleaner and more precise color boundaries \\
		
		Chromatic number ratio as an evaluation metric & Not used & Incorporates the chromatic number ratio to evaluate color diversity and distribution in the colorized images \\
		
		Color class activation ratio as an evaluation metric & Not employed & Uses the color class activation ratio to assess the accuracy and activation level of different color classes \\
		
		True activation ratio as an evaluation metric & Not measured & Utilizes the true activation ratio to evaluate the model’s performance in activating the correct color classes based on the input \\
		\hline
	\end{tabular}
	\label{table1}
	\vspace{-2mm}
\end{table}

 The following are the contributions to this work:
\begin{enumerate}
 
  \item We propose a series of equations to convert continuous color values from double-channel color spaces to discrete color classes in a single-channel color space and vice versa. These baseline formulas can be used to convert any regression problem with feature imbalance into a classification task.
  
  \item We experiment on different bin size for color class transformation. Observing class appearance, standard deviation, and model parameter on a variety of extremely large-scale real-time images in practice, we propose bin size 6 for our colorization task.

  \item We optimize class levels analyzing different numerous images and propose 532 color classes for the colorization task.
 
  \item We propose a class re-weighting formula to prioritize high gradient values from misclassified low-frequency or rare classes. This ensures that all classes make a balanced contribution to the loss function. This process eliminates any biases related to majority features and the issues of desaturation and over-saturation in the color distribution, thereby ensuring accurate and standard predictions.
  
  \item In our class re-weight formula we propose a hyper-parameter for finding the optimal trade-off between the major and minor appeared classes. This hyper-parameter testing prevent minor classes domination over the major classes, while prioritizing the minor classes.

  \item We propose an innovative approach for enhancing the color harmony of selected objects using the SAM, which results in improved edge definition and refinement.
   
  \item We propose two novel color image evaluation metrics: the Color Class Activation Ratio (CCAR) and the True Activation Ratio (TAR). These metrics assess the abundance of color classes in generated images relative to ground truth images, offering a full assessment of the color spectrum.
 
  \item We provide a plethora of quantitative and qualitative findings that clearly show that our approach surpasses existing SOTA benchmarks and yields satisfactory outcomes.

\end{enumerate}

The rest of the paper is structured as follows. Section \ref{Literature Review} is a review of the relevant literature regarding image colorization. In Section \ref{Methodology}, the entire methodology, including problem formulation and solution approach, is presented. In Section \ref{Experiments}, we present the experimental outcomes and a comparative analysis with other cutting-edge techniques. In Section \ref{Conclusion}, the conclusion is presented.

\begin{figure}[!h]
    \centering
    \begin{subfigure}
        \centering
        {\rotatebox{90}{\small Gray}} 
        \includegraphics[width=.148\linewidth,height = 1.5cm]{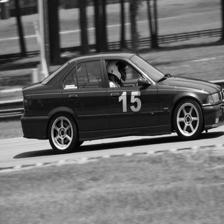}
        \includegraphics[width=.148\linewidth,height = 1.5cm]{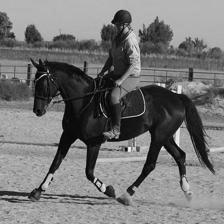}
        \includegraphics[width=.148\linewidth,height = 1.5cm]{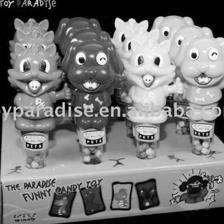}
        \includegraphics[width=.148\linewidth,height = 1.5cm]{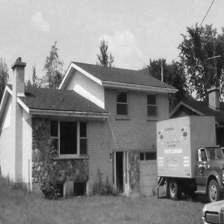}
        \includegraphics[width=.148\linewidth,height = 1.5cm]{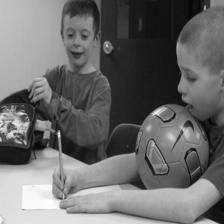}
        \includegraphics[width=.148\linewidth,height = 1.5cm]{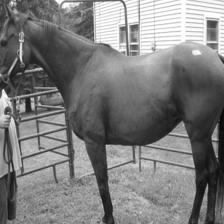}
    \end{subfigure}%
    \hfill
    \begin{subfigure}
        \centering
        {\rotatebox{90}{\small Nobalance}} 
        \includegraphics[width=.148\linewidth,height = 1.5cm]{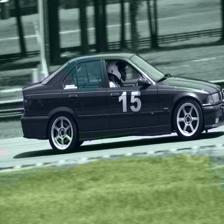}
        \includegraphics[width=.148\linewidth,height = 1.5cm]{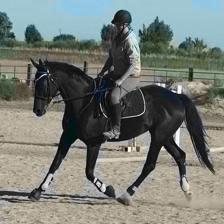}
        \includegraphics[width=.148\linewidth,height = 1.5cm]{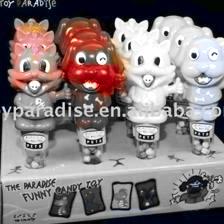}
        \includegraphics[width=.148\linewidth,height = 1.5cm]{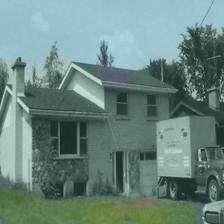}
        \includegraphics[width=.148\linewidth,height = 1.5cm]{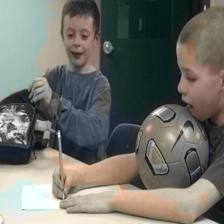}
        \includegraphics[width=.148\linewidth,height = 1.5cm]{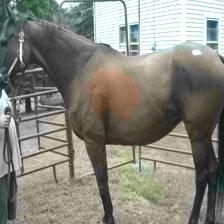}
    \end{subfigure}
    \hfill
    \begin{subfigure}
        \centering
        {\rotatebox{90}{\small Rebalance}} 
        \includegraphics[width=.148\linewidth,height = 1.5cm]{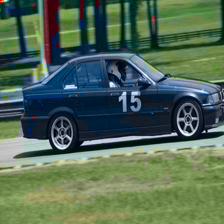}
        \includegraphics[width=.148\linewidth,height = 1.5cm]{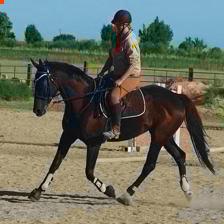}
        \includegraphics[width=.148\linewidth,height = 1.5cm]{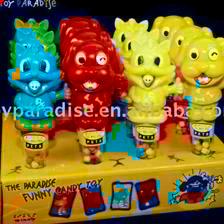}
        \includegraphics[width=.148\linewidth,height = 1.5cm]{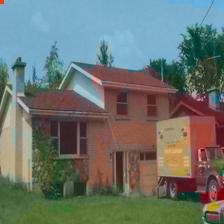}
        \includegraphics[width=.148\linewidth,height = 1.5cm]{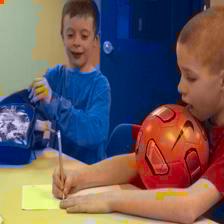}
        \includegraphics[width=.148\linewidth,height = 1.5cm]{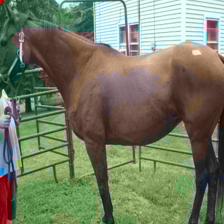}
    \end{subfigure}
    \hfill
    \begin{subfigure}
        \centering
        {\rotatebox{90}{\small Original}} 
        \includegraphics[width=.148\linewidth,height = 1.5cm]{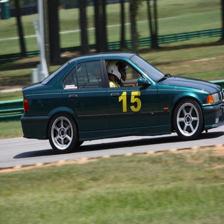}
        \includegraphics[width=.148\linewidth,height = 1.5cm]{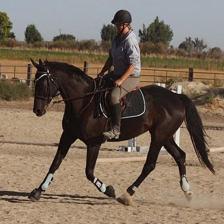}
        \includegraphics[width=.148\linewidth,height = 1.5cm]{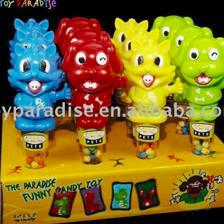}
        \includegraphics[width=.148\linewidth,height = 1.5cm]{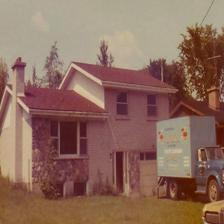}
        \includegraphics[width=.148\linewidth,height = 1.5cm]{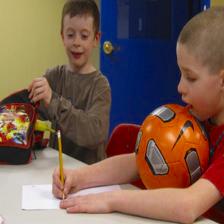}
        \includegraphics[width=.148\linewidth,height = 1.5cm]{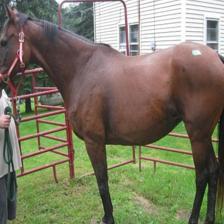}
    \end{subfigure}
    \caption{Results with rebalance and without rebalance}
    \label{fig:reb_no}
    \vspace{-2mm}
\end{figure}
\section{Literature Review} \label{Literature Review}
\subsection{Background}
This section will describe various colorization methodologies and the concept of how colorization methods function. It has a long history of research and is a prominent issue in computer graphics. Early approaches to image colorization relied on hand-crafted rules and heuristics, and it is referred to as user-guided colorization. In general, scribbles and examples-based coloring fall into two distinct user-interacted categories. It is unrealistic to assume that one or more reference images will provide sufficient color information to produce adequate colorization results. These approaches often lacked the ability to generate realistic outcomes. Colorization is now performed using techniques based on deep learning. Due to the fact that data-driven learning-based techniques function without human intervention, a large number of source images can be used to train the models. In terms of colorization, this refers to the automatic identification of colors that correspond well with real-world objects. Increasing the number of training samples and network layers (some cases) contributes to enhanced results. Learning-based colorization can be broadly categorized into five groups: CNN-based regression approach, object segmentation approach, Generative Adversarial Network (GAN) based generative approach, transformer and diffusion based approach and feature balancing approach. This section provides a concise explanation of these methods.
\subsubsection{User Guided Colorization}
\paragraph{Scribble Based Colorization}
The scribble-based method is considered one of the most ancient techniques for colorization. The software utilizes color interpolation technique to accurately fill in missing or incomplete sections of an image, based on the user's input in the form of scribbles.\\
Levin et. al.\cite{Levin} proposed a technique based on optimization for propagating user-specific color scribbles to all pixels within an image. As indicated by the user's color scribbles, adjacent pixels with identical intensity levels were designated the same color. For this endeavor, the quadratic cost function was utilized. Huang et. al.\cite{Huang} proposed a method that combines non-iterative techniques with adaptive edge extraction to reduce the computational time necessary for colorization optimization and mitigate color leakage effects. Yatziv et. al.\cite{Yatziv} introduced the concept of color blending, in which the chromatic value of a pixel is determined by the contribution of specified colors. Qu et. al.\cite{Qu} proposed a technique for propagating color effectively in both pattern-continuous and intensity-continuous regions. To improve the visual impact of sparsely distributed user strokes, Luan et. al.\cite{Luan} suggested incorporating nearby pixels with similar intensity and distant pixels with similar characteristics. In pursuance of the same objective, Xu et. al.\cite{Xu} utilized a probability distribution to determine the most trustworthy stroke color for each pixel. Zhang et. al.\cite{Zhang_tog} incorporated the U-Net structure into their colorization framework to enhance the visual effect of colorization while minimizing user interactions. 
\paragraph{Example Based Colorization}
These methods provide a more user-friendly approach to minimizing user effort by providing a reference image that is closely related to the input grayscale. Similar to the approach demonstrated by Reinhard et. al.\cite{Reinhard}, the initial study conducted by Welsh et. al.\cite{Welsh} involved the transmission of colors using global color statistics. Due to its disregard for spatial pixel data, the methods frequently produced unsatisfactory results in a number of instances. Various correspondence techniques have been investigated in order to obtain a more precise local transfer. These techniques include segmented region-level approaches proposed by Irony et. al. \cite{Ironi}, Tai et. al.\cite{Tai}, and Charpiat et. al.\cite{Charpiat}, super-pixel level methods proposed by Gupta et. al.\cite{Gupta} and Chia et. al.\cite{Chia}, and pixel-level methods proposed by Liu et. al.\cite{Liu} and Bugeau et. al. \cite{Bugeau}. However, the process of identifying low-level feature correspondences using manually devised similarity metrics, such as SIFT and Gabor wavelet, can be prone to error in scenarios with substantial variation in intensity and content. According to Sousa et. al.\cite{Sousa}, the color ascribed to each pixel in a grayscale image is determined by its intensity in a similarly themed reference color image. The research carried out by He et al.\cite{He} has used deep features derived from a pre-trained VGG-19 network to achieve precise matching between visually dissimilar pictures that have semantic similarity. They used the approach of style transfer and color transfer, respectively.
 \begin{figure}[!h]
    \centering
    \begin{subfigure}
        \centering
        {\rotatebox{90}{\small Gray}} 
        \includegraphics[width=.148\linewidth,height = 1.5cm]{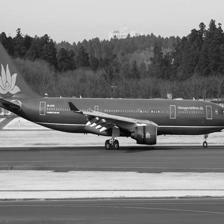}
        \includegraphics[width=.148\linewidth,height = 1.5cm]{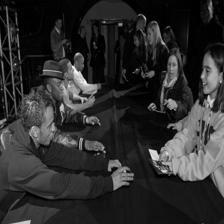}
        \includegraphics[width=.148\linewidth,height = 1.5cm]{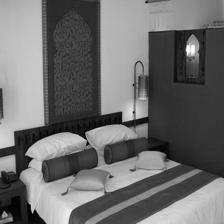}
        \includegraphics[width=.148\linewidth,height = 1.5cm]{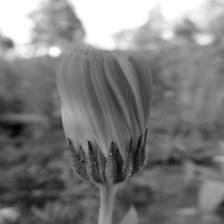}
        \includegraphics[width=.148\linewidth,height = 1.5cm]{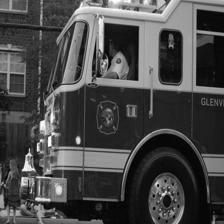}
        \includegraphics[width=.148\linewidth,height = 1.5cm]{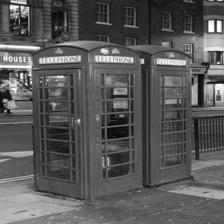}
    \end{subfigure}%
    \hfill
    \begin{subfigure}
        \centering
        {\rotatebox{90}{\small Our}} 
        \includegraphics[width=.148\linewidth,height = 1.5cm]{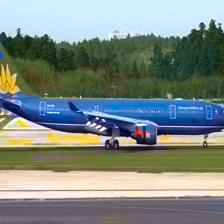}
        \includegraphics[width=.148\linewidth,height = 1.5cm]{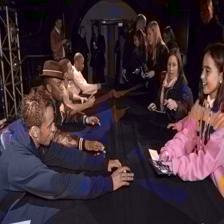}
        \includegraphics[width=.148\linewidth,height = 1.5cm]{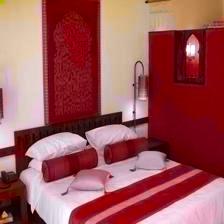}
        \includegraphics[width=.148\linewidth,height = 1.5cm]{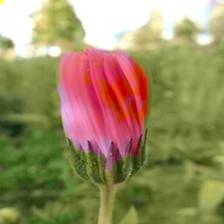}
        \includegraphics[width=.148\linewidth,height = 1.5cm]{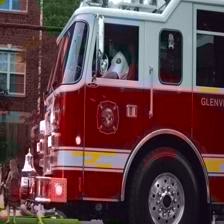}
        \includegraphics[width=.148\linewidth,height = 1.5cm]{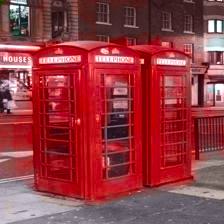}
    \end{subfigure}
    \hfill
    \begin{subfigure}
        \centering
        {\rotatebox{90}{\small SOTA}} 
        \includegraphics[width=.148\linewidth,height = 1.5cm]{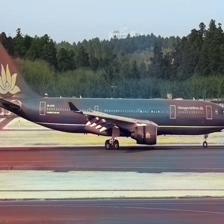}
        \includegraphics[width=.148\linewidth,height = 1.5cm]{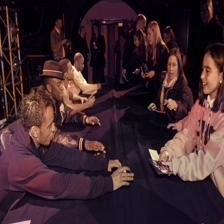}
        \includegraphics[width=.148\linewidth,height = 1.5cm]{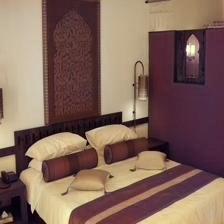}
        \includegraphics[width=.148\linewidth,height = 1.5cm]{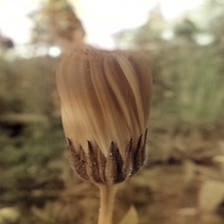}
        \includegraphics[width=.148\linewidth,height = 1.5cm]{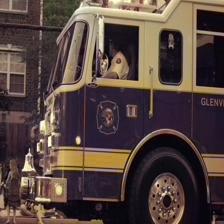}
        \includegraphics[width=.148\linewidth,height = 1.5cm]{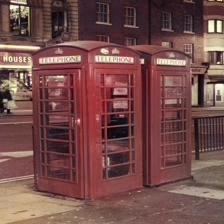}
    \end{subfigure}
    \hfill
    \begin{subfigure}
        \centering
        {\rotatebox{90}{\small Original}} 
        \includegraphics[width=.148\linewidth,height = 1.5cm]{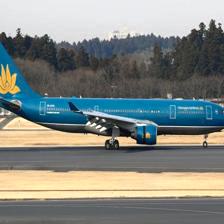}
        \includegraphics[width=.148\linewidth,height = 1.5cm]{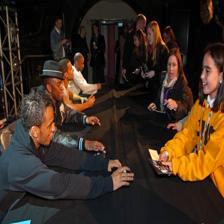}
        \includegraphics[width=.148\linewidth,height = 1.5cm]{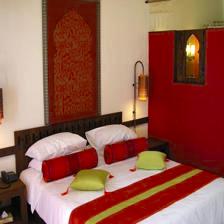}
        \includegraphics[width=.148\linewidth,height = 1.5cm]{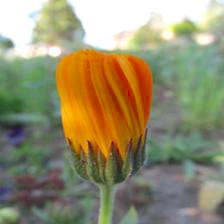}
        \includegraphics[width=.148\linewidth,height = 1.5cm]{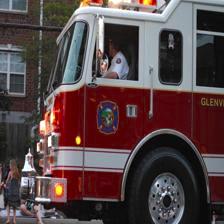}
        \includegraphics[width=.148\linewidth,height = 1.5cm]{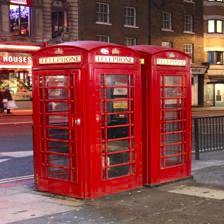}
    \end{subfigure}
    \caption{(A) Results of our proposed method with respect to State of The Art(SOTA) rebalance method}
    \label{fig:com_sota1}
    \vspace{-2mm}
\end{figure}
\subsubsection{Learning Based Colorization}
Learning-based colorization automatically applies color to grayscale images using deep learning, specifically CNNs, which learn from large datasets of colored images. The CNN is trained to convert grayscale values into corresponding colors by optimizing a loss function that measures the difference between predicted and actual colors. This enables the model to apply learned colorization to new images. Methods include basic regression-based, object segmentation-based, GAN-based, and transformer and diffusion-based approaches. The major issue regarding colorization using machine learning techniques is feature balancing for focused objects and background of the images. A few research have been done on this issue for machine learning techniques.

\paragraph{Basic Regression Based Colorization}
For colorization, conventional CNN or specialized architectures such as InceptionNet, VGGNet, ResNet, DenseNet, etc. are utilized. Grayscale images are fed to the network as input and color channels are estimated. The MSE, MAE, RMSE, etc. are then employed to calculate the loss between estimated color channels and true color channels. An optimizer is utilized to minimize loss and achieves ideal color distribution. Using four pre-trained VGG16\cite{VGG} layers, Dahl et al.\cite{Dahl} developed an automated method for generating full-color channels from gray images. Hwang et. al.\cite{Hwang} proposed a colorization method  based on the baseline regression model. Baldassarre et al.\cite{Baldassarre} proposed a technique that incorporates Deep CNN and Inception-ResNet-v2. Iizuka et al.\cite{Iizuka} developed an end-to-end method that simultaneously learns global and local image attributes.
Encoder-decoder based colorization model proposed by Nguyen-Quynh et al.\cite{Nguyen-Quynh} using both global and local priors. Encoder-decoder CNN architecture with filtering-based rebalancing methods was presented by Gain et al.\cite{Gain1}. Yun et al.\cite{icolorit} propose a novel point-interactive colorization Vision Transformer (iColoriT) that utilizes the global receptive field of Transformers to distribute user stimuli to appropriate locations.
\paragraph{Object Segmentation Based Colorization}
Objects within an image are segmented, and the network then takes these segments as input along with the full images. The network can learn color assignment segment-wise or object-wise. After an image has been segmented, the colorization model assigns colors to its segments using a variety of techniques.
Su et al. \cite{Su} developed an instance colorization network that extracts color information from both full images and individual objects. Xu et al. \cite{Xu} proposed using semantic segmentation to automate image colorization, leveraging segmentation to improve image edges. Hesham et al. \cite{Hesham} introduced a colorization model using the scaled-YOLOv4 detector to differentiate objects in multi-object images. Kong et al. \cite{Kong} presented an adversarial edge-aware model combining multitask output and semantic segmentation, using a generator to learn colorization from chromatic ground truth. Xia et al. \cite{Xia} proposed a method for segmenting nuclei with minimal supervision, combining a secondary colorization task with the segmentation task.
 \begin{figure}[!h]
	\ContinuedFloat 
	\centering
	\begin{subfigure}
		\centering
		{\rotatebox{90}{\small Gray}} 
		\includegraphics[width=.148\linewidth,height = 1.5cm]{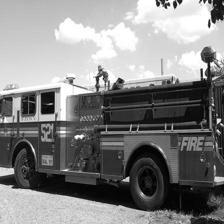}
		\includegraphics[width=.148\linewidth,height = 1.5cm]{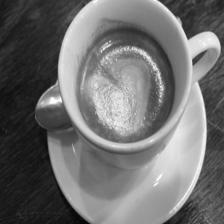}
		\includegraphics[width=.148\linewidth,height = 1.5cm]{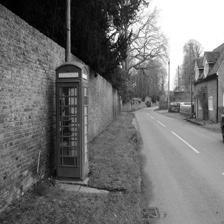}
		\includegraphics[width=.148\linewidth,height = 1.5cm]{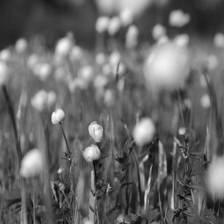}
		\includegraphics[width=.148\linewidth,height = 1.5cm]{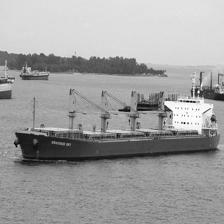}
		\includegraphics[width=.148\linewidth,height = 1.5cm]{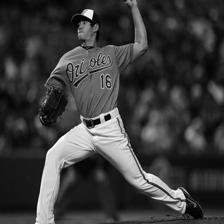}
	\end{subfigure}%
	\hfill
	\begin{subfigure}
		\centering
		{\rotatebox{90}{\small Our}} 
		\includegraphics[width=.148\linewidth,height = 1.5cm]{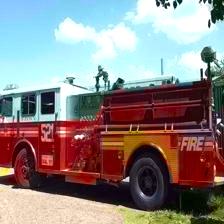}
		\includegraphics[width=.148\linewidth,height = 1.5cm]{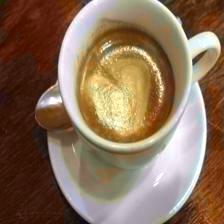}
		\includegraphics[width=.148\linewidth,height = 1.5cm]{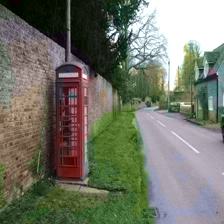}
		\includegraphics[width=.148\linewidth,height = 1.5cm]{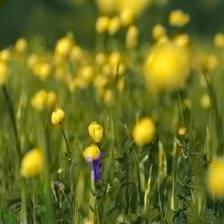}
		\includegraphics[width=.148\linewidth,height = 1.5cm]{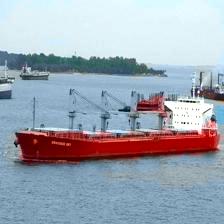}
		\includegraphics[width=.148\linewidth,height = 1.5cm]{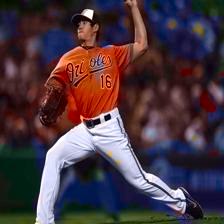}
	\end{subfigure}
	\hfill
	\begin{subfigure}
		\centering
		{\rotatebox{90}{\small SOTA}} 
		\includegraphics[width=.148\linewidth,height = 1.5cm]{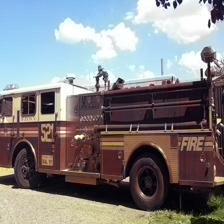}
		\includegraphics[width=.148\linewidth,height = 1.5cm]{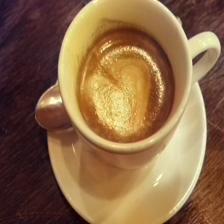}
		\includegraphics[width=.148\linewidth,height = 1.5cm]{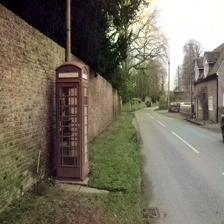}
		\includegraphics[width=.148\linewidth,height = 1.5cm]{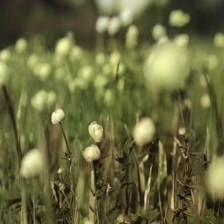}
		\includegraphics[width=.148\linewidth,height = 1.5cm]{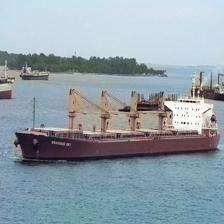}
		\includegraphics[width=.148\linewidth,height = 1.5cm]{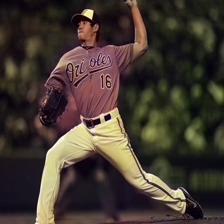}
	\end{subfigure}
	\hfill
	\begin{subfigure}
		\centering
		{\rotatebox{90}{\small Original}} 
		\includegraphics[width=.148\linewidth,height = 1.5cm]{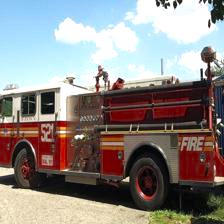}
		\includegraphics[width=.148\linewidth,height = 1.5cm]{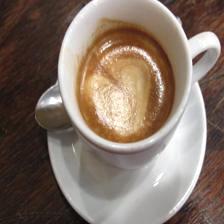}
		\includegraphics[width=.148\linewidth,height = 1.5cm]{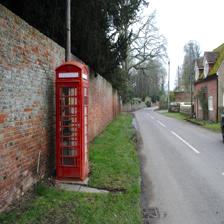}
		\includegraphics[width=.148\linewidth,height = 1.5cm]{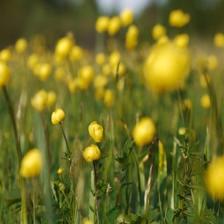}
		\includegraphics[width=.148\linewidth,height = 1.5cm]{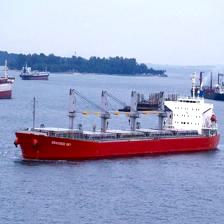}
		\includegraphics[width=.148\linewidth,height = 1.5cm]{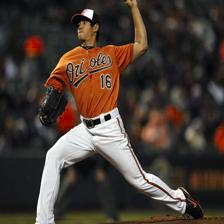}
	\end{subfigure}
	\caption{(B) Results of our proposed method with respect to State of The Art(SOTA) rebalance method}
	\label{fig:com_sota2}
	\vspace{-2mm}
\end{figure}
\paragraph{GAN Based Colorization}
GAN models for image colorization consist of a generator that produces colorized images from grayscale photos and a discriminator that differentiates between real and generated images. This adversarial process helps the model learn patterns and color distributions from the training dataset, resulting in enhanced colorization. Wu et al.\cite{Wu1} introduced a GAN method for image colorization that incorporates detailed semantic information. Guo et al.\cite{Guo} introduced a GAN-based, bilateral Res-U-net model for the purpose of image colorization. Ozbulak et al.\cite{Ozbulak} utilized the Capsule Network (CapsNet) technique to colorize images. Wu et al.\cite{Wu2} introduced a GAN based model. By utilizing a GAN encoder, the researchers detected corresponding characteristics that were similar to examples and subsequently adjusted these characteristics during the colorization process. Bahng et. al.\cite{Bahng} introduced a model comprising of two conditional generative adversarial networks. The initial network transforms text into a palette, while the subsequent network employs the palettes to add color to images. To enhance the quality of synthetic images and safeguard colourized medical images, Liang et al.\cite{Liang} proposed a colorization network that utilizes the cycle generative adversarial network (CycleGAN) model. This network contains a perceptual loss function and a total variation (TV) loss function. Treneska et al.\cite{Treneska} introduced a technique that employs GAN to produce color images. Wu et al.\cite{Wu3} proposed an innovative method for colorizing remote sensing images by employing deep convolution GAN.  Zang et al.\cite{Zang} created the bi-stream generative adversarial network colorization model (BS-GAN) specifically for the task of adding color to photographs. The BS-GAN generator may capture both global and local characteristics by combining two parallel streams.

\paragraph{Transformer and Diffusion based Colorization}
According to Yun et al.\cite{icolorit}, iColoriT is a vision-transformer-based colorization method that uses a global receptive field to effectively distribute user-provided color hints across the image. Kang et al. propose DDColor\cite{DD}, an end-to-end dual decoder method for precise and vibrant image colorization. Tang et al.\cite{codebook} introduce a two-stage strategy with a pixel decoder and a query-based color decoder, refining colorization with a predefined color codebook. Lee et al.\cite{Lee} present AdaColViT, an efficient ViT architecture for real-time interactive colorization, reducing redundant image patches and layers for faster processing. Chang et al.\cite{lcad} develop a unified model for language-based colorization, interpreting descriptive text input for versatile applications. Carrillo et al.\cite{diffusart} offer Diffusart, an interactive line art colorization method using Diffusion Probabilistic Models (DPMs). Liang et al.\cite{colorcontrol} introduce CtrlColor, a multi-modal colorization method using the pre-trained stable diffusion model, enabling fine-grained control over the colorization process.

\begin{figure*}[!t]
	\centering
	\includegraphics[width=\textwidth, height=8cm]{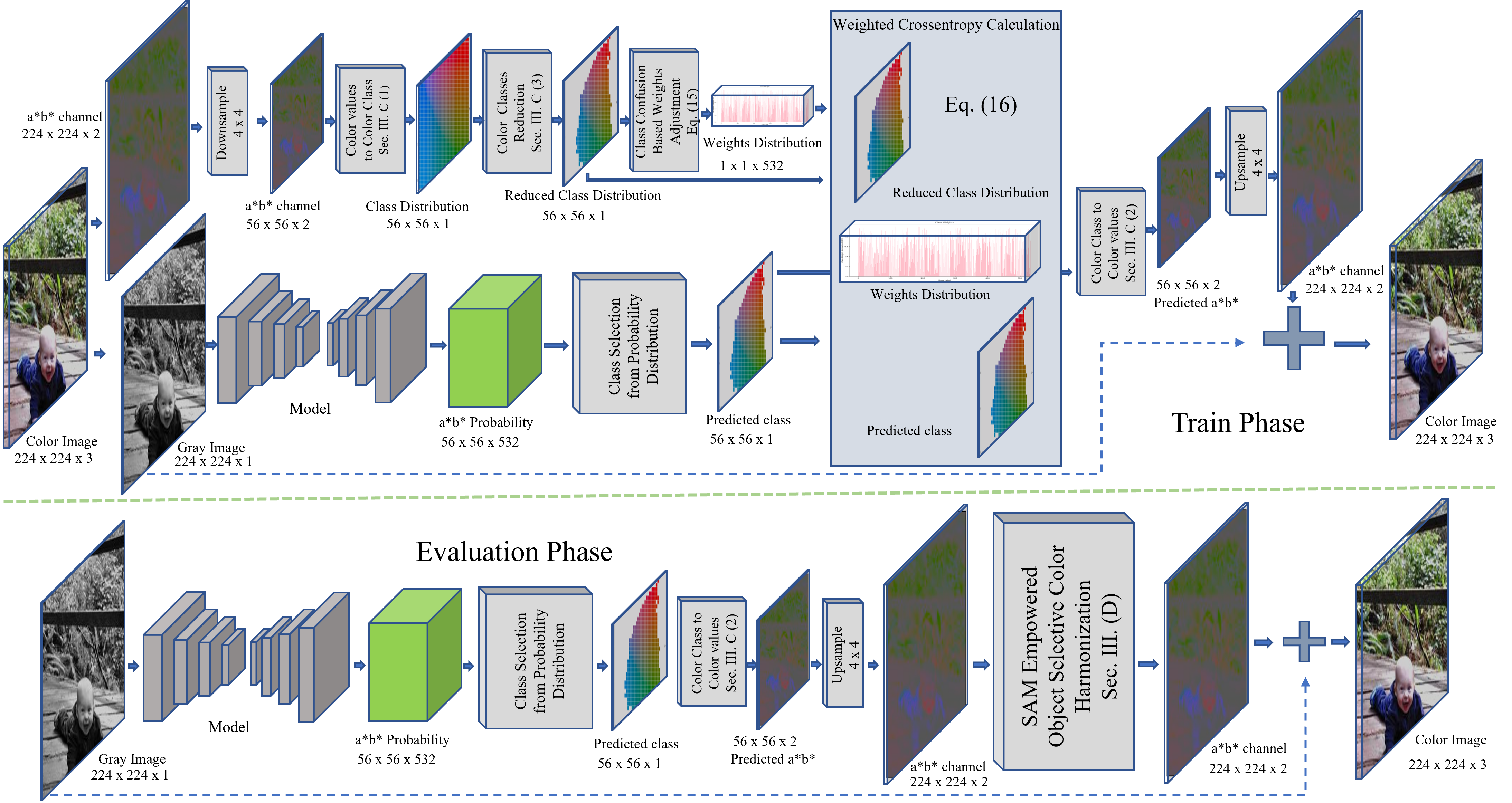}
	\caption{CCC++ with SAM empowered object selective color harmonization.}
	\label{ccc++}
	\vspace{-2mm}
\end{figure*}

\paragraph{Feature Balancing for Colorization}
 Zhang et al.\cite{Zhang_eccv} proposed an automatic colorization method (CIC) using CNN. The model learns to classify intensity into predetermined color level and then assigns corresponding color based on classified class level. Zhang's approach defines 313 classes. During training, they utilized class rebalancing to broaden the color palette of the output image for solving feature imbalance problem. An et al.\cite{An} used a VGG-16 CNN model based classification model for colorization. They set 313 classes as in Zhang et al. method. The method also utilized a color rebalancing technique to solve feature imbalance problem. Gustav et al. \cite{Larsson} used the unbalanced loss of classiﬁcation for solving feature imbalance. Their model predicts hue and chroma distributions for each pixel intensity. Gain et. al.\cite{Gain2} proposed a deep localized network which is a new model for image colorization. Instead of employing the global loss, they utilized the color-specific local loss to solve feature imbalance issue. These methods are promising for colorization problems. However, some methods show good performance on typical scenes (e.g. landscapes, sky, ground, etc.), but still their success is limited on complex images with foreground objects.

In this work, we introduce a data-driven feature balancing approach designed to achieve seamless and harmonized automatic colorization. Our method leverages advanced algorithms to dynamically balance features, ensuring that colorization results are both natural and visually coherent across a wide range of images.

\section{Methodology} \label{Methodology}
\subsection{Color Space}
The most commonly and frequently used color space is conventional RGB, where red, green, and blue are the three main colors. All conceivable combinations of the three colors are used in standard RGB. However, in RGB space, color information and content information cannot be separated. Therefore, it is not suitable for color manipulation in colorization tasks because there is a chance to change the context information during color manipulation. Therefore, common usage is CIELAB\cite{CIE} color space instead of RGB. Because color information can be separated from context information in La*b* (LAB) space and color information can be manipulated while keeping context information unchanged. In La*b* space, L denotes the brightness or luminosity of the image. Intensities fall between $[0, 100]$, where the value $0$ designates black and $100$ designates white. Colors get brighter as $L$ rises. The a* denotes the image's proportion of red or green tones. The red is represented by a positive a* value and the green is represented by a significant negative a* value. The b* denotes the image's proportion of yellow or blue tones. Yellow is represented by a high positive b* value. Blue is represented by a significant negative b* value. Although a* and b* does not have a single range, values frequently lie between $[-128, 127]$.
\begin{figure}[!t]
	\centering
	\begin{subfigure}
		\centering
		\includegraphics[width=.49\linewidth,height=3cm]{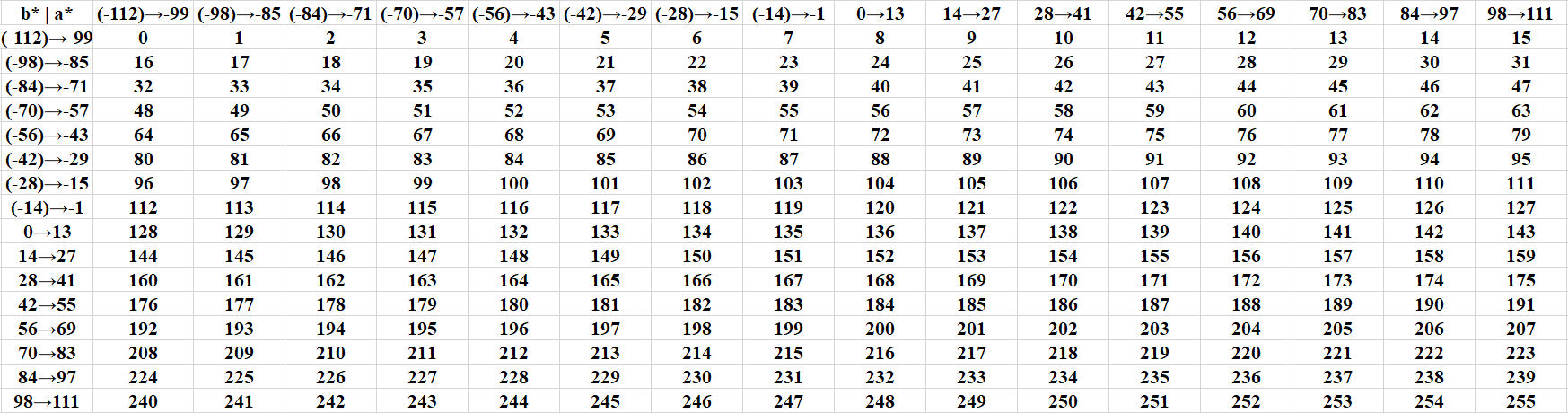}
		\includegraphics[width=.49\linewidth,height=3cm]{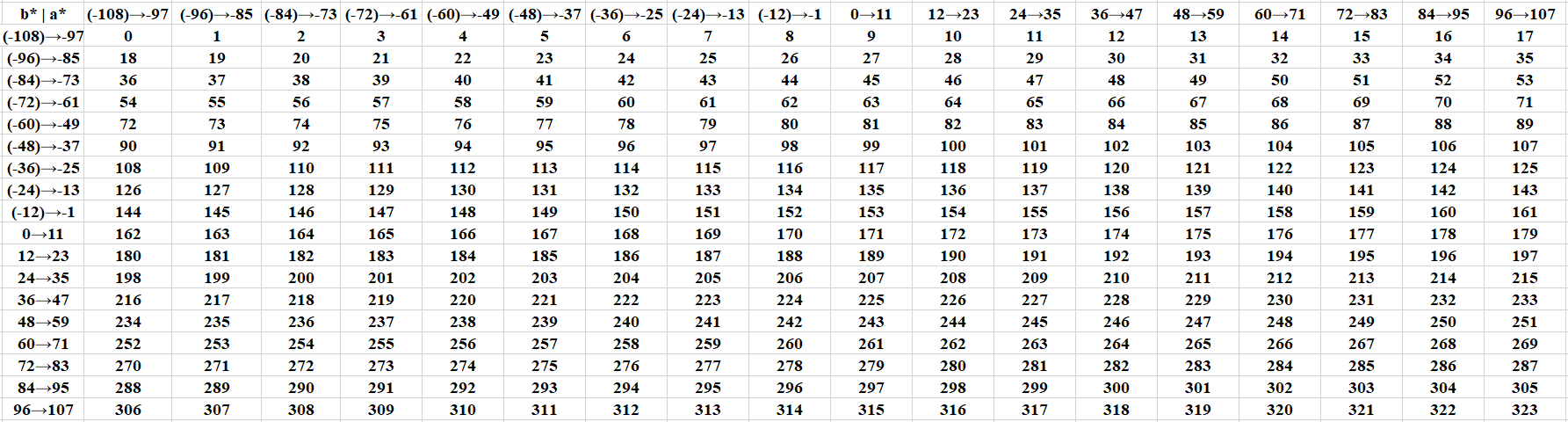}\\
		{Left: bin size = 14;  Right: bin size = 12}
		\includegraphics[width=.49\linewidth,height=3cm]{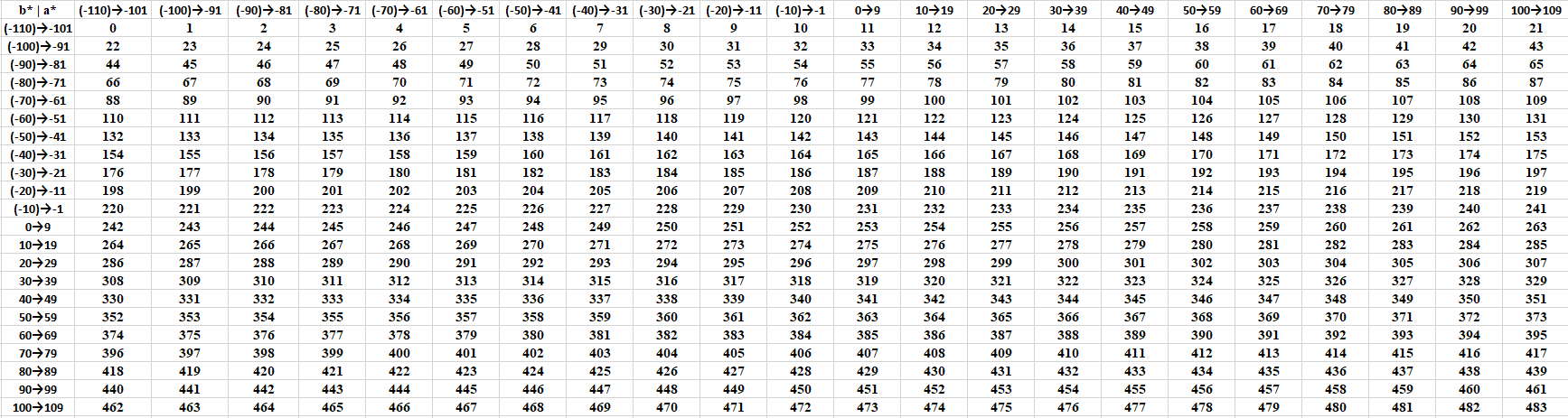}
		\includegraphics[width=.49\linewidth,height=3cm]{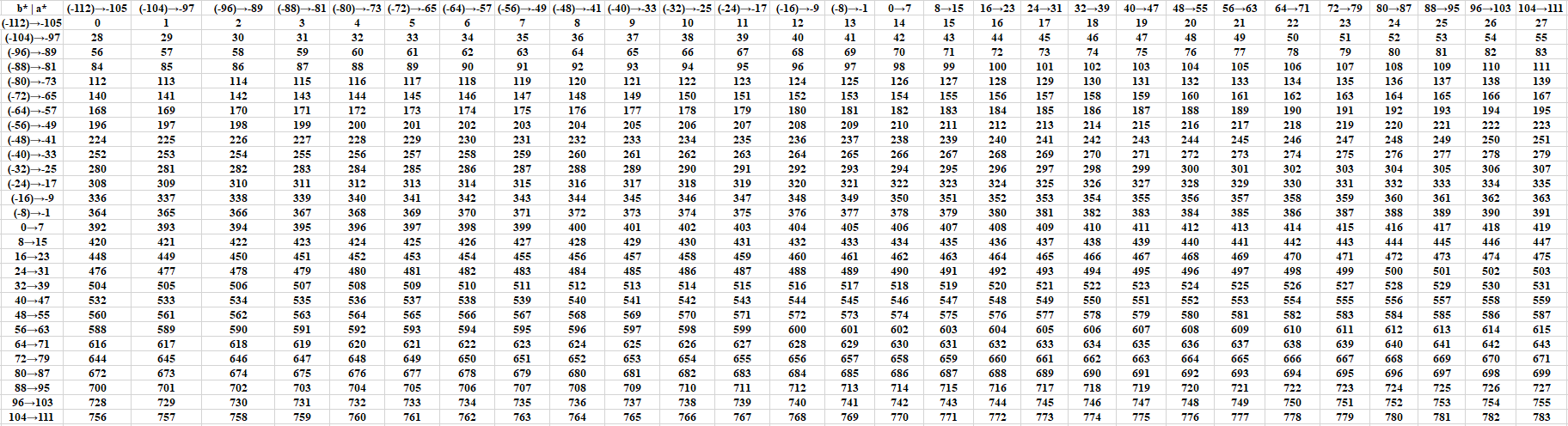}\\
		{Left: bin size = 10;  Right: bin size = 8}
		\includegraphics[width=.49\linewidth,height=3cm]{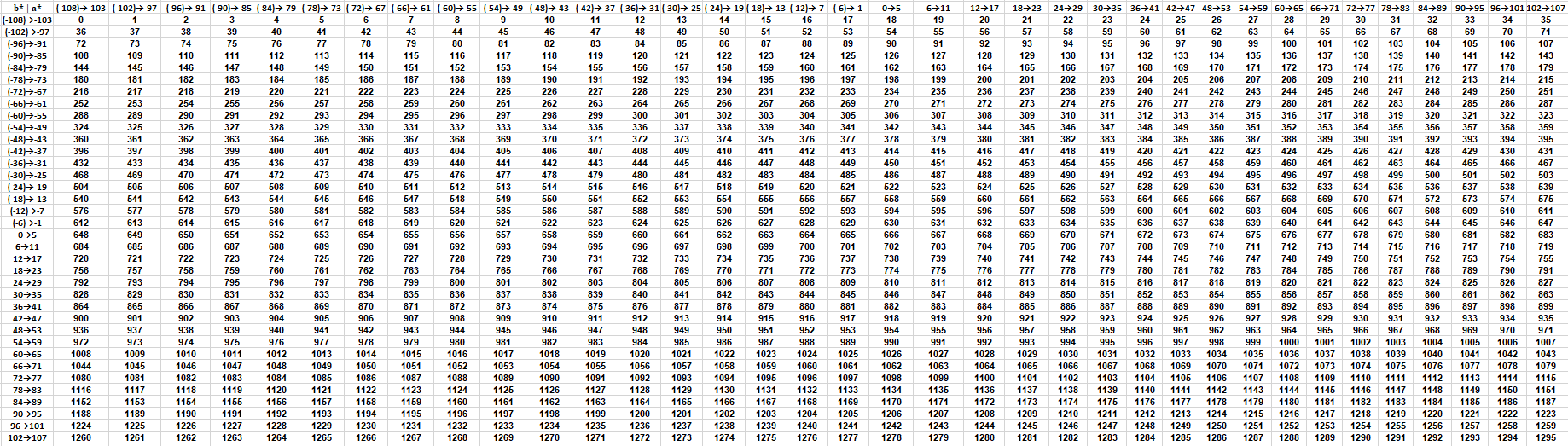}
		\includegraphics[width=.49\linewidth,height=3cm]{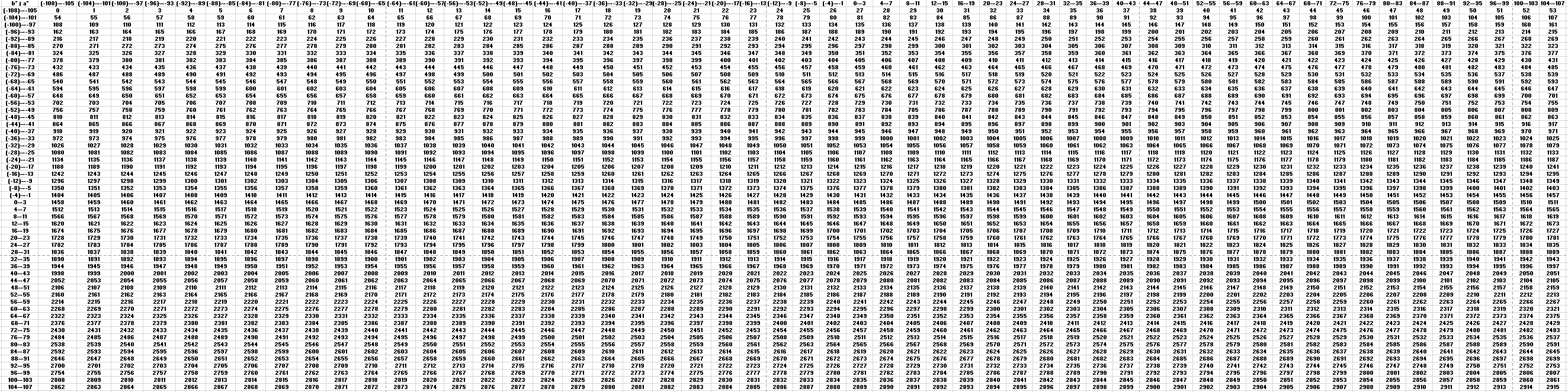}\\
		{Left: bin size = 6;  Right: bin size = 4}
	\end{subfigure}
	\caption{Continuous color values to discrete color classes for different bin sizes.}
	\label{cont_dis}
	\vspace{-2mm}
\end{figure}
\subsection{Problem Definition}
The colorization problem involves predicting color channels based on a given gray channel. The Lightness (L) channel of the La*b* color space can be converted to the gray channel (intensity) and vice versa\cite{CIE}. Additionally, it is possible to convert RGB values into LAB color space and vice versa. The task can be defined as follows in Equation \ref{problem_formulation_1}, \ref{problem_formulation_2}, \ref{problem_formulation_3}, and \ref{problem_formulation_4}. 
\begin{equation}
\label{problem_formulation_1}
X_{ab} = f(X_L) 
\end{equation}
\begin{equation}
\label{problem_formulation_2}
Distance_{min} = (Y_{ab}, X_{ab})
\end{equation}
\begin{equation}
\label{problem_formulation_3}
X_{Lab} = concat(X_L, X_{ab})
\end{equation}
\begin{equation}
\label{problem_formulation_4}
X_L \in \mathbb{R}^{H\times W \times 1} ,
X_{ab} \in  \mathbb{R}^{H\times W \times 2},
Y_{ab} \in  \mathbb{R}^{H\times W \times 2}
\end{equation}
 where $X_L$ represents the lightness channel, $X_{ab}$ represents the predicted color channel, $X_{Lab}$ represents the predicted color image, $Y_{ab}$ represents the ground truth color channel. The function $f(.)$ represents the mapping function achieved through deep learning. The function $Distance_{min}(.)$ represents the objective function, which can be any loss function the optimizer uses to make the learning process efficient. The set $\mathbb{R}$ represents the entirety of the image component, whereas $\mathbb{H}$ and $\mathbb{W}$ represent the dimensions of the image.

\begin{figure}[!t]
    \centering
    \begin{subfigure}
        \centering
        \includegraphics[width=.49\linewidth,height=3cm]{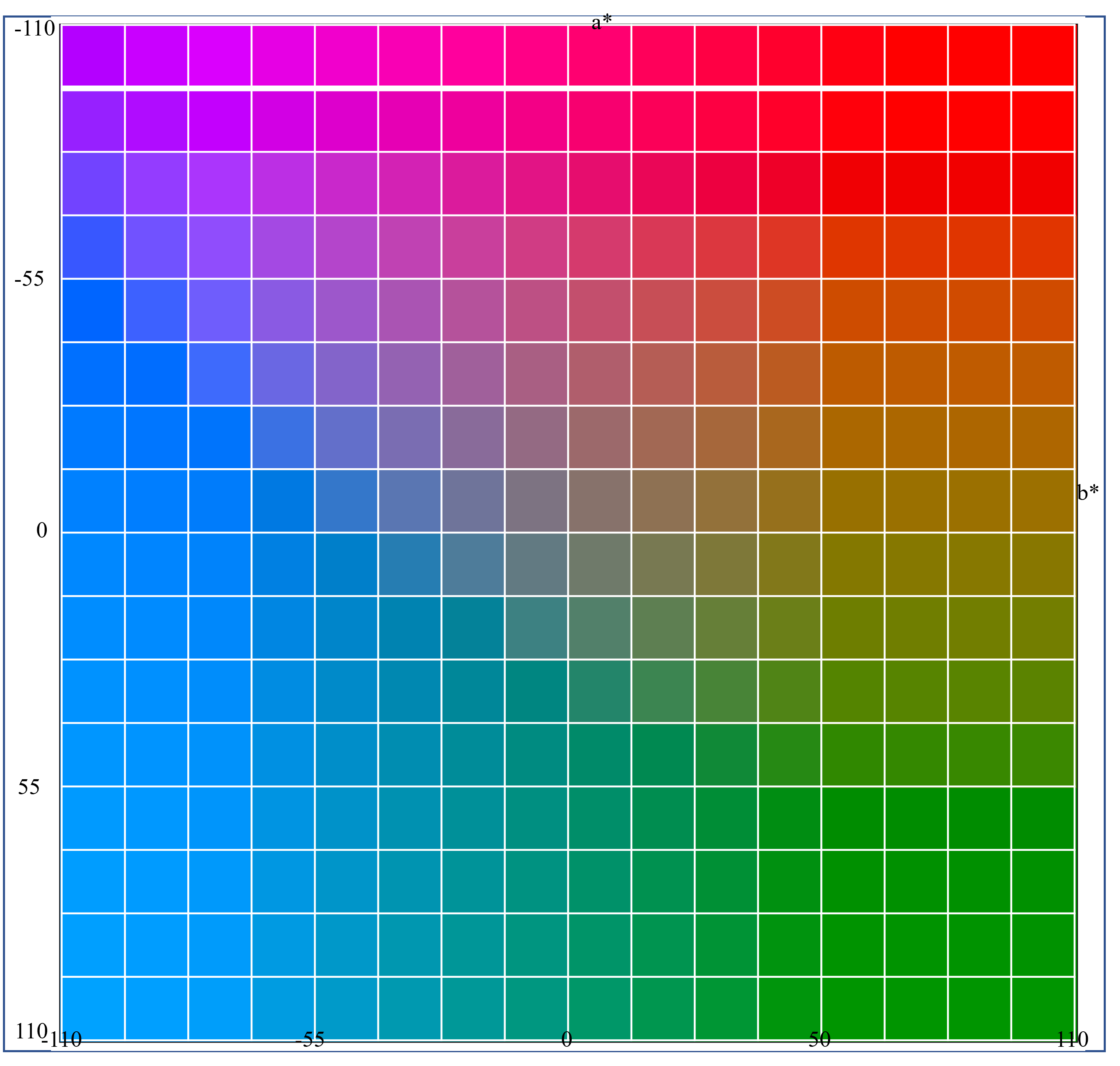}
        \includegraphics[width=.49\linewidth,height=3cm]{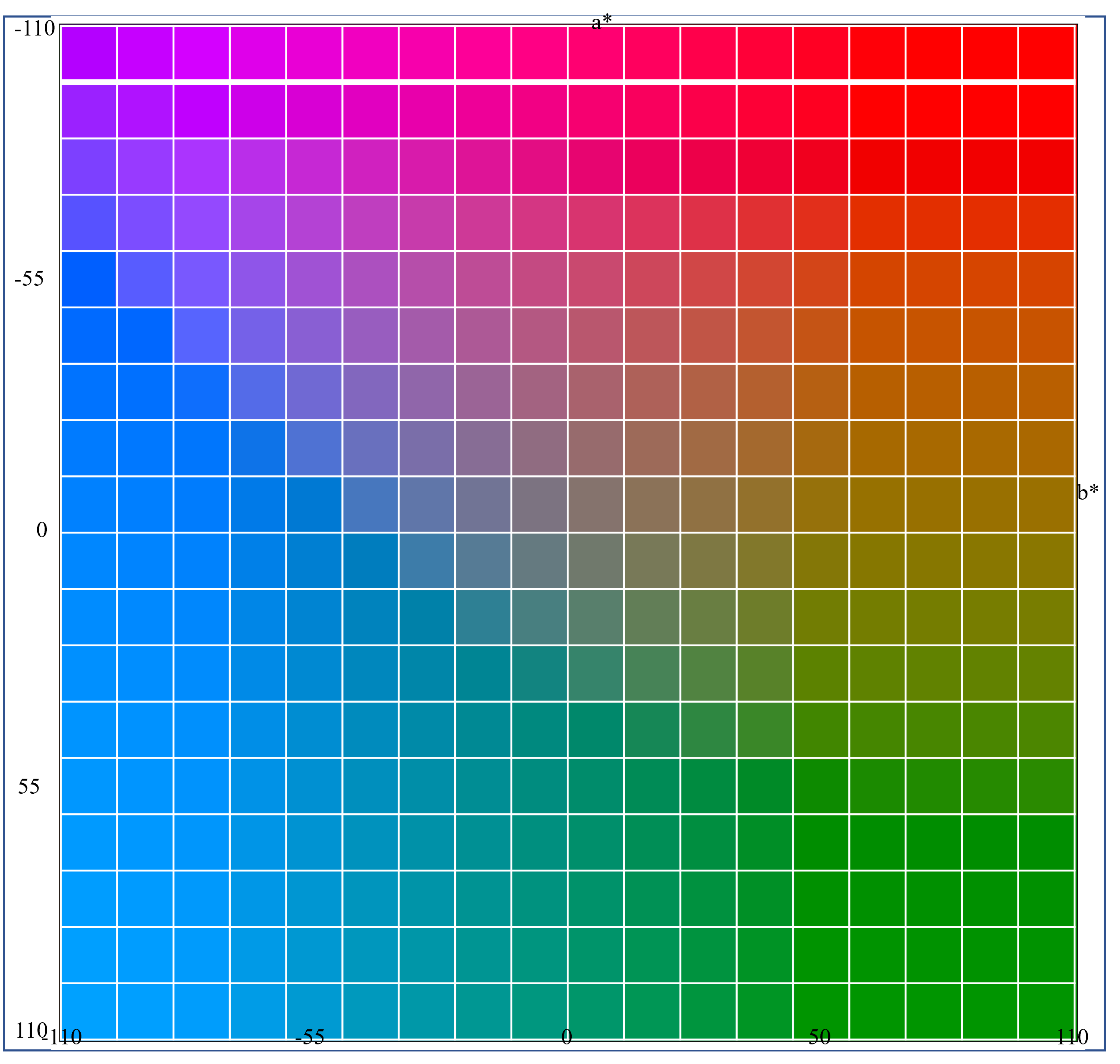}\\
        {Left: bin size = 14;  Right: bin size = 12}
        \includegraphics[width=.49\linewidth,height=3cm]{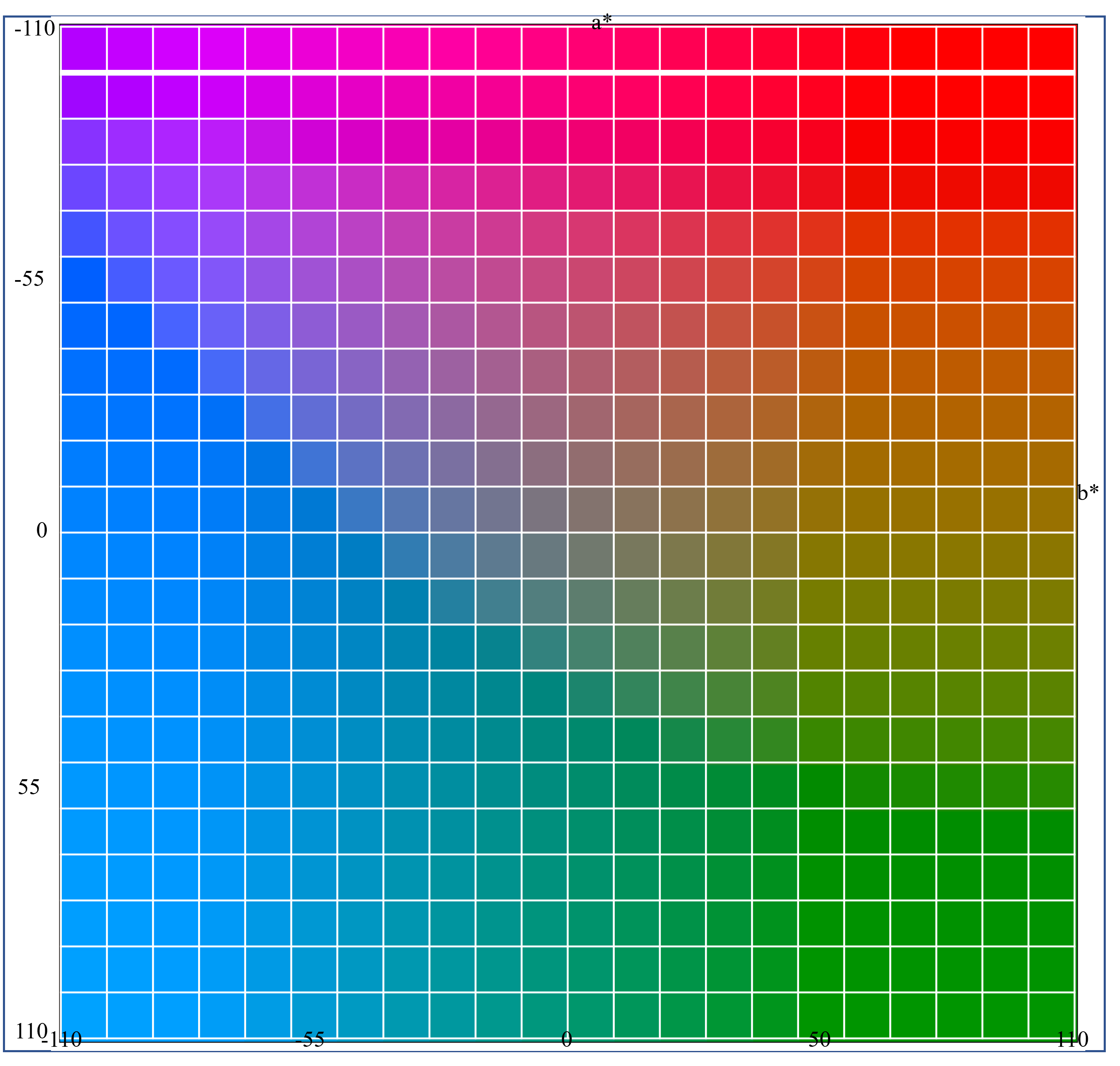}
        \includegraphics[width=.49\linewidth,height=3cm]{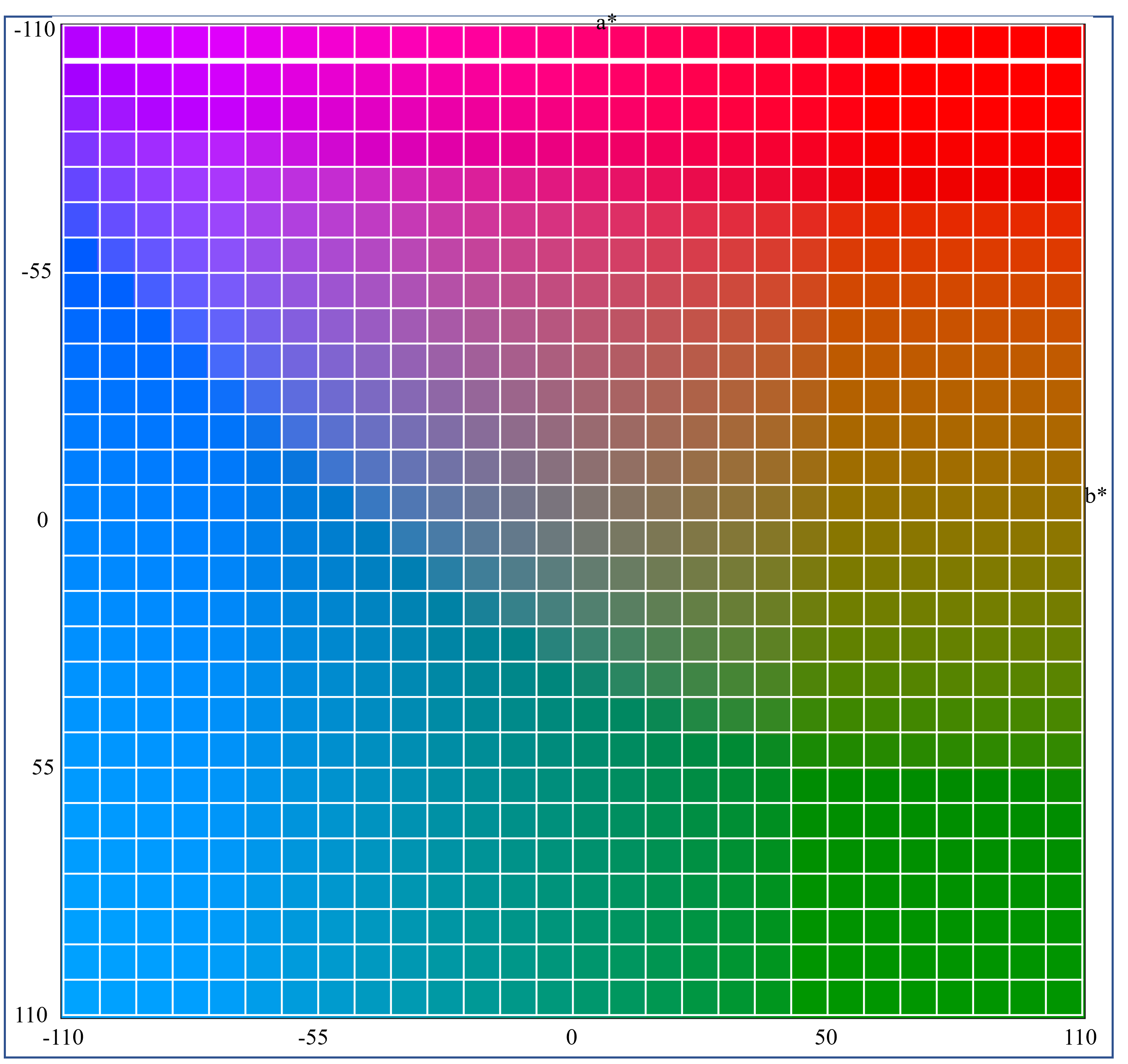}\\
        {Left: bin size = 10;  Right: bin size = 8}
        \includegraphics[width=.49\linewidth,height=3cm]{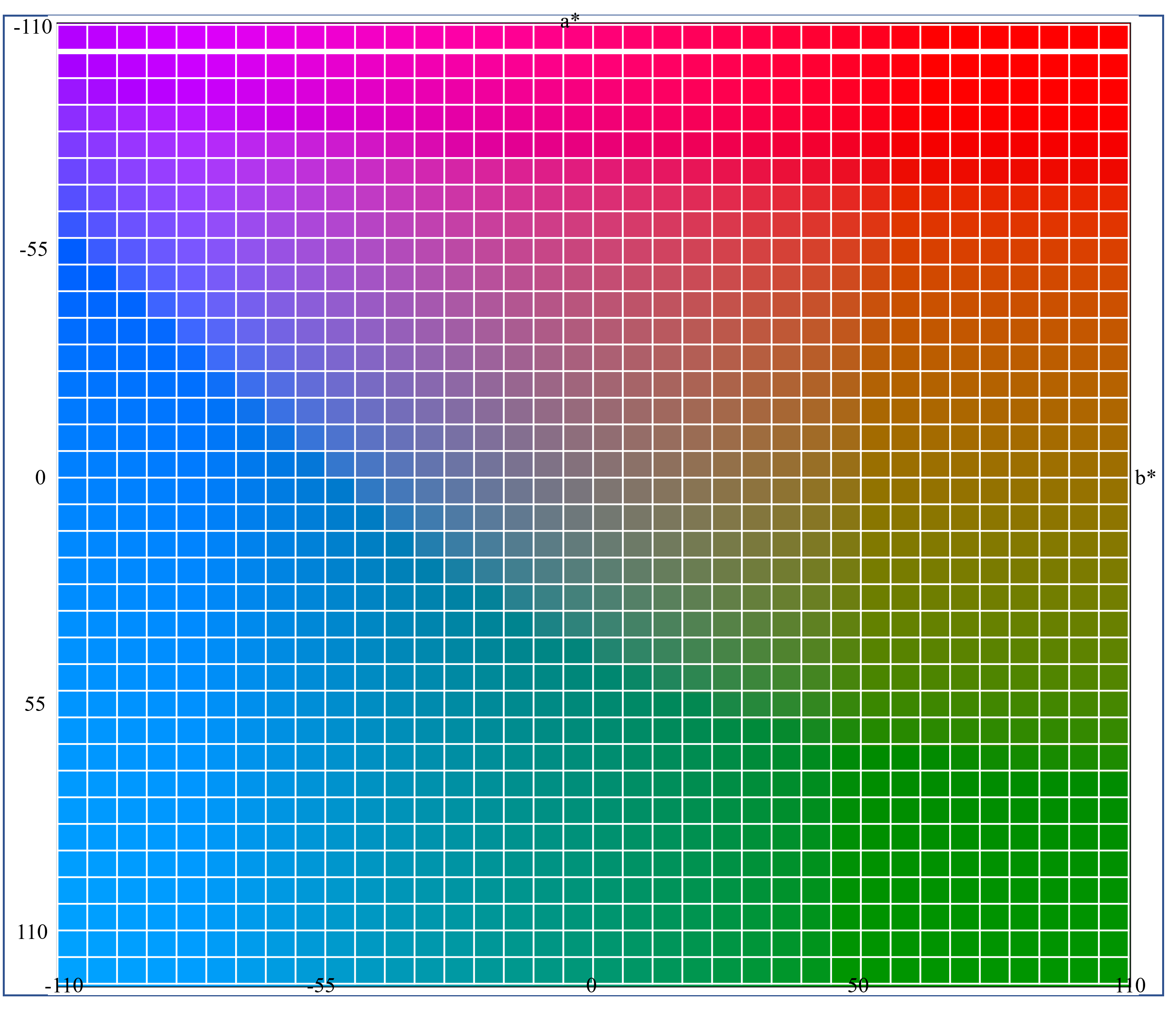}
        \includegraphics[width=.49\linewidth,height=3cm]{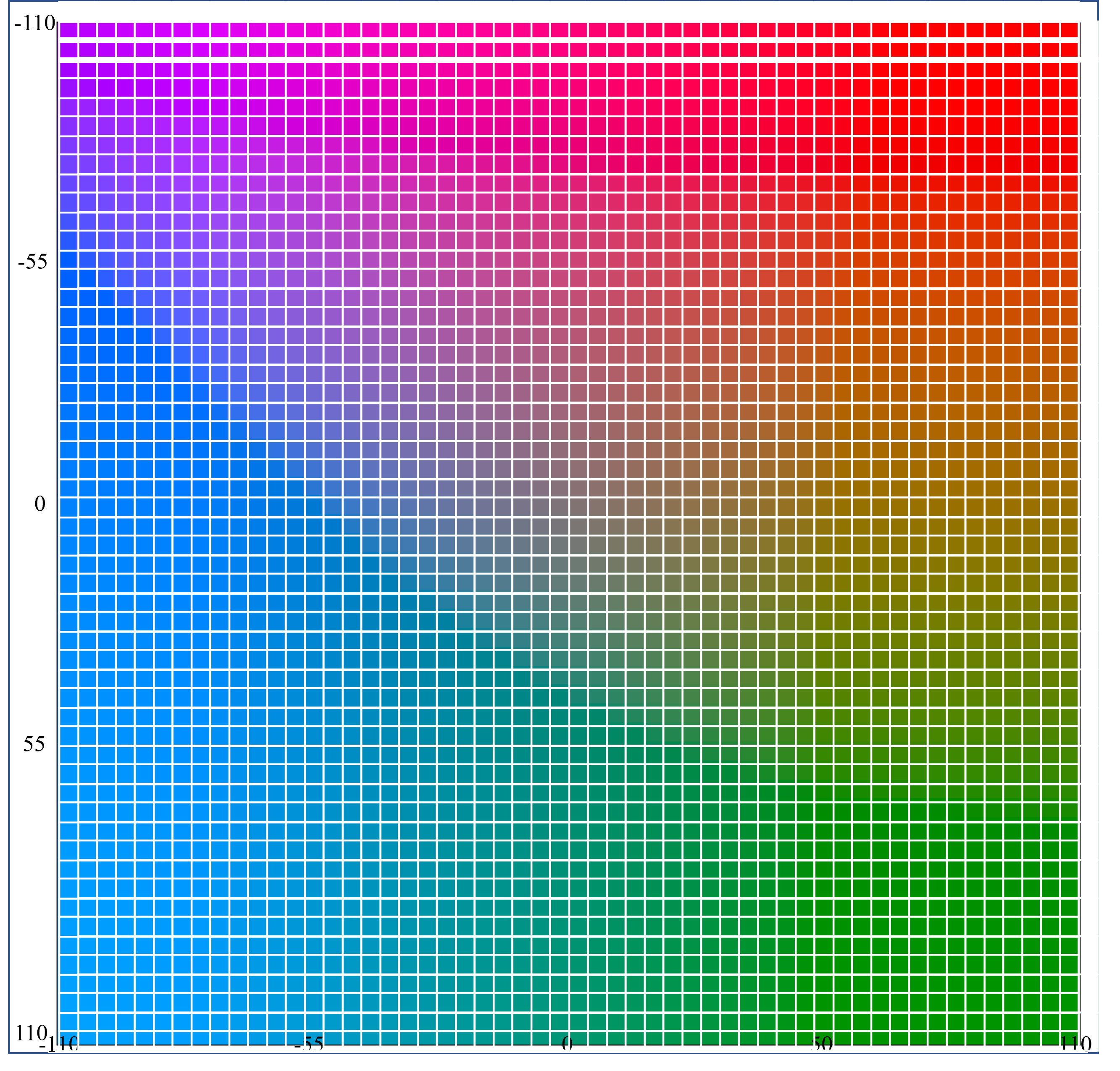}\\
        {Left: bin size = 6;  Right: bin size = 4}
    \end{subfigure}
    \caption{Color visualization of discrete color classes for different sizes.}
    \label{visualize1}
    \vspace{-2mm}
\end{figure}
Theoretically, the values of the a* and b* channels range from -128 to 127, inclusively, and are continuous. Therefore, the task at hand is classified as a regression problem. Hence, the $Distance_{min}(.)$ in equation \ref{problem_formulation_2} can be expressed as either L1 loss, L2 loss, Huber loss, Log-cosh loss, or a similar regression loss. The loss functions are displayed in Equation \ref{l1}, \ref{l2}, \ref{huber} and , \ref{log-cosh}. 
\begin{equation}
\label{l1}
L_1(Y_{ab}, X_{ab}) = \frac{1}{N}\sum_{N}|Y_{ab} - X_{ab}|
\end{equation}
\begin{equation}
\label{l2}
L_2(Y_{ab}, X_{ab}) = \frac{1}{N}\sum_{N}(Y_{ab} - X_{ab})^2
\end{equation}
\begin{equation}
\label{huber}
L_{\delta}=
    \left\{\begin{matrix}
        \frac{1}{N}\sum_{N}\frac{1}{2}(Y_{ab}-X_{ab})^{2}, & |Y_{ab}-X_{ab}|<\delta\\
        \frac{1}{N}\sum_N\delta ((Y_{ab} - X_{ab}) - \frac1 2 \delta), & otherwise
    \end{matrix}\right.
\end{equation}
\begin{equation}
\label{log-cosh}
Log-Cosh(Y_{ab}, X_{ab}) =  \frac{1}{N}\sum_{N}log(cosh(Y_{ab} - X_{ab}))
\end{equation}
The variable $N$ represents the total number of pixels in a batch, whereas the variable $\delta$ represents the acceptable threshold. The real-time photographs predominantly feature background colors that resemble clouds, earth, pavement, and walls. The color components of the major or focal objects are significantly less than the total color components, which leads to an uneven allocation of characteristics. Addressing feature imbalance is essential as the smaller groups of features are the focal point for the learning process. The intrinsic ambiguity and multimodality of the colorization problem render the above loss functions susceptible. If an object can have different values for a*b*, the most efficient way to solve the loss indicated in equations \ref{l1}, \ref{l2}, \ref{huber}, and \ref{log-cosh} is by calculating the mean of the set. The averaging error effect prefers color values that are mostly present in the ground truth image. When there is an uneven distribution of features, the training process becomes biased towards the larger sets of features. As a result, the models generated from this method lose the colors of smaller objects. Hence, the distribution of a*b* values is biased towards desaturated values, resulting in the disappearance of color in tiny objects.
\subsection{Solution Approach}
The system architecture of our proposed approach is shown in Fig. \ref{ccc++}. The details of our proposed model is described below.

\begin{figure}[!t]
	\centering
	\begin{subfigure}
		\centering
		\includegraphics[width=.49\linewidth,height=3cm]{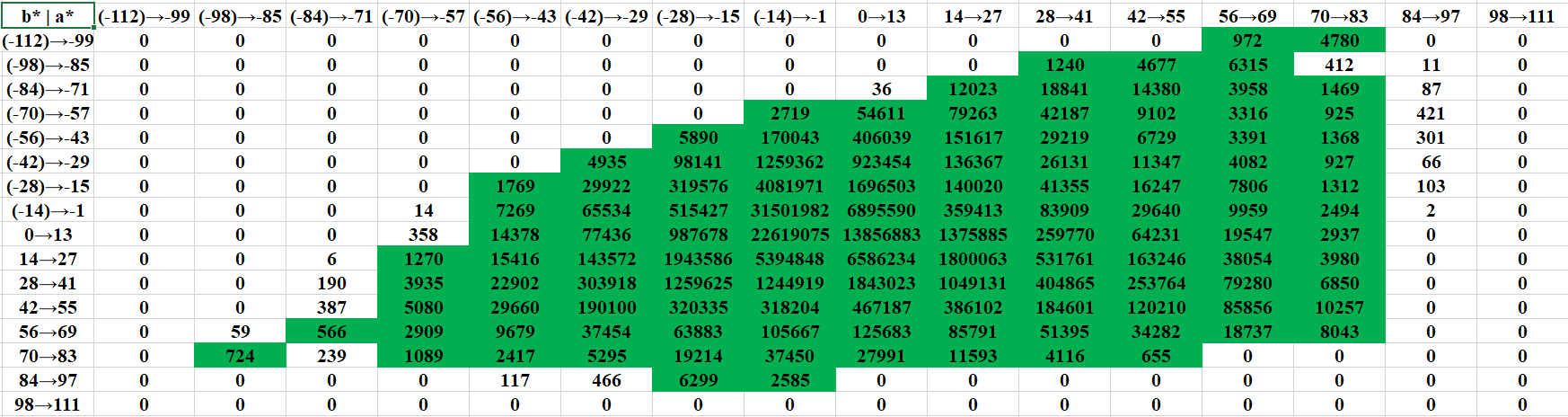}
		\includegraphics[width=.49\linewidth,height=3cm]{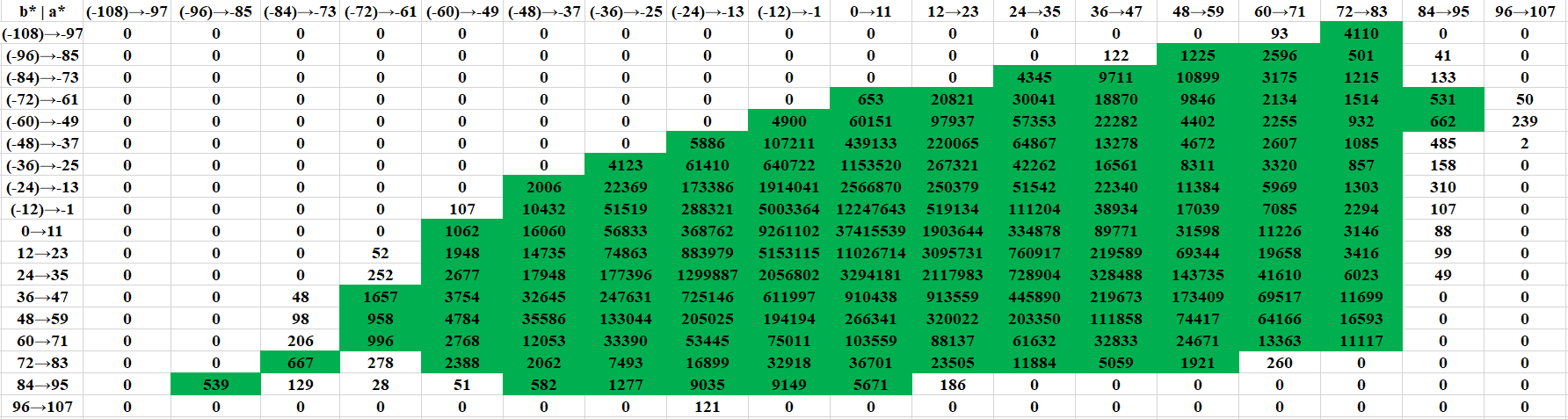}\\
		{Left: bin size = 14;  Right: bin size = 12}
		\includegraphics[width=.49\linewidth,height=3cm]{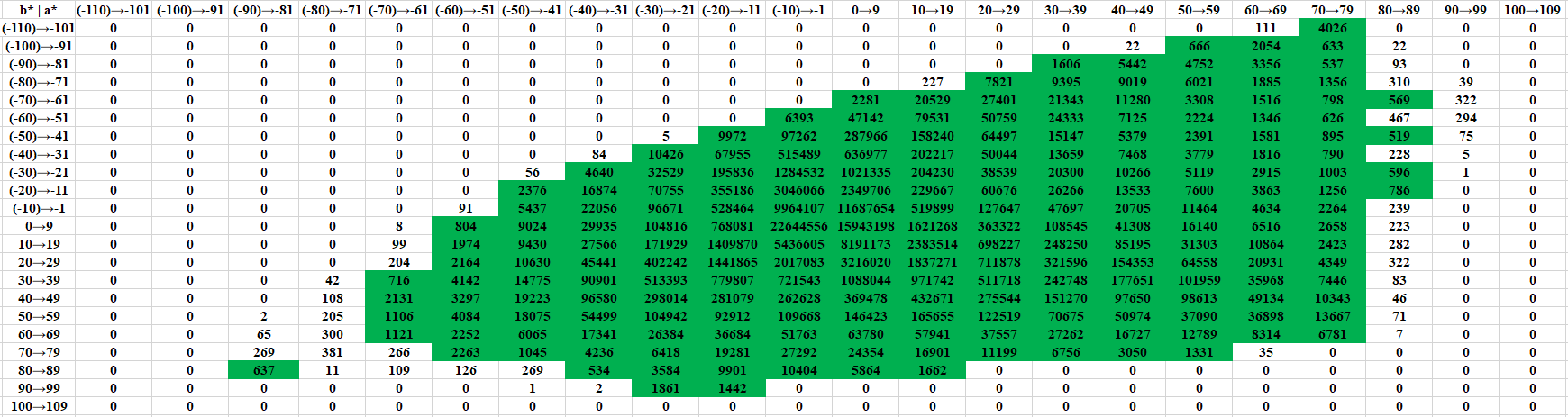}
		\includegraphics[width=.49\linewidth,height=3cm]{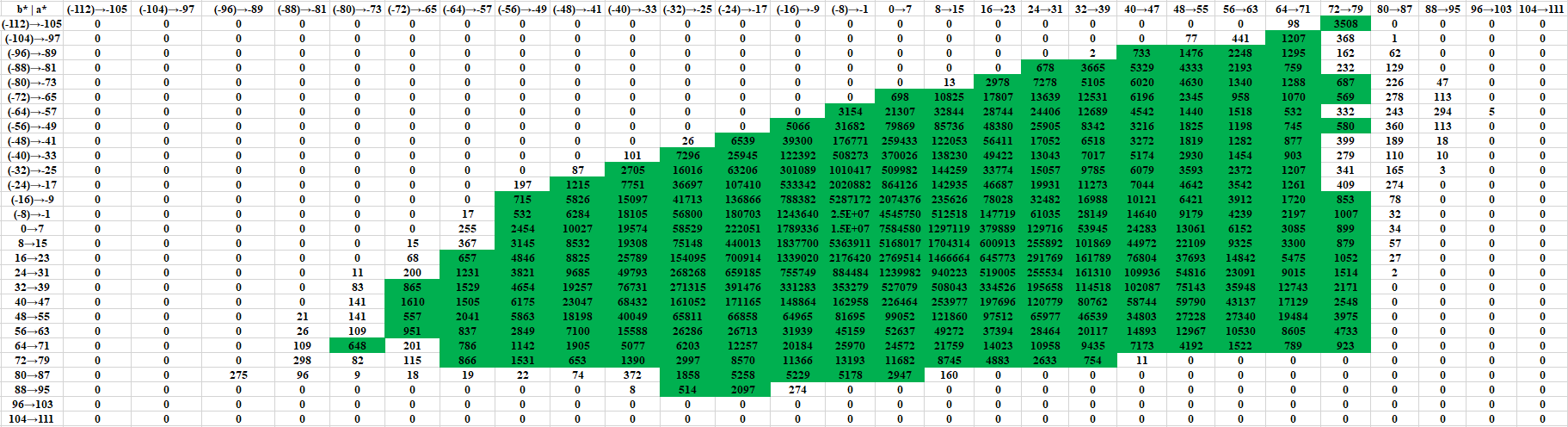}\\
		{Left: bin size = 10;  Right: bin size = 8}
		\includegraphics[width=.49\linewidth,height=3cm]{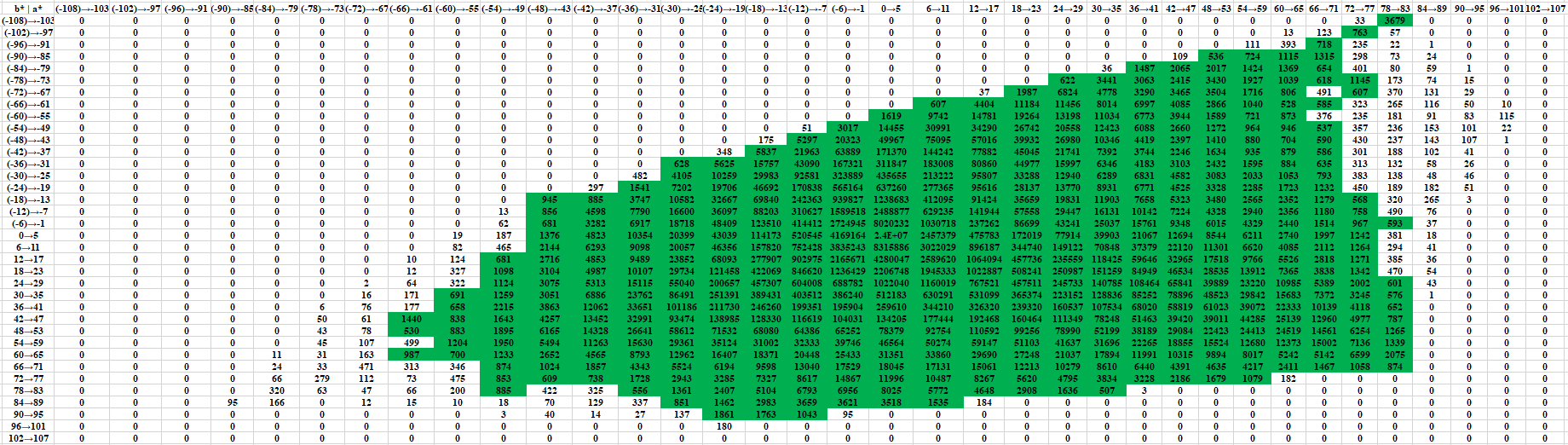}
		\includegraphics[width=.49\linewidth,height=3cm]{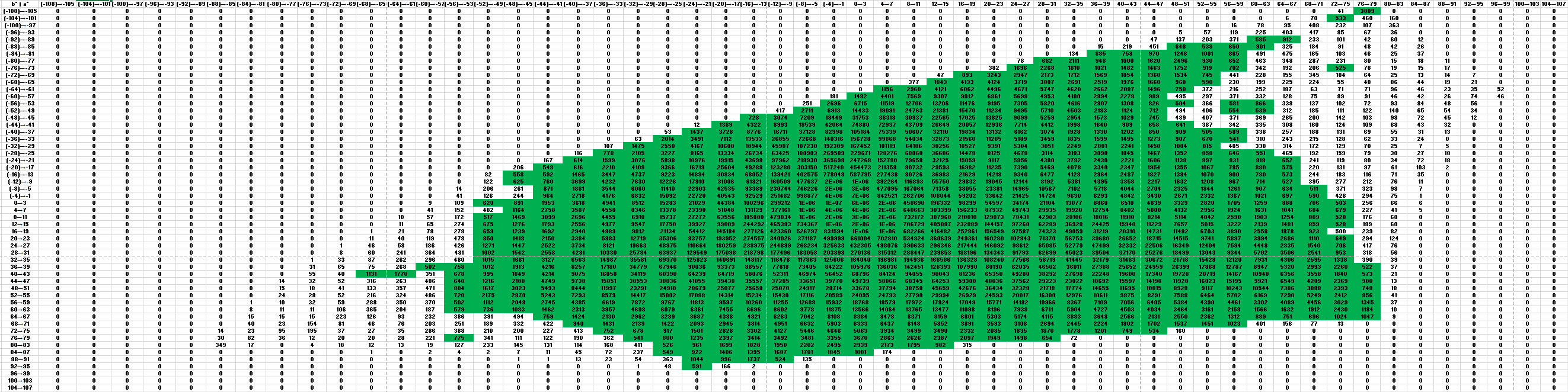}\\
		{Left: bin size = 6;  Right: bin size = 4}
		\label{subfig:ab_to_class_reduced}
	\end{subfigure}
	\caption{Real-time appearance of Color classes in place365 validation dataset.}
	\label{rta}
	\vspace{-2mm}
\end{figure}
\subsubsection{Continuous Color Range to Discrete Color Classes} \label{Continuous Color Range to Discrete Color Classes}
The color channels a* and b* exhibit continuous values throughout the range of -128 to 127. An RGB color pixel is formed by each pair of a*b* values with a lightness value $L$. A pair of a*b* can be obtained from a*b* color space, a two-dimensional space where a* represents one direction and b* represents another. For a constant value of $L$, a slight alteration in the pair of variables a and b produces no noticeable visual impact. Thus, the human perception of information in a picture typically does not entail a quantitative examination of each pixel value inside the image. Colorization is a regression problem in which a regression model predicts the continuous values of a* and b* based on a given $L$. By exploiting the psycho-visual characteristics of humans, the colorization problem can be framed as a classification problem in which the learning model predicts a discrete class level for a given a*b* pair. The a*b* color space is partitioned into bins of predetermined grid size, and each bin is assigned a discrete class level to formulate the problem. The formula is provided in Equation \ref{color_to_class}. 
\begin{equation}
\label{color_to_class}
C = \frac{b^*+\beta}{\alpha}\times \Delta + \frac{a^*+\beta}{\alpha}
\end{equation}
where a* and b* represent the continuous color channels, $C$ represents the discrete class level of color values within a bin, $\alpha^2$ represents the area of a bin, $\beta$ represents a shifting constant that shifts a*b* color values into a positive quadrant, $\delta$ represents the number of grids in each a* or b* color channel.

\begin{figure}[!t]
    \centering
    \begin{subfigure}
        \centering
        \includegraphics[width=.49\linewidth,height=3cm]{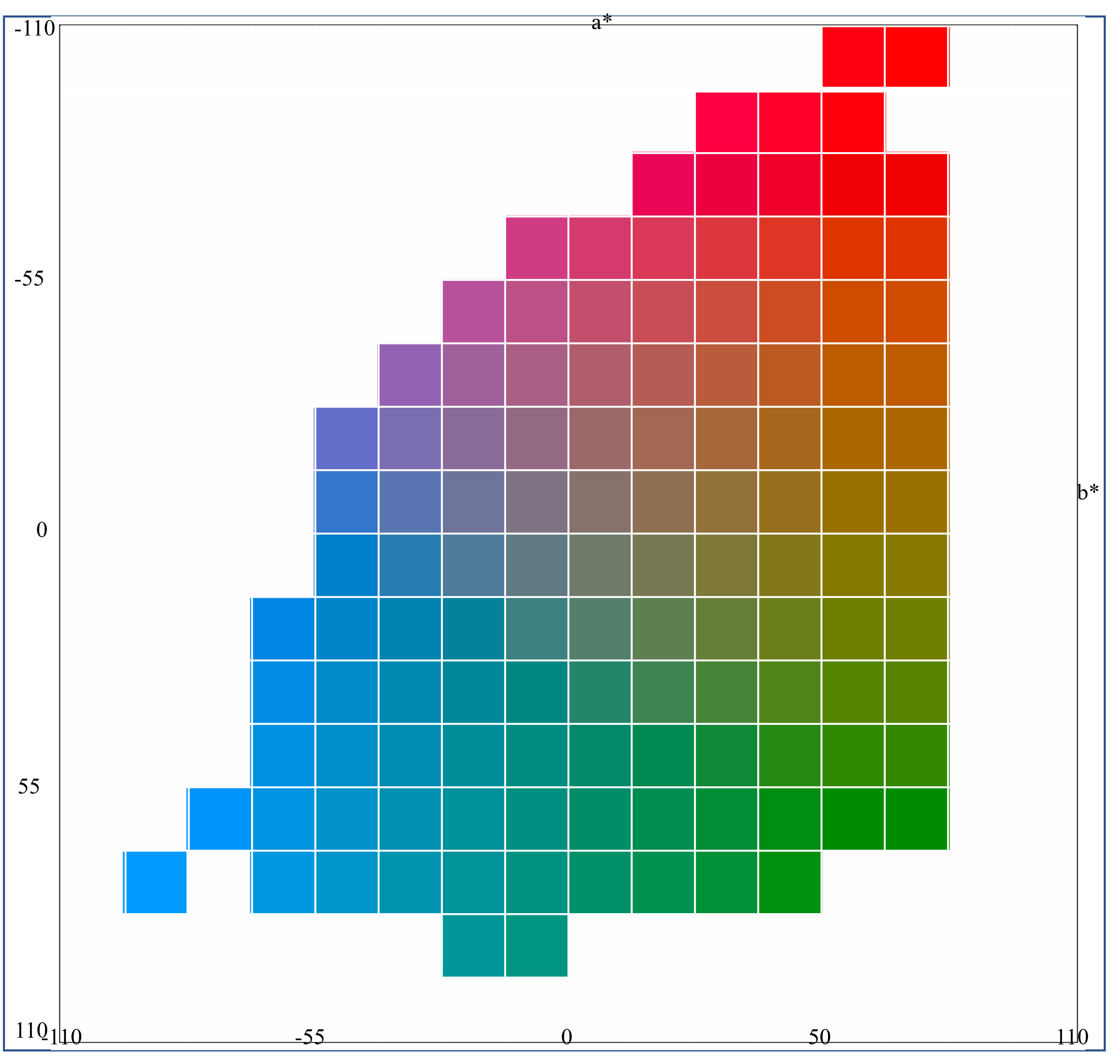}
        \includegraphics[width=.49\linewidth,height=3cm]{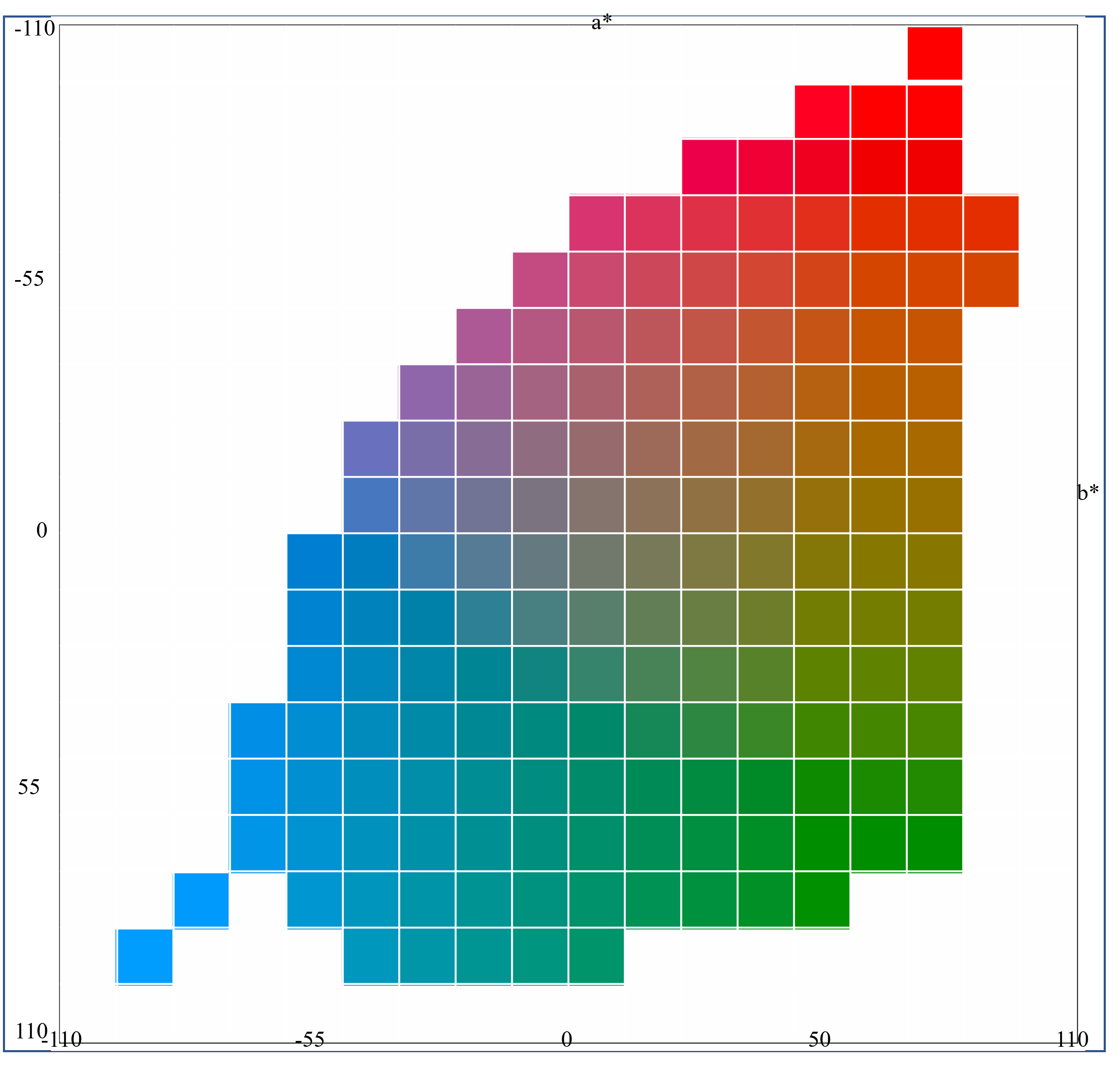}\\
        {Left: bin size = 14;  Right: bin size = 12}
        \includegraphics[width=.49\linewidth,height=3cm]{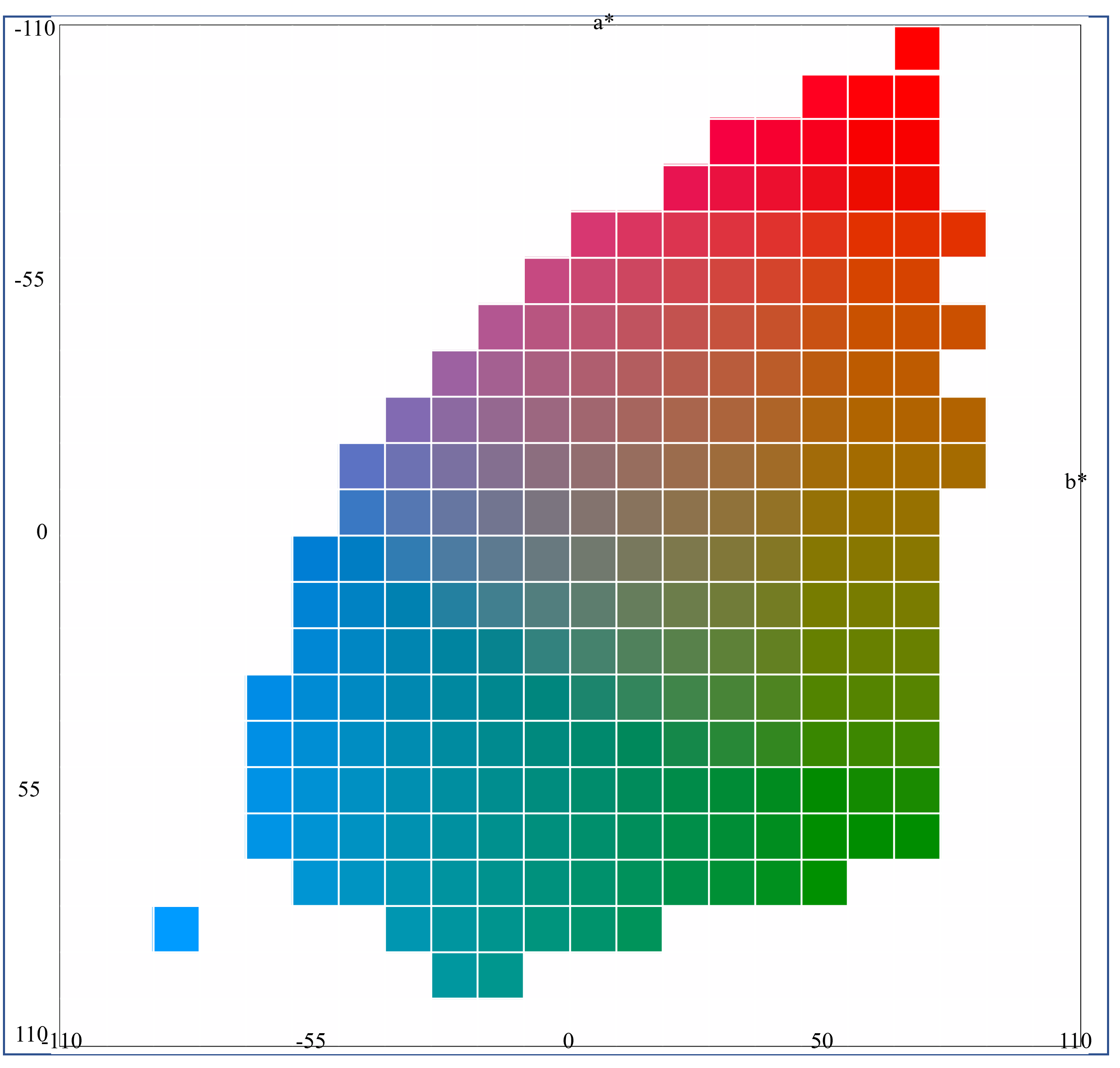}
        \includegraphics[width=.49\linewidth,height=3cm]{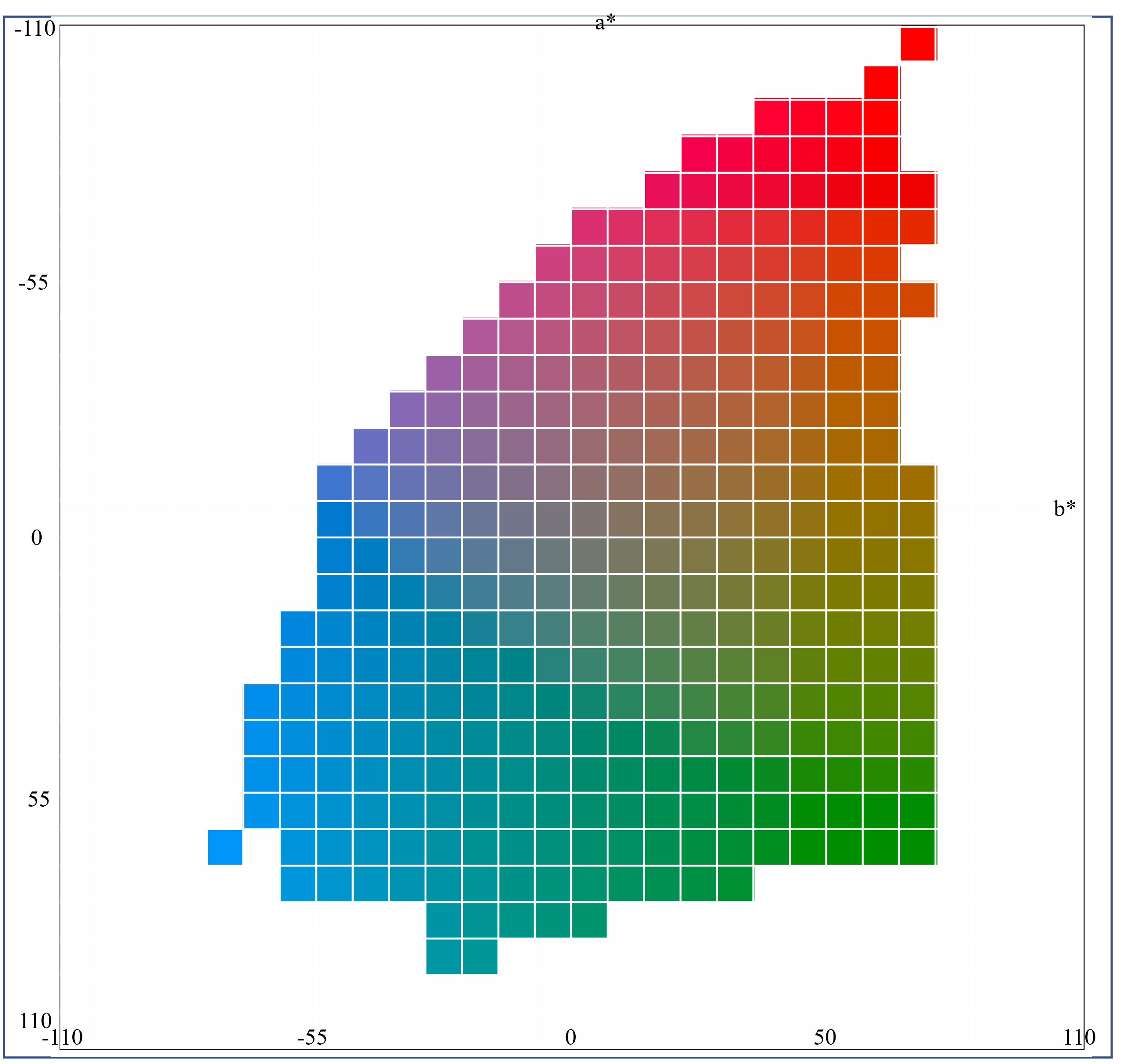}\\
        {Left: bin size = 10;  Right: bin size = 8}
        \includegraphics[width=.49\linewidth,height=3cm]{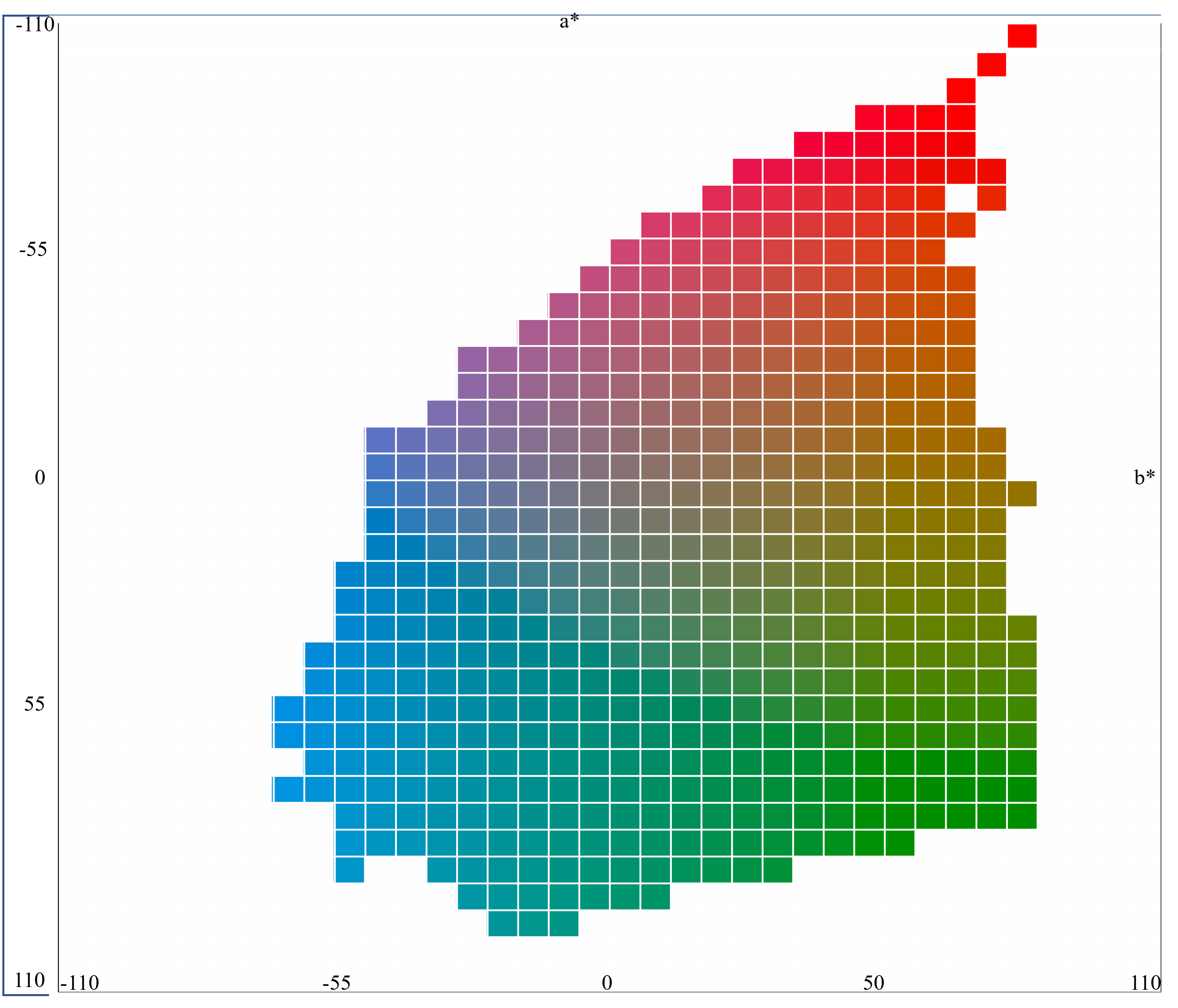}
        \includegraphics[width=.49\linewidth,height=3cm]{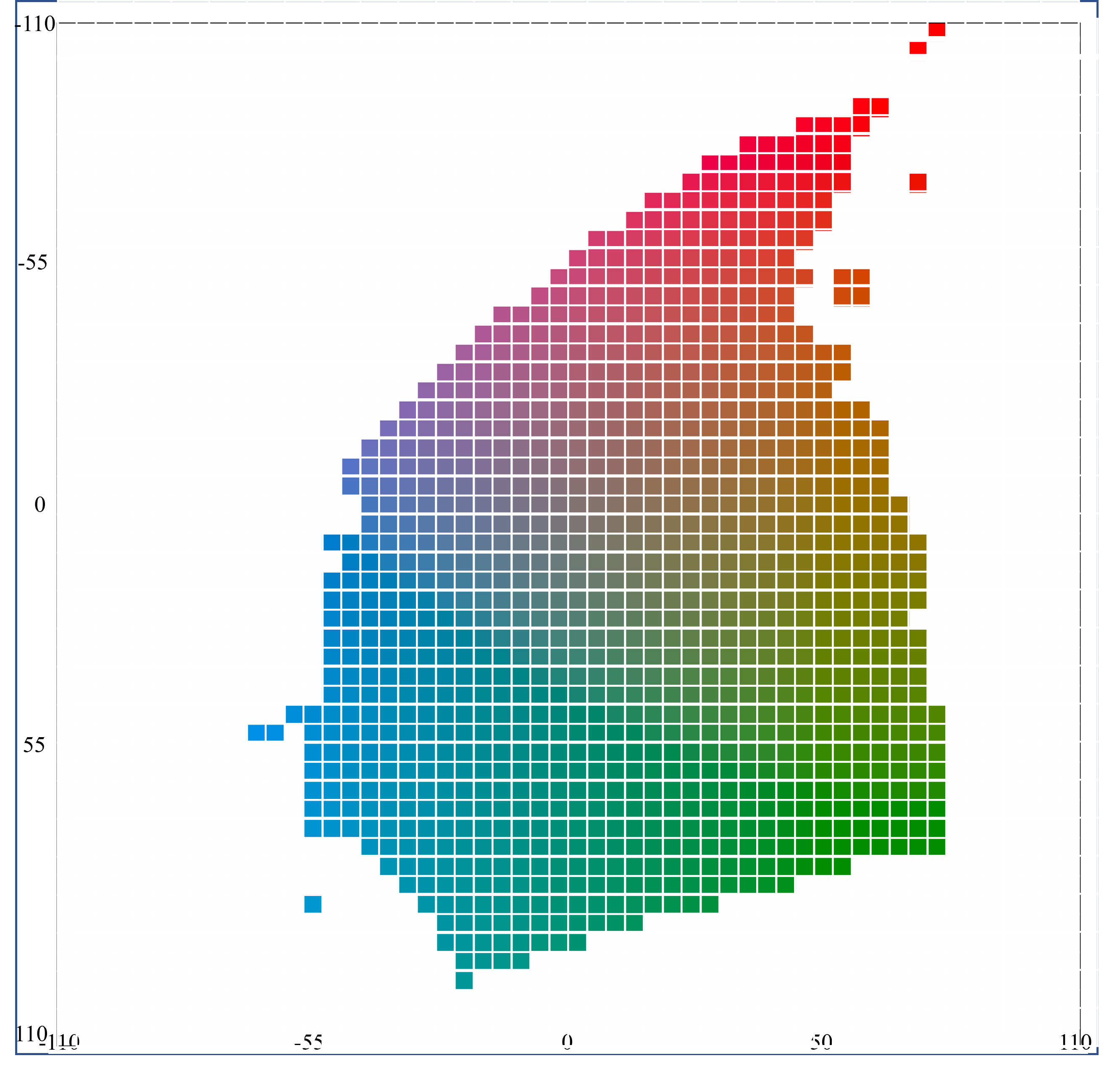}\\
        {Left: bin size = 6;  Right: bin size = 4}
    \end{subfigure}
    \caption{Visualization of real-time appearance of Color classes in place365 validation dataset.}
    \label{visualize2}
    \vspace{-2mm}
\end{figure}

\begin {table*}[h]
\caption {Bin size approximation} \label{tab:binsize} 
\begin{center}
\begin{tabular}{c c c c c c c c c c c c c} 
 \hline
   & Total & Optimized & \multicolumn{2}{c}  {a*b*}  & \multicolumn{8}{c} {RGB (L $\in$ 50)}\\
 \cline{4-13}
Bin & Class & Class & Max.  & Avg.  & \multicolumn{4}{c} {Maximum Deviation}& \multicolumn{4}{c} {Average Deviation}\\ 
\cline{6-13}
Size & Points& Points & Dev. & Dev. & a, b $\in$ 50 & a, b $\in$ 0 & a, b $\in$ -50  & Avg.& a, b $\in$ 50  & a, b $\in$ 0 & a, b $\in$ -50 & Avg.\\
 \hline
  4 & 2916 & 1049 & 2 & 1 & 2.39  & 2.43 & 1.9 & 2.24 & 1.59  & 1.60  & 0.71 & 1.30\\ 
 \hline
  6 & 1296 & 532 & 3 & 1.5 & 4.85  & 4.66 & 3.25 & 4.25 & 2.39  & 2.43  & 1.9 & 2.24\\ 
 \hline
  8 & 784 & 325 & 4 & 2 & 6.55 & 6.33 & 4.75 & 5.88 & 3.20 & 3.24  & 2.45 & 2.96\\ 
 \hline
  10 & 484 & 218 & 5  & 2.5 & 8.29 & 7.78 & 5.5& 7.19 & 4.01  & 3.95  & 2.92 & 3.62\\ 
 \hline
  12 & 324 & 159 & 6  & 3 & 10.10 & 9.44 & 6.65& 8.73 & 4.85  & 4.66  & 3.25 & 4.25\\ 
 \hline
  14 & 256 & 121 & 7  & 3.5 & 12.00 & 10.87 & 7.75& 10.20 & 5.70  & 5.60  & 4 & 5.10\\ 
 \hline
\end{tabular}
\end{center}
\vspace{-2mm}
\end {table*}

\subsubsection{Color Class to Visual Color Mapping} 
We need to retrieve pairs of a*b* values from the color classes, $C$s, predicted by the color generation learning model. Each bin is allocated a specific color class level $C$ based on the values of $a^*$ and $b^*$.  The formulas are provided below as Equation \ref{class_to_color_a} and \ref{class_to_color_b}, which represent the inverse of Equation \ref{color_to_class}.  

\begin{equation}
\label{class_to_color_a}
a^{*\prime} = [(C\mod\Delta)\times\alpha]-\beta +\frac{\alpha}{2}
\end{equation}
\begin{equation}
\label{class_to_color_b}
b^{*\prime} = [(C\div\Delta)\times\alpha]-\beta +\frac{\alpha}{2}
\end{equation}

Based on the given equations, the maximum loss for each value of a* or b* equals $\frac{\alpha}{2} -1$. Increasing the amount of $\alpha$ decreases the number of classes, but it also causes the representation to become lossy as a broad continuous range is compressed into a single class. Nevertheless, addressing the issue concerning the lower socioeconomic class is a straightforward task. Decreasing the value of $\alpha$ leads to an increase in the number of classes. Generally, there is no assurance adjusting the number of classes would enhance or diminish classification accuracy. Increasing the number of classes in the colorization problem leads to decreased prediction accuracy. It is crucial to calibrate the number of class levels for the a*b* color space to accurately represent the image's color characteristics using the updated $a^*b^*$ values.   

\subsubsection{Color Class Reduction Based on Practical Appearance}
The color values a* and b* range continuously from -128 to 127 in the a*b* color channel. Nevertheless, in practical terms, the range is generally confined to between -100 to 100. The continuous color channel a*b* in the range of $[-108, 108)$ is converted into a single plane consisting of 1296 color classes. This transformation is achieved by using the values $\alpha = 6$, $\beta = 108$, and $\delta = 36$ in Equation \ref{color_to_class}. A 2D grid of bins is created by arranging a single plane array of size $36\times36$. The horizontal axes represent the* color information, while the vertical axes represents the b* color information. A class value is assigned to each coordinate. The matrix representing the class is displayed in Fig. \ref{cont_dis}. The color class proposed for visualizing color is depicted in Fig. \ref{visualize1}. Fig. \ref{visualize1} demonstrates that the entire image seamlessly depicts various hues. The colors of the neighboring bins or blocks are nearly the same. The color transitions occur progressively, one block at a time. We choose bin size 6 as we find it optimal for the colorization task. High bin size generates less class points and requires less model parameter. However, it increases the maximum and average deviation from the actual pixel during color values retrieval. Low bin size decreases the deviation but generates much classes and requires much model parameter. The bin size approximation is shown in TABLE \ref{tab:binsize}, and relative comparison is shown in Fig. \ref{bin_size}.
\begin{figure}[!h]
	\centering
	\includegraphics[width=\linewidth]{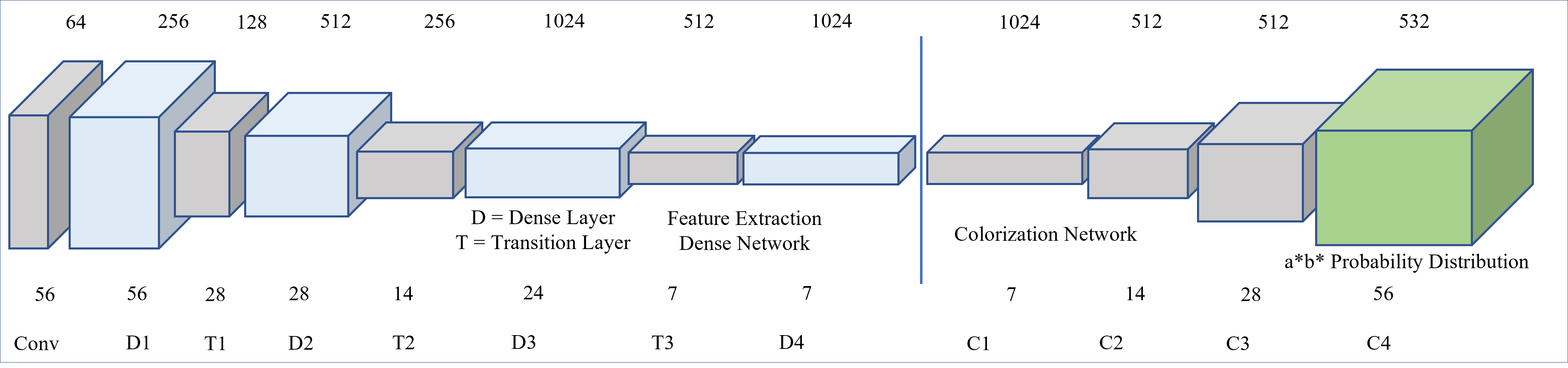}
	\caption{Network architecture of CCC++}
	\label{network}
	\vspace{-2mm}
\end{figure}
With 1296 color classes we experimented with the \textit{Place365 Validation} dataset \cite{Place} as a reference dataset. We see that some color classes face starvation in real-time images. The result is shown in Fig. \ref{rta}. In the \textit{Place365 Validation} dataset, there are 36,500 images in total. We randomly selected 35040 photos from the dataset and extracted classes from each of them. To decrease the number of class samples during the training phase, we resized the images to dimensions of $56\times56$. Consequently, there was a total of 109,885,440 (109.88 million) class samples, each belonging to one of the 1,296 class levels. Every pixel represents a color class sample with a distinct color intensity. We calculate class level histogram on total class samples. We consider a class level for training samples if the class level has minimum of 500 (0.000455\%) class samples out of 109885440 class samples. The class samples with a class level below 500 are assigned to their nearest-neighbor class levels using fixed centroid $k$-means clustering, as described in Equation \ref{kmeans}. The class in the final color bin, which is shown in Fig. \ref{visualize2}. 500 pixels of a color is very minimum compared to the total pixels of 35040 images because very few images have very few number of that color and the most of the images do not contain this color. In the end, we obtained 532 color classes out of a total of 1296 color classes that contain more than 500 pixels. Efficient optimization of the class is a significant concern for the colorization model. Reducing the number of classes in the model may improve its accuracy by reducing errors, but it may also result in some color visuals appearing outside of the designated range. One of our objectives is to maintain the presence of infrequently occurring color values in the projected distribution. Therefore, it is necessary to ensure that these rare color visions remain active during the training process. 
\begin{equation}
\label{kmeans}
 kmeans(C,\mu) = arg min\sum_{i=1}^k\sum_{C}||c- \mu_i||^2
\end{equation}
$C$ represents the input color class vector, $\mu$ represents the approved color classes for training, and $k$ represents the number of color classes equal to 532. The term $\mu$ can be defined as the stationary centroid value. The loop will occur only once, and the centroid value will remain unchanged. 
\subsubsection{Network Architecture}
Our model is constructed using an encoder-decoder architecture. Our feature extractor for the encoder part is DenseNet. The DenseNet is a sophisticated feature extractor that is well-suited for generating high-quality color values. Conventional CNN is employed for the decoder component. The network structure of our suggested approach is illustrated in Fig. \ref{network}. 

\begin{itemize}
\item {Feature Extraction:}
The connections between the layers of DenseNet are robust. It minimizes the gradient vanishing problem and has less semantic information loss during feature extraction than previous CNN models \cite{DenseNet}. The output from each subsequent layer is concatenated by the DenseNet. To adapt the model to grayscale input, we change the first convolutional layer. To build a $\frac{H}{32}\times\frac{W}{32}\times 1024$ feature representation from DenseNet, the final linear layer is discarded. These characteristics are utilized as input in the CNN-based colorization network. TABLE \ref{tab:densenet} displays the various DenseNet convolutional layers and outputs.
\end{itemize}
\begin {table}[H]
\caption {The model structure of the DenseNet} \label{tab:densenet}
\scriptsize 
\begin{center}
	\begin{tabular}{c c c c c } 
		\hline
		Layers & Output Size & \multicolumn{2}{c}{DenseNet-121} & Outputs \\  
		\hline
		Convolutional & $112\times112$ & \multicolumn{2}{c}{ }  & 64 \\ 
		\hline
		Pooling & $56\times56$ &  \multicolumn{2}{c}{ } & 64 \\
		\hline
		\multirow{ 2 }{*}{Dense Block 1}  & \multirow{ 2 }{*}{$56\times56$} & $1\times1$ conv & \multirow{ 2 }{*}{$\times6$} & \multirow{ 2 }{*}{256} \\
		& & $3\times3$ conv &  & \\
		\hline
		\multirow{ 2 }{*}{Transition 2}  & $56\times56$ & $1\times1$ conv & & \multirow{ 2 }{*}{128} \\
		& $28\times28$ & $3\times3$ conv & & \\
		\hline
		\multirow{ 2 }{*}{Dense Block 2}  &  \multirow{ 2 }{*}{$28\times28$} & $1\times1$ conv &\multirow{ 2 }{*}{$\times12$} &  \multirow{ 2 }{*}{512} \\
		&  & $3\times3$ conv & &  \\
		\hline
		\multirow{ 2 }{*}{Transition 2}  & $28\times28$ & $1\times1$ conv && \multirow{ 2 }{*}{256} \\
		& $14\times14$ & $3\times3$ conv & &  \\
		\hline
		\multirow{ 2 }{*}{Dense Block 3} & \multirow{ 2 }{*}{$14\times14$} & $1\times1$ conv &\multirow{ 2 }{*}{$\times24$}& \multirow{ 2 }{*}{1024} \\ 
		&   & $3\times3$ conv & &  \\
		\hline
		\multirow{ 2 }{*}{Transition 3} & $14\times14$ & $1\times1$ conv && \multirow{ 2 }{*}{512} \\ [1ex] 
		& $7\times7$ & $3\times3$ conv & & \\
		\hline
		\multirow{ 2 }{*}{Dense Block 4} & \multirow{ 2 }{*}{$7\times7$} & $1\times1$ conv &\multirow{ 2 }{*}{$\times16$}& \multirow{ 2 }{*}{1024} \\ 
		&   & $3\times3$ conv & &  \\
		\hline
	\end{tabular}
\end{center}
\vspace{-2mm}
\end {table}

\begin{itemize}
    \item {Colorization Network}
After obtaining a feature representation of size $\frac{H}{32}\times\frac{W}{32}\times1024$, the network leverages many convolutional and up-sampling layers. We utilize the fundamental nearest-neighbor method for up-sampling. The network outputs a tensor of dimensions $H\times W\times2$ with elements denoted by a*b*. TABLE \ref{tab:cnn} presents the convolutional layers and their corresponding outcomes. 
\end{itemize}

\begin{figure}[!b]
	\centering
	\includegraphics[width=\linewidth, height = 4cm]{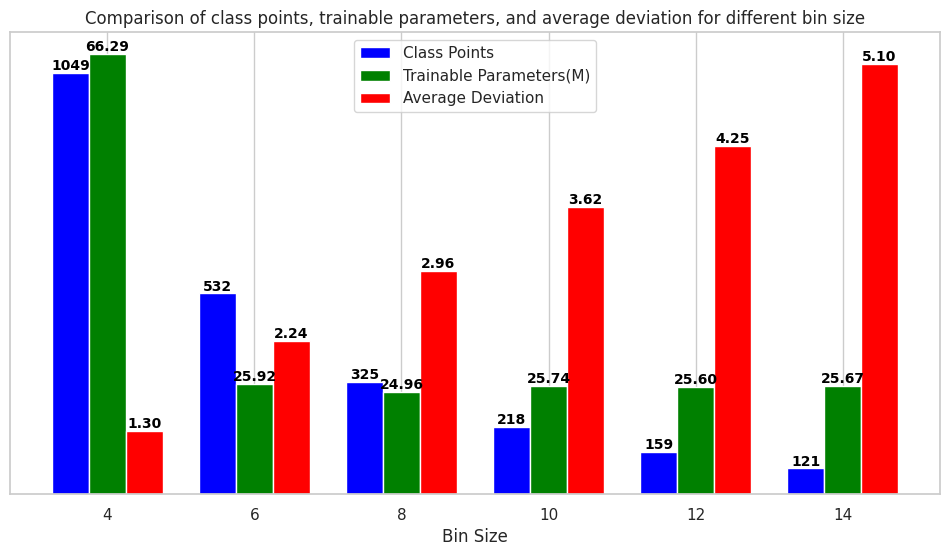}
	\caption{Comparison of class points, trainable parameters, and average deviation for different bin size}
	\label{bin_size}
	\vspace{-2mm}
\end{figure}

\subsubsection{Loss Calculation}
Colorization is often regarded as a regression problem due to the continuous nature of color values. However, we convert the continuous color values into the discrete color classes outlined in Section \ref{Continuous Color Range to Discrete Color Classes}. Thus, we approach the problem as a classification problem and employ cross-entropy loss instead of mean squared error (MSE) or other regression loss functions. The loss function is depicted in Equation \ref{crossentropy}. 

\begin{equation}
	\label{crossentropy}
	Loss_{CE} =- \sum_{H,W}\mathcal{W}_c\sum_{\mathcal{C}} \mathcal{K} . log(\overline{\mathcal{K}}).
\end{equation}

Here, $H$ and $W$ represent the height and width of the output $\kappa$ distribution. $\kappa$ represents the actual color category, while $\overline{\kappa}$ represents the predicted color category. The $\omega_c$ represents the vector of weights for color classes. The symbol $\omega_c$ is defined in Equation \ref{normal_weights}. 
\begin{equation}
	\label{normal_weights}
	\mathcal{W}_c =  \Big(\frac{1}{n_\mathcal{C}}\Big), \forall_c \in \mathcal{C}
\end{equation}
where $N_c$ is the number of color classes.

\begin {table}[H]
\caption {The model structure of the proposed Colorization Network} \label{tab:cnn} 
\begin{center}
	\begin{tabular}{c c c c c} 
		\hline
		Layers & Output Size & Kernel & Stride & Outputs \\  
		\hline
		Conv-1 & $7\times7$ & $3\times3$ & $1\times1$  & 1024 \\ 
		\hline
		Upsample(scale factor=2) & $14\times14$ & - & -  & 1024 \\ 
		\hline
		Conv-2 & $14\times14$ & $3\times3$ & $1\times1$  & 512 \\ 
		\hline
		Upsample(scale factor=2) & $28\times28$ & - & -  & 512 \\ 
		\hline
		Conv-3 & $28\times28$ & $3\times3$ & $1\times1$  & 512 \\ 
		\hline
		Upsample(scale factor=2) & $56\times56$ & - & -  & 512 \\ 
		\hline
		Conv-4 & $56\times56$ & $3\times3$ & $1\times1$  & 532 \\ 
		\hline
	\end{tabular}
\end{center}
\vspace{-2mm}
\end {table}
\subsubsection{Class Confusion Based Weights Adjustment}
In actual photographs, the distribution of color classes is not uniform. The prevalence of grayish visual color classes is higher than luminous color classes, primarily because of the extensive background areas. The categorical cross-entropy loss assigns a weight of $\frac{1}{N}$ to each actual class when calculating the loss, as seen in Equation \ref{normal_weights}. Due to the relatively low frequency of bright color classes, the gradients steadily diminish with each iteration. To preserve the infrequently occurring color categories, we amplify the weights of these categories to a greater extent than those of the predominantly occurring color categories.
Nevertheless, this procedure amplifies the overall detriment on a worldwide scale. Making trade-offs while determining the weights is necessary to achieve both realistic colors and minimize loss. We have introduced a novel formula to balance the weights, as shown in Equation \ref{weights_trade_off}. 
\begin{equation}
	\label{weights_trade_off}
	\mathcal{W}_{\text{new}} = \left( \frac{\sum_{i=1}^{\mathcal{H}} \sum_{j=1}^{\mathcal{W}} \sum_{k=1}^{\mathcal{B}} 1}{N^{adj}_c + \frac{\max(N_c)}{\Psi}} \right), \quad \forall c \in \mathcal{C}
\end{equation}
where, 
\begin{equation*}
	\label{bin_adj}
	N^{adj}_c = \begin{cases}
		\Psi & \text{if } N_c < \Psi \\
		N_c & \text{otherwise}
	\end{cases}
	\quad \forall c \in \mathcal{C}
\end{equation*}
where,
\begin{equation*}
	\label{Psi}
	\Psi = \Bigg(\frac{\sum_{i=1}^{\mathcal{H}} \sum_{j=1}^{\mathcal{W}} \sum_{k=1}^{\mathcal{B}} 1}{n_\mathcal{C}}\Bigg) \cdot \mathcal{P}(\%)
\end{equation*}
where $\mathcal{H}$, $\mathcal{W}$, and $\mathcal{B}$ represent the height, width, and batch size of the true class, respectively. $\mathcal{C}$ represents the set of color classes. ${\max}(N_c)$ refers to a class's highest value of appearance. $N_c$ represents the appearance value of a specific class $c$. $\mathcal{W}_{new}$ is the matrix of weights for the current batch. $N^{adj}_c$ represents the adjusted appearance value of a class. The variables $c$ and $\Psi$ represent the threshold value used to change the appearance value of the classes. The variable $n_c$ represents the total number of classes in the optimized class set, which is set to 532 according to a rule. The variable $\mathcal{P}$ represents the percentage value used to determine the threshold value $\Psi$. 

\begin{figure}[!ht]
	\centering
	\vspace{-1mm}
	\begin{subfigure}
		\centering
		{\rotatebox{90}{\scriptsize Gray}} 
		\includegraphics[width=.106\linewidth,height = 1cm]{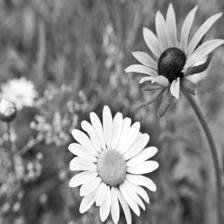}
		\includegraphics[width=.106\linewidth,height = 1cm]{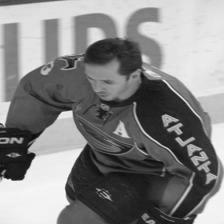}
		\includegraphics[width=.106\linewidth,height = 1cm]{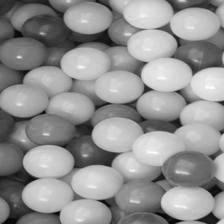}
		\includegraphics[width=.106\linewidth,height = 1cm]{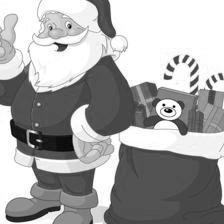}
		\includegraphics[width=.106\linewidth,height = 1cm]{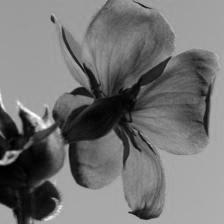}
		\includegraphics[width=.106\linewidth,height = 1cm]{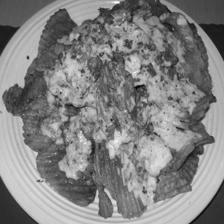}
		\includegraphics[width=.106\linewidth,height = 1cm]{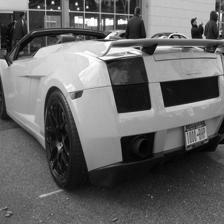}
		\includegraphics[width=.106\linewidth,height = 1cm]{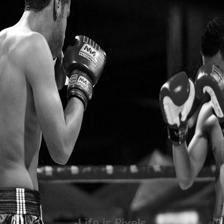}
	\end{subfigure}
	\vspace{-1mm}
	\begin{subfigure}
		\centering
		\rotatebox{90}{\tiny Deold.\cite{Deoldify}}
		\includegraphics[width=.106\linewidth,height = 1cm]{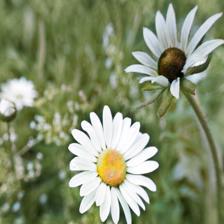}
		\includegraphics[width=.106\linewidth,height = 1cm]{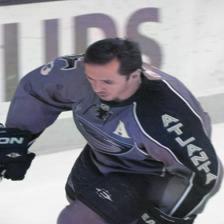}
		\includegraphics[width=.106\linewidth,height = 1cm]{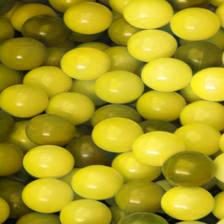}
		\includegraphics[width=.106\linewidth,height = 1cm]{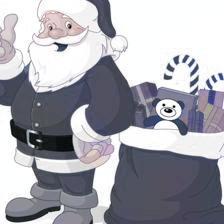}
		\includegraphics[width=.106\linewidth,height = 1cm]{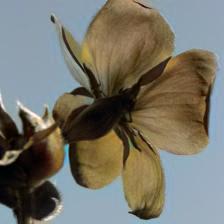}
		\includegraphics[width=.106\linewidth,height = 1cm]{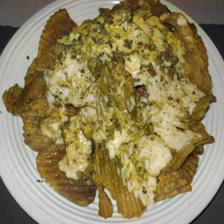}
		\includegraphics[width=.106\linewidth,height = 1cm]{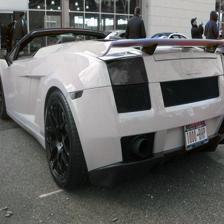}
		\includegraphics[width=.106\linewidth,height = 1cm]{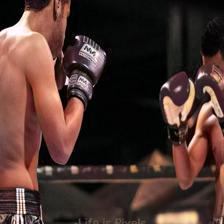}
	\end{subfigure}
	\vspace{-1mm}
	\begin{subfigure}
		\centering
		{\rotatebox{90}{\tiny Iizuka\cite{Iizuka}}} 
		\includegraphics[width=.106\linewidth,height = 1cm]{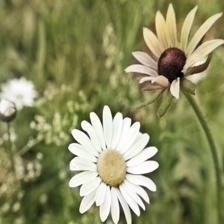}
		\includegraphics[width=.106\linewidth,height = 1cm]{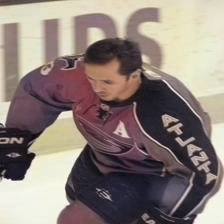}
		\includegraphics[width=.106\linewidth,height = 1cm]{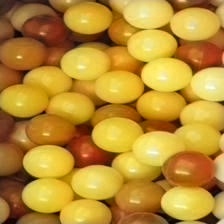}
		\includegraphics[width=.106\linewidth,height = 1cm]{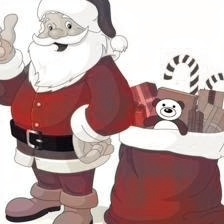}
		\includegraphics[width=.106\linewidth,height = 1cm]{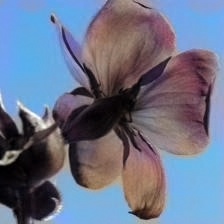}
		\includegraphics[width=.106\linewidth,height = 1cm]{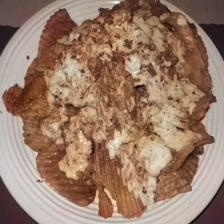}
		\includegraphics[width=.106\linewidth,height = 1cm]{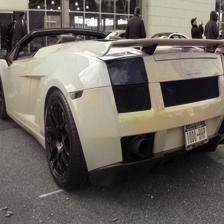}
		\includegraphics[width=.106\linewidth,height = 1cm]{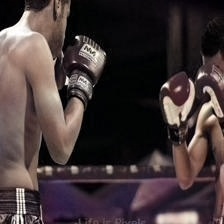}
	\end{subfigure}
	\vspace{-1mm}
	\begin{subfigure}
		\centering
		{\rotatebox{90}{\scriptsize Larss.\cite{Larsson}}} 
		\includegraphics[width=.106\linewidth,height = 1cm]{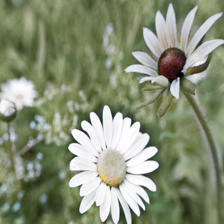}
		\includegraphics[width=.106\linewidth,height = 1cm]{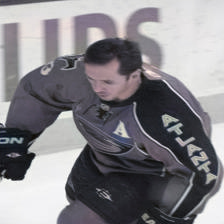}
		\includegraphics[width=.106\linewidth,height = 1cm]{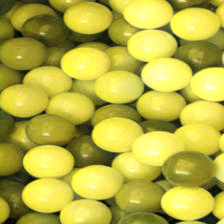}
		\includegraphics[width=.106\linewidth,height = 1cm]{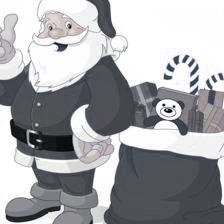}
		\includegraphics[width=.106\linewidth,height = 1cm]{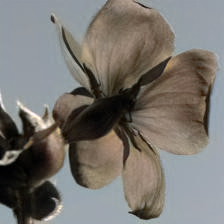}
		\includegraphics[width=.106\linewidth,height = 1cm]{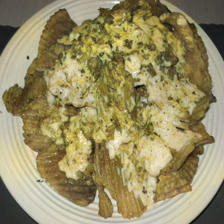}
		\includegraphics[width=.106\linewidth,height = 1cm]{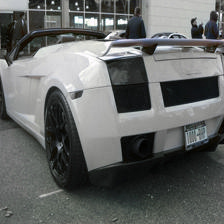}
		\includegraphics[width=.106\linewidth,height = 1cm]{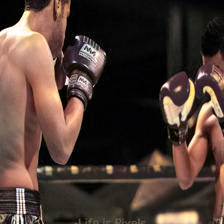}
	\end{subfigure}
	\vspace{-1mm}
	\begin{subfigure}
		\centering
		{\rotatebox{90}{\scriptsize CIC\cite{Zhang_eccv}}} 
		\includegraphics[width=.106\linewidth,height = 1cm]{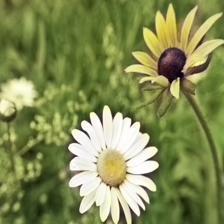}
		\includegraphics[width=.106\linewidth,height = 1cm]{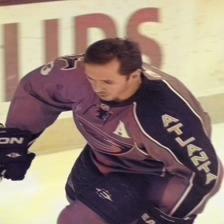}
		\includegraphics[width=.106\linewidth,height = 1cm]{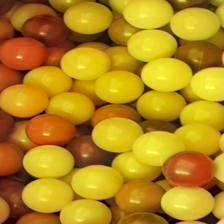}
		\includegraphics[width=.106\linewidth,height = 1cm]{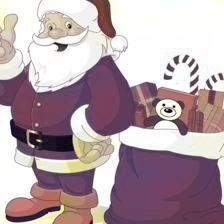}
		\includegraphics[width=.106\linewidth,height = 1cm]{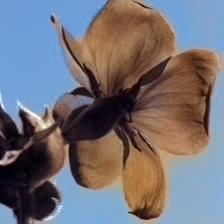}
		\includegraphics[width=.106\linewidth,height = 1cm]{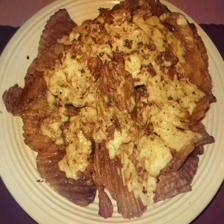}
		\includegraphics[width=.106\linewidth,height = 1cm]{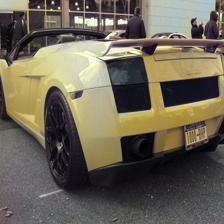}
		\includegraphics[width=.106\linewidth,height = 1cm]{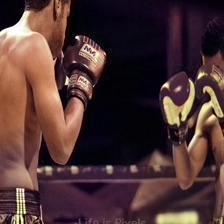}
	\end{subfigure}
	\vspace{-1mm}
	\begin{subfigure}
		\centering
		{\rotatebox{90}{\tiny Zhang\cite{Zhang_tog}}} 
		\includegraphics[width=.106\linewidth,height = 1cm]{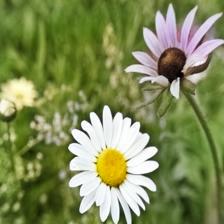}
		\includegraphics[width=.106\linewidth,height = 1cm]{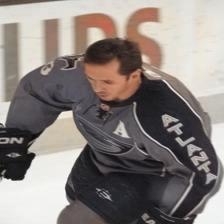}
		\includegraphics[width=.106\linewidth,height = 1cm]{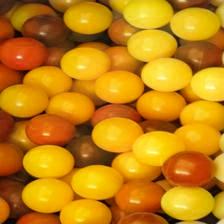}
		\includegraphics[width=.106\linewidth,height = 1cm]{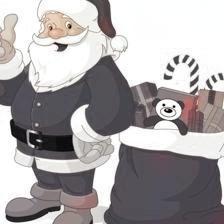}
		\includegraphics[width=.106\linewidth,height = 1cm]{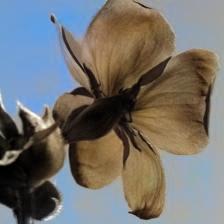}
		\includegraphics[width=.106\linewidth,height = 1cm]{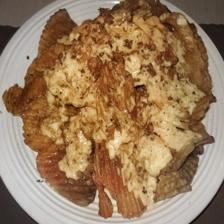}
		\includegraphics[width=.106\linewidth,height = 1cm]{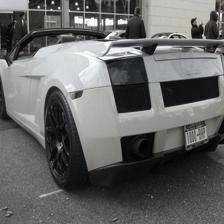}
		\includegraphics[width=.106\linewidth,height = 1cm]{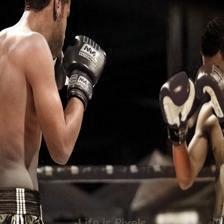}
	\end{subfigure}
	\vspace{-1mm}
	\begin{subfigure}
		\centering
		{\rotatebox{90}{\scriptsize Su\cite{Su}}} 
		\includegraphics[width=.106\linewidth,height = 1cm]{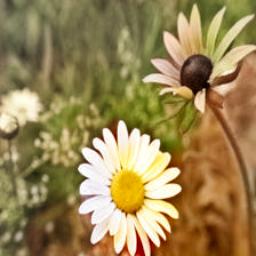}
		\includegraphics[width=.106\linewidth,height = 1cm]{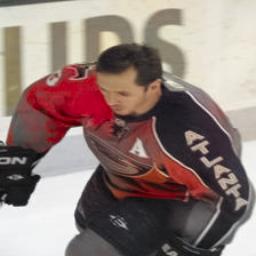}
		\includegraphics[width=.106\linewidth,height = 1cm]{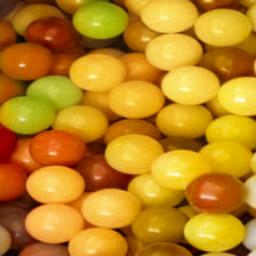}
		\includegraphics[width=.106\linewidth,height = 1cm]{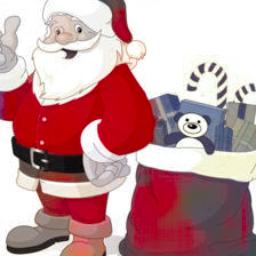}
		\includegraphics[width=.106\linewidth,height = 1cm]{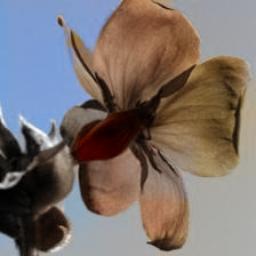}
		\includegraphics[width=.106\linewidth,height = 1cm]{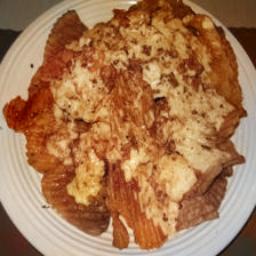}
		\includegraphics[width=.106\linewidth,height = 1cm]{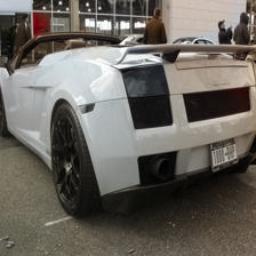}
		\includegraphics[width=.106\linewidth,height = 1cm]{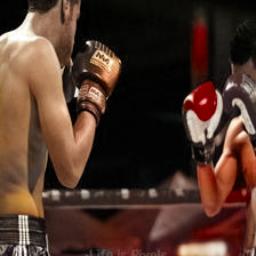}
	\end{subfigure}
	\vspace{-1mm}
	\begin{subfigure}
		\centering
		{\rotatebox{90}{\scriptsize Gain\cite{Gain2}}} 
		\includegraphics[width=.106\linewidth,height = 1cm]{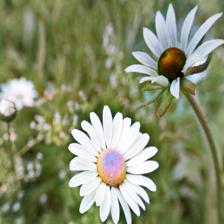}
		\includegraphics[width=.106\linewidth,height = 1cm]{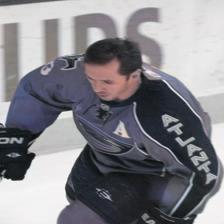}
		\includegraphics[width=.106\linewidth,height = 1cm]{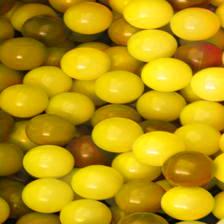}
		\includegraphics[width=.106\linewidth,height = 1cm]{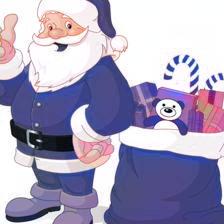}
		\includegraphics[width=.106\linewidth,height = 1cm]{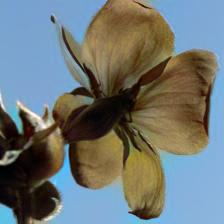}
		\includegraphics[width=.106\linewidth,height = 1cm]{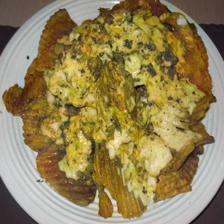}
		\includegraphics[width=.106\linewidth,height = 1cm]{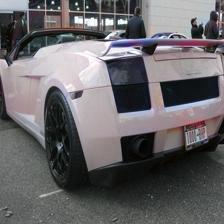}
		\includegraphics[width=.106\linewidth,height = 1cm]{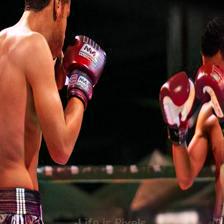}
	\end{subfigure}
	\vspace{-1mm}
	\begin{subfigure}
		\centering
		{\rotatebox{90}{\scriptsize DD\cite{DD}}} 
		\includegraphics[width=.106\linewidth,height = 1cm]{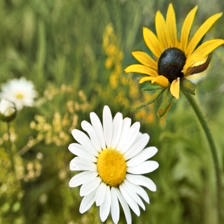}
		\includegraphics[width=.106\linewidth,height = 1cm]{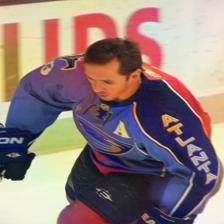}
		\includegraphics[width=.106\linewidth,height = 1cm]{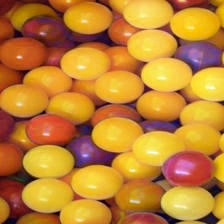}
		\includegraphics[width=.106\linewidth,height = 1cm]{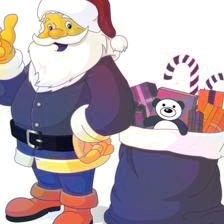}
		\includegraphics[width=.106\linewidth,height = 1cm]{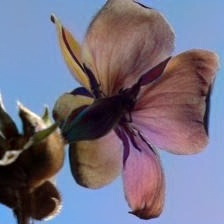}
		\includegraphics[width=.106\linewidth,height = 1cm]{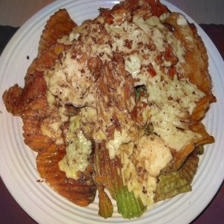}
		\includegraphics[width=.106\linewidth,height = 1cm]{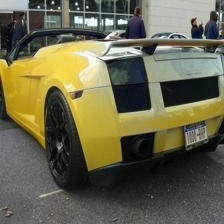}
		\includegraphics[width=.106\linewidth,height = 1cm]{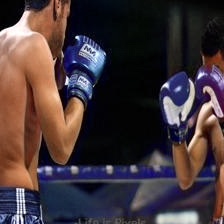}
	\end{subfigure}
	\vspace{-1mm}
	\begin{subfigure}
		\centering
		{\rotatebox{90}{\scriptsize CCC++}} 
		\includegraphics[width=.106\linewidth,height = 1cm]{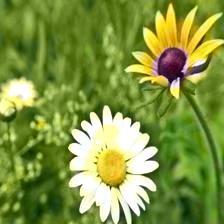}
		\includegraphics[width=.106\linewidth,height = 1cm]{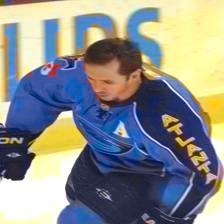}
		\includegraphics[width=.106\linewidth,height = 1cm]{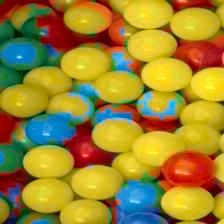}
		\includegraphics[width=.106\linewidth,height = 1cm]{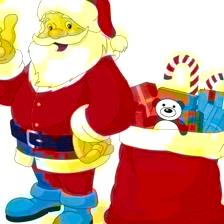}
		\includegraphics[width=.106\linewidth,height = 1cm]{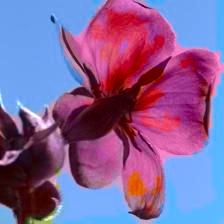}
		\includegraphics[width=.106\linewidth,height = 1cm]{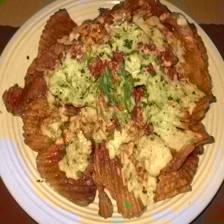}
		\includegraphics[width=.106\linewidth,height = 1cm]{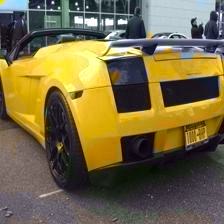}
		\includegraphics[width=.106\linewidth,height = 1cm]{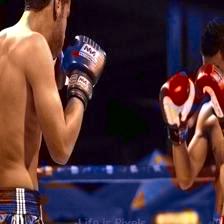}
	\end{subfigure}
	\vspace{-1mm}\hspace{0.5mm}
	\begin{subfigure}
		\centering
		{\rotatebox{90}{\scriptsize G. Truth}} 
		\includegraphics[width=.106\linewidth,height = 1cm]{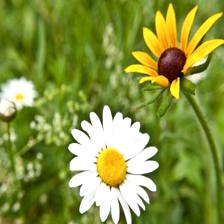}
		\includegraphics[width=.106\linewidth,height = 1cm]{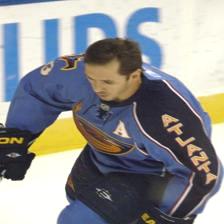}
		\includegraphics[width=.106\linewidth,height = 1cm]{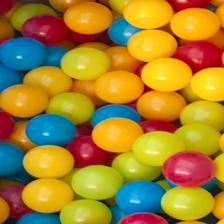}
		\includegraphics[width=.106\linewidth,height = 1cm]{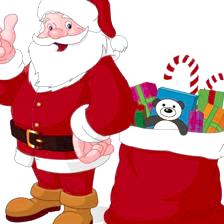}
		\includegraphics[width=.106\linewidth,height = 1cm]{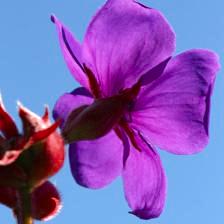}
		\includegraphics[width=.106\linewidth,height = 1cm]{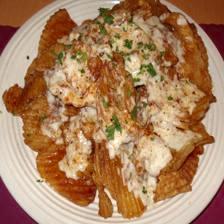}
		\includegraphics[width=.106\linewidth,height = 1cm]{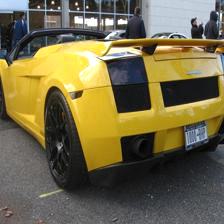}
		\includegraphics[width=.106\linewidth,height = 1cm]{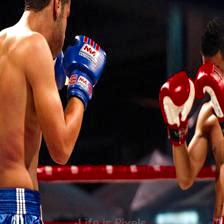}
	\end{subfigure}
	\caption{Some results of our proposed method compared to other state-of-the-art methods. }
	\label{fig:comparison}
	\vspace{-8mm}
\end{figure}
Initially, we regularize weights by dividing the total count of each class in a batch by its individual appearance value. This ensures that the weight of the most frequently appeared class is set to a minimum while proportionally upscaling the weights of others. However, this approach leads to a significant increase in the weight of classes appearing very infrequently. When a particular class appear 0 or tends to 0 then the weight of that class will be extremely high that makes the learning process imbalance again. To strike a balance, we introduce a trade-off mechanism. Firstly, we adjust appearance value($N^{adj}_c$) of class $c$ using a threshold factor $\Psi$. By this we prevent every class for being holding extreme high weight. However, some minor classes appear very few times compared to some major classes. Therefore, those major classes go under the domination of some minor classes. To address the problem, we propose an additional term ($\frac{\max(N_c)}{\Psi}$), to be incorporated into the $N^{adj}_c$ term. When the difference between the maximum appeared class and minimum appeared $(\Psi)$ class is very high, the ratio will also be high. Therefore, high value will be added to the adjusted class value. This reduce the weights high appeared classes little but low appeared classes much. This seems little contradictory with our initial target as we are trying to establish the presence of minor class in the generated image and therefore, increasing the weights of the minor classes. However, unrestrained increase of weights of minor class can bias the learning towards minor classes. Our target is to ensure the learning unbiased. Therefore we need a perfect trade-off between the major and minor classes. To address this, we introduce a new hyperparameter $\mathcal{P}$ which denotes how much percentage we will take for determining the value of minimum appeared class. This additional term assists in managing the impact of infrequent class instances, resulting in a more refined and equitable method for determining class weights. Consequently, the loss function has been altered, as depicted in the Equation \ref{crossentropy_new}.
\begin{equation}
	\vspace{-4mm}
	\label{crossentropy_new}
	Loss_{\text{CBCE}} =- \sum_{H,W}\mathcal{W}_{new}\sum_{C} \mathcal{K} . log(\overline{\mathcal{K}}).
	\vspace{-1mm}
\end{equation}

\subsubsection{Class Probabilities Estimation for Loss Calculation}
The network outputs $H \times W \times C \times batch $ tensor. We extract $H \times W \times C \times batch$ softmax probability distribution of the class representation to calculate the loss for backpropagation using the Equation \ref{softmax}.
\begin{equation}
\label{softmax}
\sigma(\overline{\kappa})_{i} = \frac{e^{\kappa_{i}}}{\sum_{j=1}^k{e{\kappa_{i}}}}
\end{equation}
\subsubsection{Class Selection from Probability Distribution for Image Reconstruction}
The network outputs a tensor distribution $P$ of dimension $H \times W \times C \times batch $. We extract the class distribution $\mathcal{C}$ of dimension $H \times W \times 1 \times batch$ for a*b* color channel construction using the Equation \ref{argmax}.
\begin{equation}
	\label{argmax}
	\mathcal{C} \in \mathbb{R}^{H \times W}, \quad \mathcal{C}_{ij} = \underset{c}{\operatorname{argmax}}(P_{ijk})
	\vspace{-1mm}
\end{equation}

Here, $\quad \mathcal{C}_{ij}$ represents the element at the \textit{i-th} row and \textit{j-th} column of matrix $\mathcal{C}$, and it is the index of the maximum probability along the third dimension of matrix $P$ at the corresponding position.

 \begin{figure}[!h]
    \centering
    \vspace{-1mm}
    \begin{subfigure}
        \centering
        \includegraphics[width=.113\linewidth,height = 1.0cm]{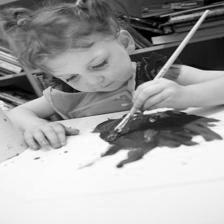}
        \includegraphics[width=.113\linewidth,height = 1.0cm]{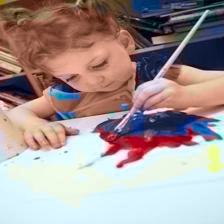}
        \includegraphics[width=.113\linewidth,height = 1.0cm]{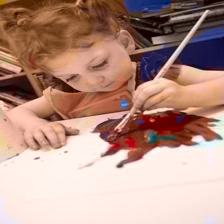}
        \includegraphics[width=.113\linewidth,height = 1.0cm]{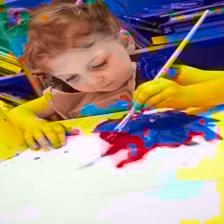}
        \includegraphics[width=.113\linewidth,height = 1.0cm]{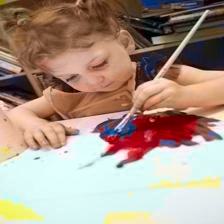}
        \includegraphics[width=.113\linewidth,height = 1.0cm]{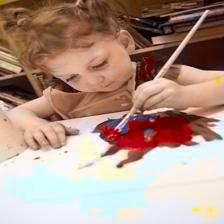}
        \includegraphics[width=.113\linewidth,height = 1.0cm]{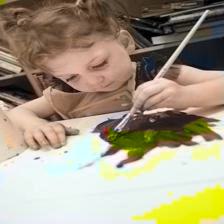}
        \includegraphics[width=.113\linewidth,height = 1.0cm]{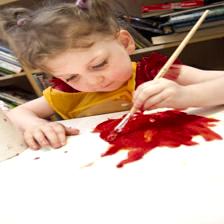}
    \end{subfigure}
    \vspace{-1mm}
    \begin{subfigure}
        \centering
        \includegraphics[width=.113\linewidth,height = 1.0cm]{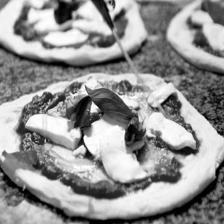}
        \includegraphics[width=.113\linewidth,height = 1.0cm]{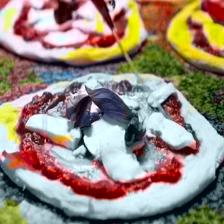}
        \includegraphics[width=.113\linewidth,height = 1.0cm]{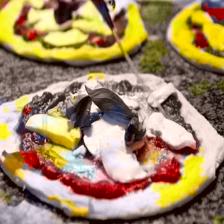}
        \includegraphics[width=.113\linewidth,height = 1.0cm]{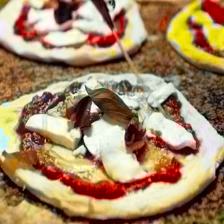}
        \includegraphics[width=.113\linewidth,height = 1.0cm]{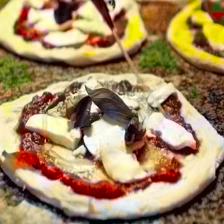}
        \includegraphics[width=.113\linewidth,height = 1.0cm]{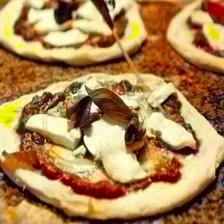}
        \includegraphics[width=.113\linewidth,height = 1.0cm]{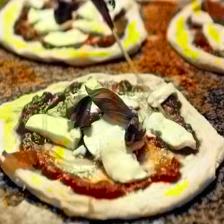}
        \includegraphics[width=.113\linewidth,height = 1.0cm]{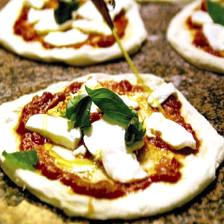}
    \end{subfigure}
    \vspace{-1mm}
    \begin{subfigure}
        \centering
        \includegraphics[width=.113\linewidth,height = 1.0cm]{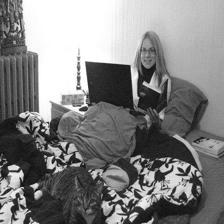}
        \includegraphics[width=.113\linewidth,height = 1.0cm]{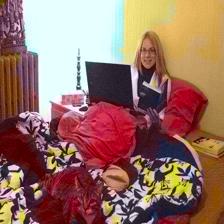}
        \includegraphics[width=.113\linewidth,height = 1.0cm]{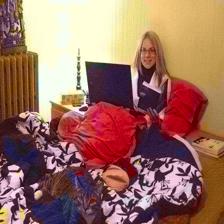}
        \includegraphics[width=.113\linewidth,height = 1.0cm]{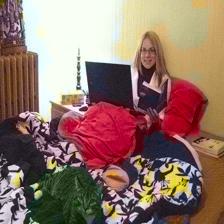}
        \includegraphics[width=.113\linewidth,height = 1.0cm]{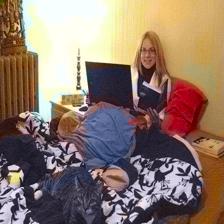}
        \includegraphics[width=.113\linewidth,height = 1.0cm]{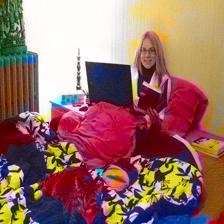}
        \includegraphics[width=.113\linewidth,height = 1.0cm]{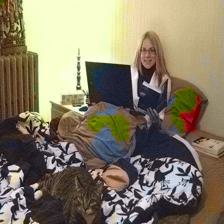}
        \includegraphics[width=.113\linewidth,height = 1.0cm]{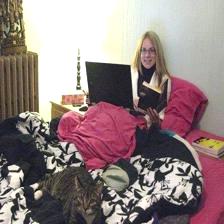}
    \end{subfigure}
    \vspace{-1mm}
    \begin{subfigure}
        \centering
        \includegraphics[width=.113\linewidth,height = 1.0cm]{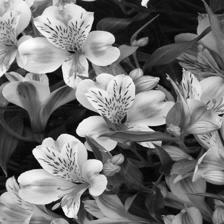}
        \includegraphics[width=.113\linewidth,height = 1.0cm]{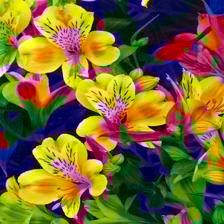}
        \includegraphics[width=.113\linewidth,height = 1.0cm]{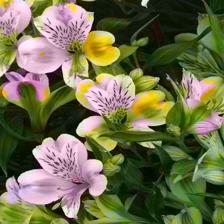}
        \includegraphics[width=.113\linewidth,height = 1.0cm]{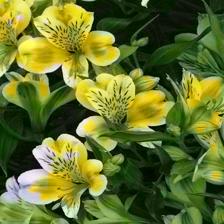}
        \includegraphics[width=.113\linewidth,height = 1.0cm]{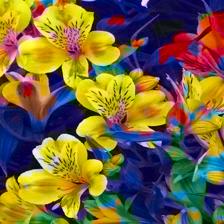}
        \includegraphics[width=.113\linewidth,height = 1.0cm]{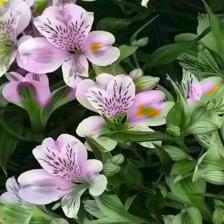}
        \includegraphics[width=.113\linewidth,height = 1.0cm]{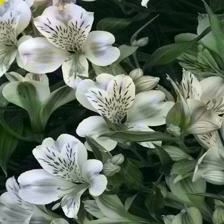}
        \includegraphics[width=.113\linewidth,height = 1.0cm]{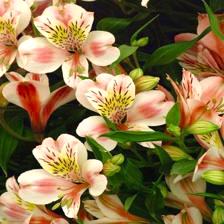}
    \end{subfigure}
    \vspace{-1mm}
    \begin{subfigure}
        \centering
        \includegraphics[width=.113\linewidth,height = 1.0cm]{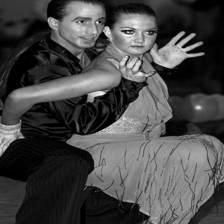}
        \includegraphics[width=.113\linewidth,height = 1.0cm]{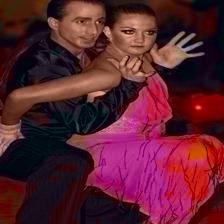}
        \includegraphics[width=.113\linewidth,height = 1.0cm]{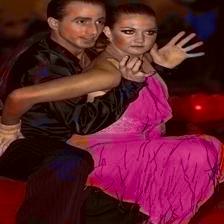}
        \includegraphics[width=.113\linewidth,height = 1.0cm]{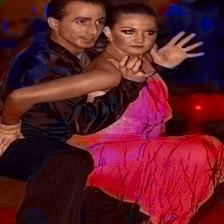}
        \includegraphics[width=.113\linewidth,height = 1.0cm]{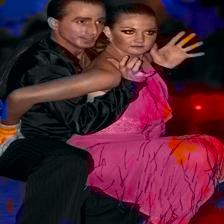}
        \includegraphics[width=.113\linewidth,height = 1.0cm]{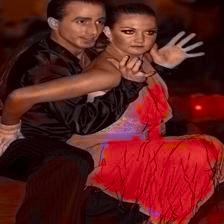}
        \includegraphics[width=.113\linewidth,height = 1.0cm]{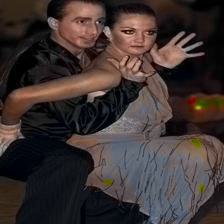}
        \includegraphics[width=.113\linewidth,height = 1.0cm]{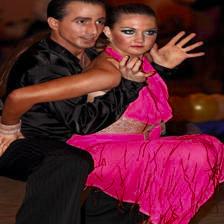}
    \end{subfigure}
    \vspace{-1mm}
    \begin{subfigure}
        \centering
        \includegraphics[width=.113\linewidth,height = 1.0cm]{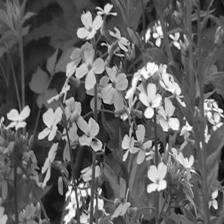}
        \includegraphics[width=.113\linewidth,height = 1.0cm]{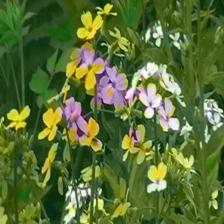}
        \includegraphics[width=.113\linewidth,height = 1.0cm]{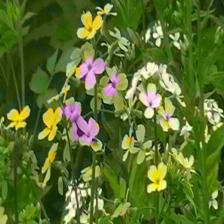}
        \includegraphics[width=.113\linewidth,height = 1.0cm]{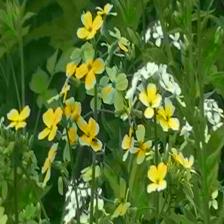}
        \includegraphics[width=.113\linewidth,height = 1.0cm]{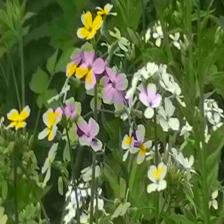}
        \includegraphics[width=.113\linewidth,height = 1.0cm]{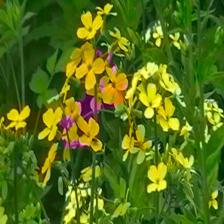}
        \includegraphics[width=.113\linewidth,height = 1.0cm]{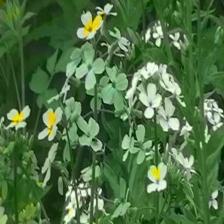}
        \includegraphics[width=.113\linewidth,height = 1.0cm]{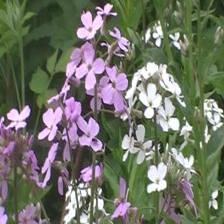}
    \end{subfigure}
    \vspace{-3mm}
    \begin{subfigure}
        \centering
   \caption*{ Gray \hspace{2mm} bs=4 \hspace{0.5mm} bs=6 \hspace{1mm} bs=8 \hspace{1mm} bs=10 \hspace{1mm} bs=12 \hspace{0.5mm} bs=14 \hspace{0.5mm} G. Tru.} 
    \end{subfigure}
    \caption{Results on different bin size }
    \label{fig:bin_size}
    \vspace{-6mm}
\end{figure}
\subsection{SAM Empowered Object-Selective Color Harmonization}
\label{sec:sam}
\begin{figure}[!t]
\centering
\includegraphics[width=\linewidth]{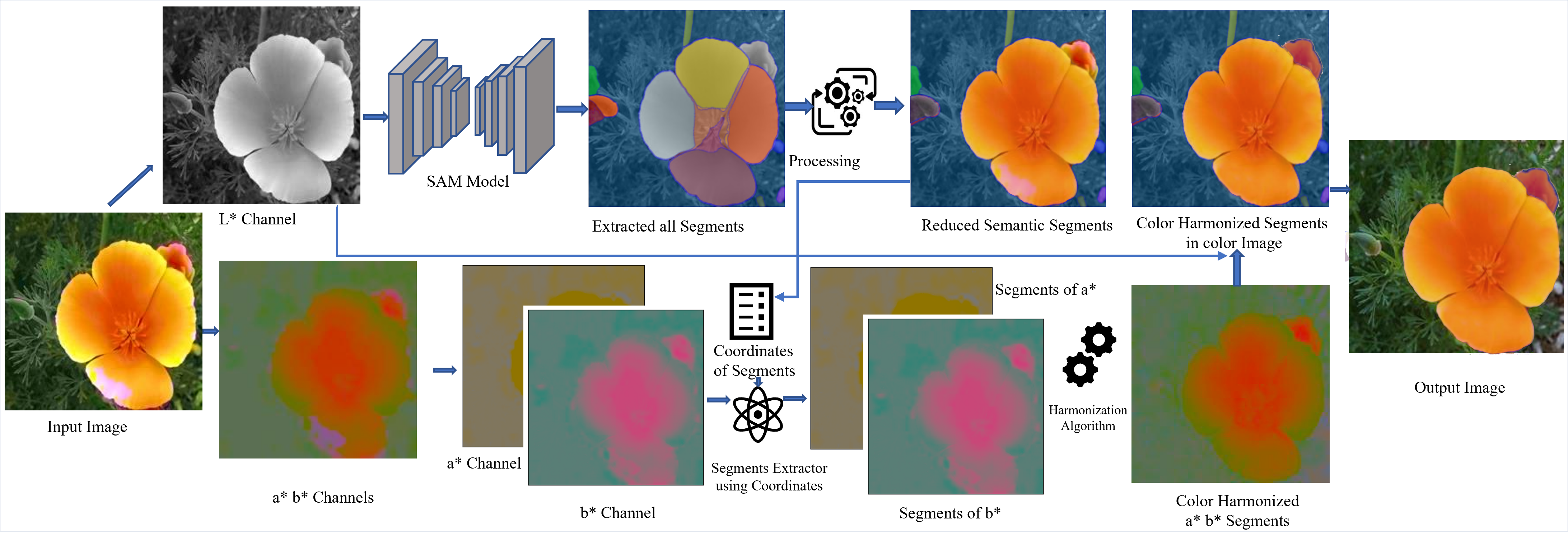}
\caption{System model for SAM empowered object selective color harmonization.}
\label{sam_model}
\vspace{-2mm}
\end{figure}

To address the noise sometimes observed at the edges of objects when regularizing the minor class, we propose SAM-empowered object-selective color harmonization, as illustrated in Fig. \ref{sam_model}. The SAM is a cutting-edge image segmentation model developed by Meta AI Research. It features zero-shot generalization, allowing it to segment various objects and scenes without additional training. SAM excels in interactive segmentation, enabling users to refine results with points, boxes, or text prompts. Known for its high precision, SAM can extract detailed masks for multiple objects in complex images, making it highly versatile and effective for various applications \cite{SAM, raha_arxiv}. Our proposed algorithm is detailed in Algorithm \ref{alg:segmentation}. This approach ensures more polished edges by leveraging SAM's advanced segmentation capabilities, leading to improved overall color harmonization.

\subsection{Chromatic Diversity}
Conventional color image assessment metrics often concentrate on measuring discrepancies between generated and ground truth images. However, such metrics may fall short in capturing scenarios where introducing distinctive color accents to specific regions enhances visual aesthetics. Conversely, even a modest area within an image, if strategically aligned with the color scheme of a more dominant counterpart or rendered colorless, can significantly contribute to the conventional measurement criteria. Therefore a small peak ratio (major area only) can give good performance in that criteria and minor areas are overlooked. To address the above mentioned problem, We introduce a new color image evaluation statistic called Chromatic Number Ratio (CNR). The CNR measures the abundance of color categories in the generated images about the ground truth photos. It thoroughly assesses the range of colors in the produced images, improving our comprehension of color variety. The metric is displayed in Equation \ref{color_diversity}. 
\begin{equation}
	\label{color_diversity}
	CNR =\frac{\sum\limits_{i=0}^{m-1} \sum\limits_{j=0}^{n-1} \left(1 - \sum\limits_{k=0}^{i-1} \sum\limits_{l=0}^{n-1} [\mathcal{P}_{i,j} = \mathcal{P}_{k,l}]\right)}{\sum\limits_{i=0}^{m-1} \sum\limits_{j=0}^{n-1} \left(1 - \sum\limits_{k=0}^{i-1} \sum\limits_{l=0}^{n-1} [\mathcal{G}_{i,j} = \mathcal{G}_{k,l}]\right)} 
	\vspace{-1mm}
\end{equation}
where the variables $\mathcal{P}_{i,j}$ and $\mathcal{G}_{i,j}$ represent the color class value at the specific position of row $i$ and column $j$ in the color image. The variables $\mathcal{P}$ and $\mathcal{G}$ represent the predicted image and ground truth image, respectively. The image's dimensions in the color class space are denoted by $m$ and $n$. The outer summation iterates over all rows ($i$) and columns ($j$) of the image in the color class space. The inner summation evaluates the uniqueness of each pixel ($\mathcal{P}_{i,j}$ / $\mathcal{G}_{i,j}$) by comparing it with all preceding pixels in the image. $[\mathcal{P}_{i,j} = \mathcal{P}_{k,l}]$ and [$\mathcal{G}_{i,j}$ = $\mathcal{G}_{k,l}$] is the indicator function that evaluates to 1 when the condition is true (when pixel values are equal) and 0 when it is false.

\subsection{Color Class Activation Ratio}
Within the Chromatic Number Ratio (CNR) context, although it accurately measures the ratio of active color classes between generated images and ground truth images, determining the exact number of activated color classes within the whole set of discretized color classes is challenging. To tackle the problems, we provide a new, innovative method for assessing color images called the Color Class Activation Ratio (CCAR). The CCAR measures the proportion of activated color categories in the colorized images. It thoroughly assesses the range of colors in the produced images, improving our comprehension of color diversity. The metric is displayed in Equation \ref{color_activation}. 
\begin{equation}
	\label{color_activation}
	CCAR =\frac{\sum\limits_{i=0}^{m-1} \sum\limits_{j=0}^{n-1} \left(1 - \sum\limits_{k=0}^{i-1} \sum\limits_{l=0}^{n-1} [\mathcal{P}_{i,j} = \mathcal{P}_{k,l}]\right)}{N_{Class}} 
\end{equation}
where, the color class value at row $i$ and column $j$ of the created color image $\mathcal{P}$ is denoted as $\mathcal{P}_{i,j}$. The picture's dimensions in the color class space are represented by $m$ and $n$. The outer summation iterates over all rows ($i$) and columns ($j$) of the image in the color class space. The inner summation evaluates the uniqueness of each pixel ($\mathcal{P}_{i,j}$) by comparing it with all preceding pixels in the image. The equality of $\mathcal{P}_{i,j}$ and $\mathcal{P}_{k,l}$ is denoted as $[\mathcal{P}_{i,j} = \mathcal{P}_{k,l}]$, which is an indicator function that evaluates to 1 when the condition is true (when pixel values are equal) and 0 when it is false. 

\subsection{True Activation Ratio}
In the evaluation metrics CNR (Chromatic Number Ratio) and CCAR (Color Class Alignment Ratio), we effectively quantify the class activation ratio between generated and ground truth images, as well as between generated and total discretized color classes, respectively. However, determining the specific count of color classes that align precisely with ground truth images remains a challenging task. While our primary goal is to enhance the diversity and plausibility of color classes, including those belonging to minor counterparts, we also emphasize the accurate alignment of color classes, particularly focusing on preserving the presence of minor classes. For this, we propose a new method for assessing color images called True Activation Ratio (TAR). The TAR measures the proportion of color classes that are accurately activated in the colorized images. It thoroughly assesses the specifically chosen range of colors in the produced images, improving our comprehension of color diversity. The metric is displayed in Equation \ref{true_activation}.

\begin{equation}
	\label{true_activation}
	TAR = \frac{1}{M} \sum_{j=1}^{M} \frac{1}{N} \sum_{i=1}^{N} I(G_{ji} = T_{ji}) 
\end{equation}
where  $G$  be the generated image,  $T$ be the ground truth image, and $N$ be the total number of pixels.
The pixel-wise matching indicator function $I$ for a pixel at index $i$ is defined as follows:
\[ I(G_{ji} = T_{ji}) = \begin{cases} 1 & \text{if } G_{ji} = T_{ji} \\ 0 & \text{otherwise} \end{cases} \]
The pixel-wise accuracy for a single image can be expressed as:
\[ \textit{Pixel Accuracy}(G, T) = \frac{1}{N} \sum_{i=1}^{N} I(G_i = T_i) \]
Now, considering this for a dataset with \( M \) images, the overall Pixel Accuracy \( PA \) can be defined as the average accuracy across all images:
where,
 \( G_{ji} \) is the intensity value of the pixel at index \( i \) in the generated image \( j \).
 \( T_{ji} \) is the intensity value of the pixel at index \( i \) in the ground truth image \( j \), and
 \( I(G_{ji} = T_{ji}) \) is the pixel-wise matching indicator function for the pixel at index \( i \) in the generated image \( j \).

\begin{algorithm}
	\small
	\caption{SAM Empowered Object-Selective Color Harmonization}
	\label{alg:segmentation}
	\begin{algorithmic}[1]
		\renewcommand{\algorithmicrequire}{\textbf{Input:}}
		\renewcommand{\algorithmicensure}{\textbf{Output:}}
		\REQUIRE Input gray image, Predicted $a^*b^*$
		\ENSURE Edge-harmonized $a^*b^*$
		
		\STATE Extract all segments ($\mathcal{S}^{\text{all}}$) from the gray image using SAM
		\STATE Select semantic segments ($\mathcal{S}^{\text{sem}}$) from the set of all segments ($\mathcal{S}^{\text{all}}$)
		
		\FOR{each segment $s$ in $\mathcal{S}^{\text{all}}$}
		\STATE Extract $\mathcal{S}^a$ from $a^*$ using the coordinates of $s$
		\STATE Calculate the mode value $\mathcal{M}_a$ of $\mathcal{S}^a$
		
		\FOR{each pixel $\mathcal{P}$ in $\mathcal{S}^a$}
		\IF{$|\mathcal{P} - \mathcal{M}_a| > \delta_a$}
		\STATE Replace the value of $\mathcal{P}$ with $\mathcal{M}_a$
		\ENDIF
		\ENDFOR
		
		\STATE Extract $\mathcal{S}^b$ from $b^*$ using the coordinates of $s$
		\STATE Calculate the mode value $\mathcal{M}_b$ of $\mathcal{S}^b$
		
		\FOR{each pixel $\mathcal{P}$ in $\mathcal{S}^b$}
		\IF{$|\mathcal{P} - \mathcal{M}_b| > \delta_b$}
		\STATE Replace the value of $\mathcal{P}$ with $\mathcal{M}_b$
		\ENDIF
		\ENDFOR
		\ENDFOR
		
		\RETURN Edge-harmonized $a^*b^*$
	\end{algorithmic}
\end{algorithm}

\section{Experiments} \label{Experiments}
This study undertakes several tests to validate the suggested method in this part. Section \ref{Datasets} describes the datasets used for training and testing. Section \ref{Implementation Set Up} explains the specifics of implementation. Section \ref{Evaluation Criteria} introduces evaluation criteria. Section \ref{Ancient colorization} presents the ancient black and white photos colorization. Ablation experiments on the network are given in Section \ref{Ablation Study}. 

\subsection{Datasets} \label{Datasets}
\subsubsection{Training}
\paragraph{Place365 Train}We train the proposed model using the \textit{Place365 Train} dataset\cite{Place}. 1.8 million images make up the Place365 training set. 365 scene categories with approximately 5000 photos each comprise the Place365 training dataset. The model we propose is constructed using a self-supervised approach. During training, we do not assign any external labels to our data. Alternatively, we create the model's supervisory signals or labels from the input data while training.  
\begin{figure}[!h]
	\centering
	\begin{subfigure}
		\centering
		\includegraphics[width=.0993\linewidth,height = 1.0cm]{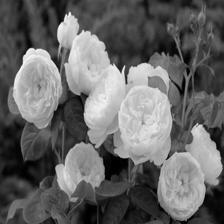}
		\includegraphics[width=.0993\linewidth,height = 1.0cm]{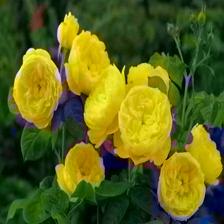}
		\includegraphics[width=.0993\linewidth,height = 1.0cm]{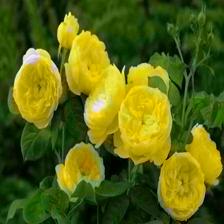}
		\includegraphics[width=.0993\linewidth,height = 1.0cm]{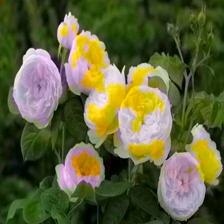}
		\includegraphics[width=.0993\linewidth,height = 1.0cm]{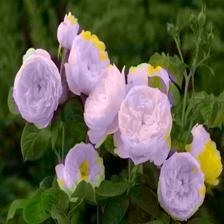}
		\includegraphics[width=.0993\linewidth,height = 1.0cm]{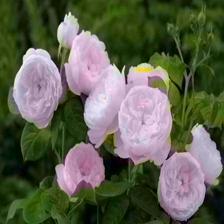}
		\includegraphics[width=.0993\linewidth,height = 1.0cm]{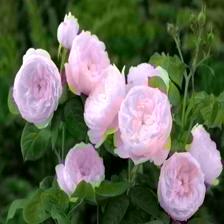}
		\includegraphics[width=.0993\linewidth,height = 1.0cm]{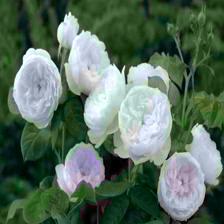}
		\includegraphics[width=.0993\linewidth,height = 1.0cm]{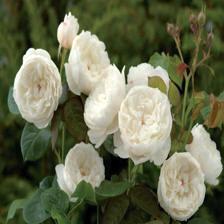}
	\end{subfigure}%
	\hfill
	\begin{subfigure}
		\centering
		\includegraphics[width=.0993\linewidth,height = 1.0cm]{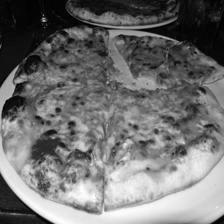}
		\includegraphics[width=.0993\linewidth,height = 1.0cm]{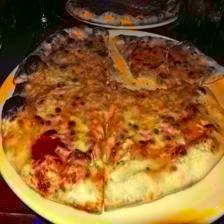}
		\includegraphics[width=.0993\linewidth,height = 1.0cm]{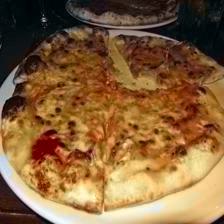}
		\includegraphics[width=.0993\linewidth,height = 1.0cm]{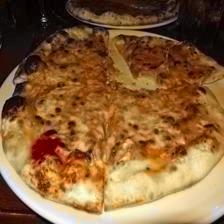}
		\includegraphics[width=.0993\linewidth,height = 1.0cm]{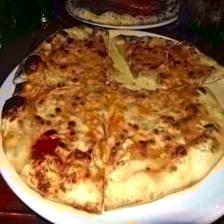}
		\includegraphics[width=.0993\linewidth,height = 1.0cm]{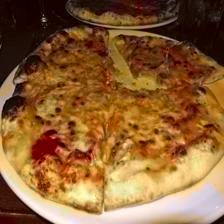}
		\includegraphics[width=.0993\linewidth,height = 1.0cm]{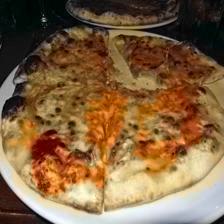}
		\includegraphics[width=.0993\linewidth,height = 1.0cm]{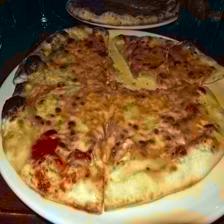}
		\includegraphics[width=.0993\linewidth,height = 1.0cm]{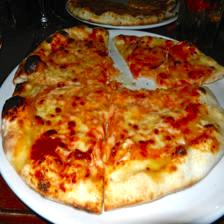}
	\end{subfigure}
	\hfill
	\begin{subfigure}
		\centering
		\includegraphics[width=.0993\linewidth,height = 1.0cm]{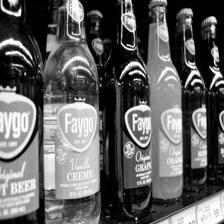}
		\includegraphics[width=.0993\linewidth,height = 1.0cm]{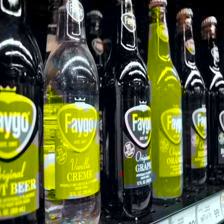}
		\includegraphics[width=.0993\linewidth,height = 1.0cm]{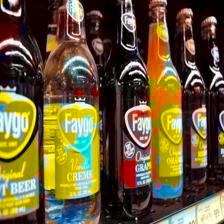}
		\includegraphics[width=.0993\linewidth,height = 1.0cm]{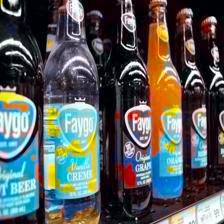}
		\includegraphics[width=.0993\linewidth,height = 1.0cm]{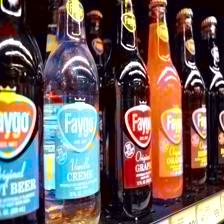}
		\includegraphics[width=.0993\linewidth,height = 1.0cm]{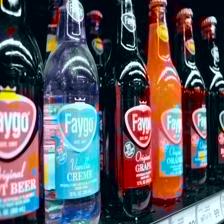}
		\includegraphics[width=.0993\linewidth,height = 1.0cm]{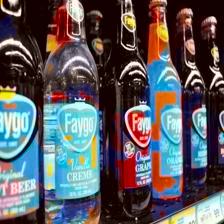}
		\includegraphics[width=.0993\linewidth,height = 1.0cm]{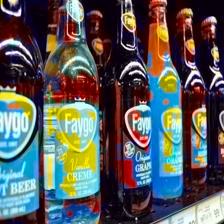}
		\includegraphics[width=.0993\linewidth,height = 1.0cm]{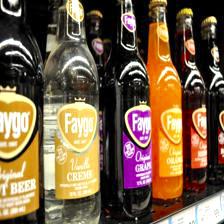}
	\end{subfigure}
	\hfill
	\begin{subfigure}
		\centering
		\includegraphics[width=.0993\linewidth,height = 1.0cm]{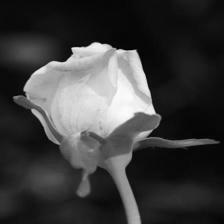}
		\includegraphics[width=.0993\linewidth,height = 1.0cm]{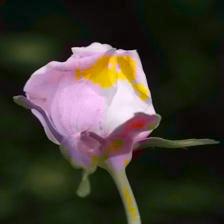}
		\includegraphics[width=.0993\linewidth,height = 1.0cm]{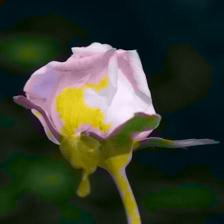}
		\includegraphics[width=.0993\linewidth,height = 1.0cm]{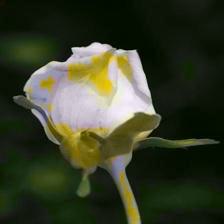}
		\includegraphics[width=.0993\linewidth,height = 1.0cm]{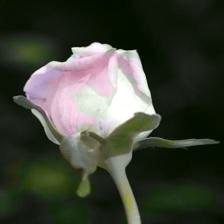}
		\includegraphics[width=.0993\linewidth,height = 1.0cm]{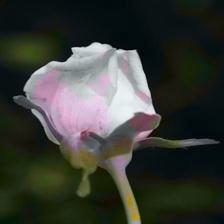}
		\includegraphics[width=.0993\linewidth,height = 1.0cm]{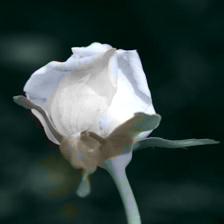}
		\includegraphics[width=.0993\linewidth,height = 1.0cm]{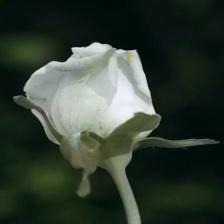}
		\includegraphics[width=.0993\linewidth,height = 1.0cm]{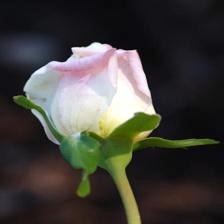}
	\end{subfigure}
	\hfill
	\begin{subfigure}
		\centering
		\includegraphics[width=.0993\linewidth,height = 1.0cm]{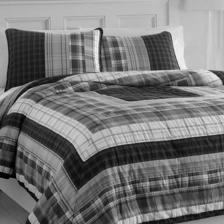}
		\includegraphics[width=.0993\linewidth,height = 1.0cm]{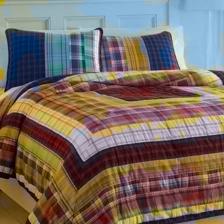}
		\includegraphics[width=.0993\linewidth,height = 1.0cm]{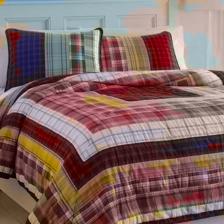}
		\includegraphics[width=.0993\linewidth,height = 1.0cm]{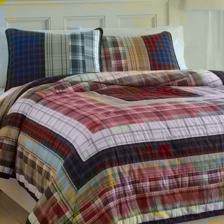}
		\includegraphics[width=.0993\linewidth,height = 1.0cm]{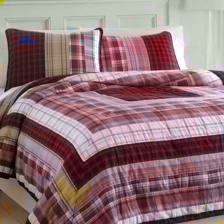}
		\includegraphics[width=.0993\linewidth,height = 1.0cm]{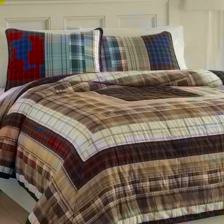}
		\includegraphics[width=.0993\linewidth,height = 1.0cm]{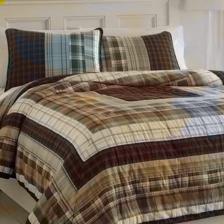}
		\includegraphics[width=.0993\linewidth,height = 1.0cm]{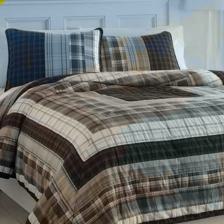}
		\includegraphics[width=.0993\linewidth,height = 1.0cm]{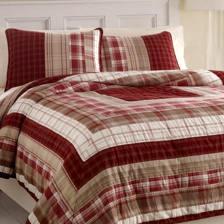}
	\end{subfigure}
	\hfill
	\begin{subfigure}
		\centering
		\includegraphics[width=.0993\linewidth,height = 1.0cm]{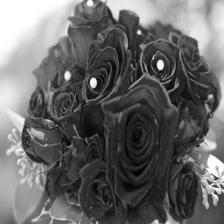}
		\includegraphics[width=.0993\linewidth,height = 1.0cm]{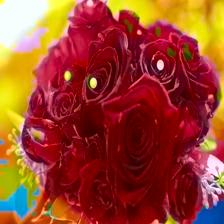}
		\includegraphics[width=.0993\linewidth,height = 1.0cm]{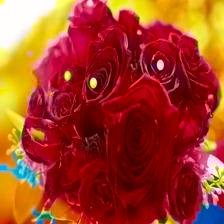}
		\includegraphics[width=.0993\linewidth,height = 1.0cm]{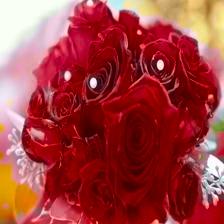}
		\includegraphics[width=.0993\linewidth,height = 1.0cm]{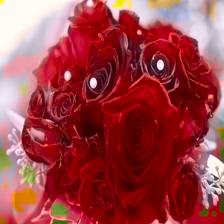}
		\includegraphics[width=.0993\linewidth,height = 1.0cm]{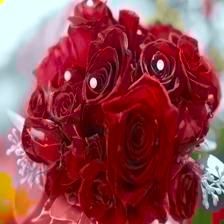}
		\includegraphics[width=.0993\linewidth,height = 1.0cm]{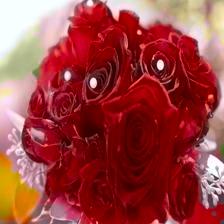}
		\includegraphics[width=.0993\linewidth,height = 1.0cm]{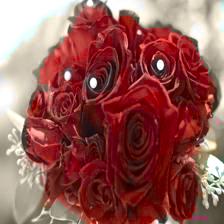}
		\includegraphics[width=.0993\linewidth,height = 1.0cm]{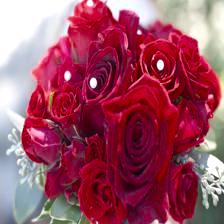}
	\end{subfigure}
	\caption*{Gray\hspace{4mm} 4\%\hspace{4mm} 6\% \hspace{3mm}8\% \hspace{3mm}10\% \hspace{3mm}12\% \hspace{3mm}14\% \hspace{2mm}16\% \hspace{1mm}G. Tru.}
	\caption{Results on hyperparameters($\mathcal{P}$) testing for determining the value of minimum appeared class for making perfect trade-off between the weights of major and minor classes. }
	\label{fig:hyperparameter}
	\vspace{-4mm}
\end{figure} 
\subsubsection{Testing}
For testing we have used 5 datasets \textit{Place365 Test}\cite{Place}, \textit{ImageNet1k Validation}\cite{Imagenet}, \textit{Oxford 102 Flower}\cite{Oxford_flower}, CelebFaces(\textit{CelebA})\cite{Celeba}, and Common Objects in Context (\textit{COCO})\cite{COCO}.
\vspace{-2mm}
\subsection{Implementation Set Up} \label{Implementation Set Up}
The studies were conducted on a workstation with an NVIDIA GEFORCE RTX 2080 Ti graphics processing unit (GPU). The PyTorch version 1.28 was utilized to implement the neural network in Python 3.10.9. During the training, we selected a batch size of 64 and employed the Adam optimizer with a learning rate of $1 \times 10^{-3}$. The momentum parameters $\beta_1 = 0.5$ and $\beta_2 = 0.999$ were used to update and calculate the network parameters. To simplify the computations of our loss function, we shrunk each ground truth tensor to a dimension of $56 \times 56$. We investigated several hyperparameters and trade-off considerations for our suggested model, establishing their values through careful experimental analysis. The precise values are delineated in TABLE \ref{hyperparameter}. 
\subsubsection{Training}
During training, the batch size is set to 64, the Adam optimizer is employed with the learning rate $1 \times 10^{-3}$, and the momentum parameters $\beta_1$ = 0.5 and $\beta_2$ = 0.999 are used to update and compute the network parameters. Each of the  epochs required for the network training takes approximately 16 hours. Each input gray image was resized to $224\times224$ pixels.  Each ground truth a*b* tensor was resized into $56 \times 56$ size for reducing the complexity of loss calculations. 
\subsubsection{Testing}
The batch size is set to 64 for testing purposes. To train the network for color prediction, the input picture is scaled to $224\times224$, then fed into the network. At the conclusion of the network, the output is deconvoluted to its original size and mixed with the grayscale input to produce the created color picture. 
\begin{table}[h]
\centering
\normalsize
\caption{Different Hyper-parameter and trade-off factor values for CCC++}
\label{hyperparameter}
\begin{tabular}{c c c c c c c c c}
    \hline
H. P.& $\alpha$ & $\beta$ & $\Delta$ & $\mathcal{P}$  &\ $\delta_a$ & $\delta_b$ & $\Psi$  \\
    \hline
Value & 6 & 108 & 36 & 10 & 8  & 8 & 30 \\
    \hline
\end{tabular}
\vspace{-3mm}
\end{table}

\subsection{Evaluation Criteria} \label{Evaluation Criteria}
\subsubsection{Evaluation Metrics}
We conduct qualitative and quantitative comparisons between our proposed model and state-of-the-art colorization methods using various metrics, including mean squared error (MSE\cite{MSE}), peak signal-to-noise ratio (PSNR\cite{Psnr_ssim}), structural similarity index measure (SSIM\cite{Psnr_ssim}), learned perceptual image patch similarity (LPIPS\cite{LPIPS}), universal image quality index (UIQI\cite{UIQI}), and Fréchet inception distance score (FID\cite{FID}). These metrics thoroughly assess our model's accuracy, perceptual quality, and overall image fidelity. 

\begin{table*}[h]
    \centering
    \caption{Quantitative comparison of our proposed method with the baseline and SOTA methods using the visuals of Fig. \ref{fig:comparison}.}
    \label{tab:visual}
    \begin{tabular}{c c c c c c c c c c c}
        \hline
         &  MSE $\downarrow$ &  PSNR $\uparrow$  & SSIM $\uparrow$  & LPIPS $\downarrow$ &  UIQI $\uparrow$  &  FID $\downarrow$ &  CNR $\uparrow$  &  CCAR $\uparrow$ &  TAR $\uparrow$  \\
        \hline
         Deoldify\cite{Deoldify} & 0.0255 & 17.15 & 0.8160 & 0.2653 & \textbf{0.8618} & 3.58 &  0.6054 & 60.69 & 5.1499 \\
        \hline
         Iizuka\cite{Iizuka} & \textbf{0.0190} & 18.09 & 0.8235 & 0.2233 & 0.8553 & 3.18 &  0.4383 & 45.63 & 4.2966 \\
        \hline
         Larsson\cite{Larsson} & 0.0274 & 16.77 & 0.8024 & 0.2815 & 0.8328 & 3.93 &  0.5219 & 5238 & 5.9590\\
        \hline
         CIC\cite{Zhang_eccv} & 0.0209 & 18.02 & 0.8179 & \textbf{0.2258} & 0.8539 & 2.41 &  0.6381 & 64.47 & 3.3255 \\
        \hline
         Zhang\cite{Zhang_tog} & 0.0250 & 17.26 & 0.8197 & 0.2513 & 0.8443 & 3.72 &  0.5146 & 51.97 & 6.3340 \\
        \hline
         Su\cite{Su} & 0.0233& 17.41 & 0.7908 & 0.2989 & 0.8405 & 2.94 &  0.6230 & 63.37 & 4.5505 \\
        \hline
         Gain\cite{Gain2} & 0.0274 & 16.94 & \textbf{0.8711} & 0.2569 & 0.8402 & 2.73 &  0.7409 & 74.53 & 4.9992 \\
        \hline
         DD\cite{DD} & 0.0212 & \textbf{18.53} & 0.8161 & 0.251 & 0.8601 & \textbf{1.66} &  0.9080 & 95.00 & 4.9077 \\
        \hline
         CCC++ & 0.0240 & 17.88 & 0.8169 & 0.2681 & 0.8517 & 2.50 & \textbf{1.0507} & \textbf{111.45} & \textbf{7.0256} \\
        \hline
    \end{tabular}
    \vspace{-2mm}
\end{table*}

\subsubsection{Qualitative Comnparison} \label{Qualitative Comnparison}
Qualitative evaluation is imperative for assessing the visual prowess of our proposed method. We present a visual ensemble featuring our model's colorized outputs juxtaposed with grayscale versions, ground truth images, and comparative works from other colorization methods. This visual presentation serves as a comprehensive means to gauge the perceptual fidelity, intricate detailing, and contextual coherence achieved by our proposed method, contributing to a nuanced understanding of its visual performance compared to existing colorization approaches.

In Fig. \ref{fig:fig1}, we have showed a visual representation of the colorized images of our proposed model along with the gray and ground truth. From the figure, we can see that our proposed model produce very plausible and vivid color compared to the ground truth. In Fig. \ref{fig:reg_cla}, we have presented a set of six images to elucidate the performance disparities between regression colorization and our proposed classification-based colorization, alongside their grayscale and ground truth counterparts. A discernible distinction emerges from the visual analysis—images generated by the regression model exhibit desaturation and a pronounced grayish tone, whereas those produced by our proposed classification colorization approach showcase vibrant saturation and a diverse color palette. This visual evidence substantiates the superior color richness and diversity achieved by our proposed classification model in comparison to the desaturated outputs of the regression model. 

\begin{table*}[h]
    \centering
    \caption{Regression loss comparison of our proposed method with the baseline and SOTA methods using multiple datasets.}
    \label{mse-psnr}
    \begin{tabular}{c c c c c c c c c c c}
        \hline
        & \multicolumn{2}{c}  {ADE} & \multicolumn{2}{c}  {COCO} & \multicolumn{2}{c}  {ImageNet} & \multicolumn{2}{c}  {CelebA} & \multicolumn{2}{c}  {Oxford Flower}  \\
         &  MSE $\downarrow$  &  PSNR $\uparrow$ &  MSE $\downarrow$ &  PSNR $\uparrow$ &  MSE $\downarrow$ &  PSNR $\uparrow$ &  MSE $\downarrow$ &  PSNR $\uparrow$ &  MSE $\downarrow$ &  PSNR $\uparrow$ \\
        \hline
        DeOldify\cite{Deoldify} & .0043 & 25.66 & 0.0067 &  23.01 & 0.0136 & 20.09 & .0045  & 26.06 & .0295 & 16.46\\
        \hline
        Iizuka\cite{Iizuka} & \textbf{.0035} & \textbf{26.22} & 0.0067 & 23.26 & 0.0115 & 21.10 & .0045 & 26.00 & .0211 & 18.01\\
        \hline
        Larsson\cite{Larsson} & .0037 & 25.94 & 0.0073 & 22.34 & 0.0168 & 19.26 & .0058 & 26.66 & .0245 & 16.85 \\
        \hline
        CIC\cite{Zhang_eccv} & .0053 & 24.33 & 0.0075 & 22.18 & 0.0109 & 20.72 & .0056 & 24.79 & .0261 & 17.16\\
        \hline
        Zhang\cite{Zhang_tog} & .0036 & 26.07 & \textbf{0.0061} & \textbf{23.32} & 0.0098 & 21.61 & \textbf{.0041} & \textbf{26.78} & .0295 & 16.80\\
        \hline
        Su\cite{Su} & .0038 & 25.37 & - & - & 0.0116 &  20.66 & .0046 & 25.70 & .0265 & 16.81\\
        \hline
        DD\cite{DD} & .0039 & 25.22 & 0.0071 & 22.73 & \textbf{0.0097} & \textbf{21.6}8 & .0066 & 25.70 & .0273 & 16.88\\
        \hline
        CCC++ & .0051 & 24.53 & 0.0095 & 20.74 & 0.0190 & 19.12 & .0058 & 24.85 & \textbf{.0181} & \textbf{19.13}\\
        \hline
    \end{tabular}
    \vspace{-2mm}
\end{table*}

In Fig. \ref{fig:com_opti}, six images have been presented, showcasing both full classes and optimized classes colorized outputs, in conjunction with their grayscale and ground truth counterparts. The visual analysis reveals a notable distinction: full class colorization struggles to accurately capture all colors, as the increased number of classes compromises prediction accuracy. Conversely, optimized classes colorization yields more realistic and vibrant color predictions. The precision of the colorization process is enhanced when employing fewer classes, elucidating the trade-off between accuracy and complexity in the optimization of color class predictions.
\begin{figure}[!h]
	\centering
	\begin{subfigure}
		\centering
		{\rotatebox{90}{\small Gray}} 
		\includegraphics[width=.18\linewidth,height = 1.5cm]{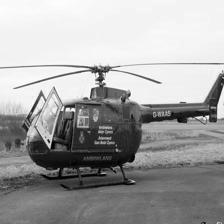}
		\includegraphics[width=.18\linewidth,height = 1.5cm]{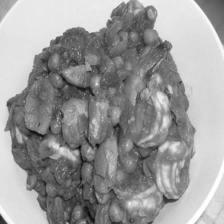}
		\includegraphics[width=.18\linewidth,height = 1.5cm]{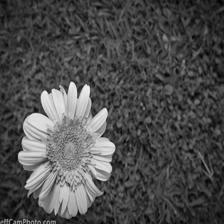}
		\includegraphics[width=.18\linewidth,height = 1.5cm]{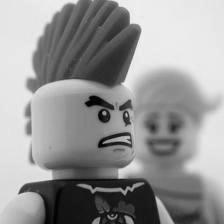}
		\includegraphics[width=.18\linewidth,height = 1.5cm]{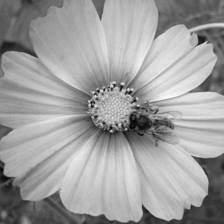}
	\end{subfigure}%
	\hfill
	\begin{subfigure}
		\centering
		{\rotatebox{90}{\small W/o Har.}} 
		\includegraphics[width=.18\linewidth,height = 1.5cm]{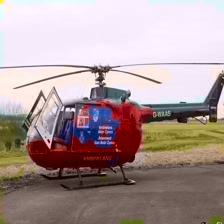}
		\includegraphics[width=.18\linewidth,height = 1.5cm]{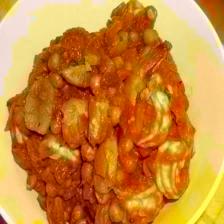}
		\includegraphics[width=.18\linewidth,height = 1.5cm]{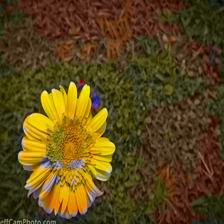}
		\includegraphics[width=.18\linewidth,height = 1.5cm]{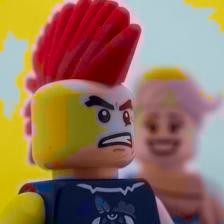}
		\includegraphics[width=.18\linewidth,height = 1.5cm]{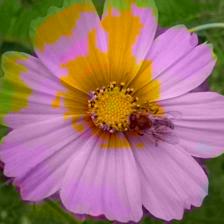}
	\end{subfigure}
	\hfill
	\begin{subfigure}
		\centering
		{\rotatebox{90}{\small W. Har.}} 
		\includegraphics[width=.18\linewidth,height = 1.5cm]{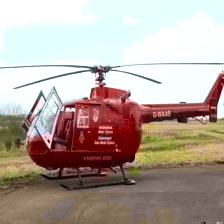}
		\includegraphics[width=.18\linewidth,height = 1.5cm]{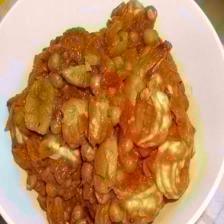}
		\includegraphics[width=.18\linewidth,height = 1.5cm]{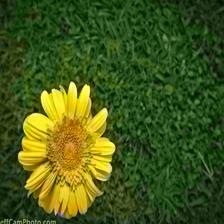}
		\includegraphics[width=.18\linewidth,height = 1.5cm]{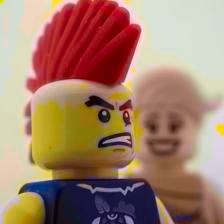}
		\includegraphics[width=.18\linewidth,height = 1.5cm]{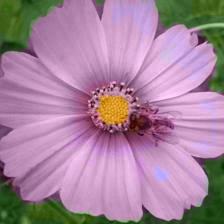}
	\end{subfigure}
	\hfill
	\begin{subfigure}
		\centering
		{\rotatebox{90}{\small G. Truth}} 
		\includegraphics[width=.18\linewidth,height = 1.5cm]{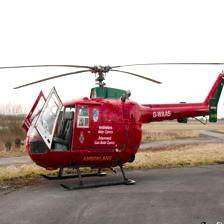}
		\includegraphics[width=.18\linewidth,height = 1.5cm]{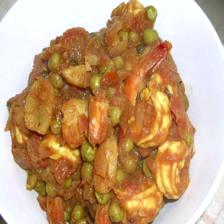}
		\includegraphics[width=.18\linewidth,height = 1.5cm]{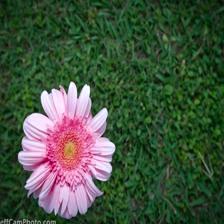}
		\includegraphics[width=.18\linewidth,height = 1.5cm]{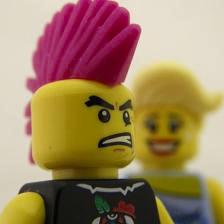}
		\includegraphics[width=.18\linewidth,height = 1.5cm]{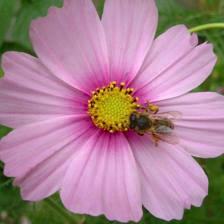}
	\end{subfigure}
	\caption{Results with and without color harmonization}
	\label{fig:harmonization}
	\vspace{-2mm}
\end{figure}
In Fig. \ref{fig:reb_no}, we present a series of six images to elucidate the outcomes of the rebalance and nobalance approaches, accompanied by their grayscale and ground truth counterparts. The visual examination reveals a substantial difference: in the nobalance approach, the accuracy of color class predictions is notably low, resulting in a desaturated visual appearance. This can be attributed to an imbalance in the number of desaturated and saturated components within an image. Conversely, the rebalance approach yields saturated and visually plausible results, demonstrating the effectiveness of addressing imbalances in color components for enhanced colorization accuracy.
\begin{table*}[h]
	\centering
	\caption{Structural Similarity comparison of our proposed method with the baseline and SOTA methods using multiple datasets.}
	\label{ssim-uiqi}
	\begin{tabular}{c c c c c c c c c c c}
		\hline
		& \multicolumn{2}{c}  {ADE} & \multicolumn{2}{c}  {COCO} & \multicolumn{2}{c}  {ImageNet}  & \multicolumn{2}{c}  {CelebA} & \multicolumn{2}{c}  {Oxford Flower}  \\
		&  SSIM $\uparrow$ &  UIQI $\uparrow$ &  SSIM $\uparrow$ &  UIQI $\uparrow$ &  SSIM $\uparrow$ &  UIQI $\uparrow$ &  SSIM $\uparrow$ &  UIQI $\uparrow$ &  SSIM $\uparrow$ &  UIQI $\uparrow$\\
		\hline
		DeOldify\cite{Deoldify} & \textbf{0.96} & \textbf{0.96} & 0.8631 & 0.9343 & 0.9040 & 0.9006 & 0.94  & \textbf{0.94} & \textbf{0.82} & 0.81\\
		\hline
		Iizuka\cite{Iizuka} & 0.95 & \textbf{0.96} & \textbf{0.9150} & \textbf{0.9429} & 0.8918 & 0.9113 & \textbf{0.95} & \textbf{0.94} & 0.80 & 0.82\\
		\hline
		Larsson\cite{Larsson} & 0.95 & \textbf{0.96} & 0.9021 & 0.9253 &  0.8893 & 0.8910 & 0.94 & 0.93 & \textbf{0.82} & \textbf{0.83}\\
		\hline
		CIC\cite{Zhang_eccv} & 0.95 & 0.95 & 0.8599 & 0.9286 & 0.9045 & 0.9053 & 0.93 & 0.92 & 0.81 & 0.80\\
		\hline
		Zhang\cite{Zhang_tog} & \textbf{0.96} & \textbf{0.96} & 0.8678 & 0.9374 & \textbf{0.9161} & \textbf{0.9184} & \textbf{0.95} & 0.93 & 0.81 & 0.81\\
		\hline
		Su\cite{Su} & 0.92 & \textbf{0.96} & - & - &  0.8577 & 0.9094 & 0.93 & 0.93 & 0.77 & 0.81\\
		\hline
		DD\cite{DD} & \textbf{0.96} & \textbf{0.96} & 0.9097 & 0.9350 & 0.9074 & 0.9160 & 0.93 & 0.92 & 0.81 & 0.80\\
		\hline
		CCC++ & 0.93 & 0.94 & 0.8860 & 0.9115 & 0.8954 & 0.8802 & 0.93 & 0.92 & 0.81 & 0.80\\
		\hline
	\end{tabular}
	\vspace{-2mm}
\end{table*}

In Fig.'s \ref{fig:com_sota1}(A) and \ref{fig:com_sota2}(B), a comprehensive comparison is presented, featuring twelve images generated by our proposed CCC++ method alongside a state-of-the-art (SOTA) classification colorization method, in conjunction with grayscale and ground truth images. The analysis from Fig. \ref{fig:com_sota1}(A) reveals that the colorization achieved by our proposed CCC++ method is characterized by more object-specific and saturated hues when compared to the SOTA classification colorization method. Notably, our method exhibits colorized images that closely resemble the ground truth, outperforming the SOTA classification colorization method. Further scrutiny in Fig. \ref{fig:com_sota2}(B) highlights the ability of our proposed method to capture a diverse range of saturated, object-specific colors, including nuanced color details in smaller objects—a distinctive strength when contrasted with the SOTA classification colorization method.

In Fig.\ref{fig:comparison}, we compare our proposed CCC++ method with some SOTA colorization method. From the figure we can see that, Deoldify\cite{Deoldify} produce overall desaturated color. Iijuka\cite{Iizuka} produce some color 3rd, 4th images but overall colorization is trends to grayish effect. Larsson\cite{Larsson} produce overall graish effects.  CIC\cite{Zhang_eccv} produces some object wise color but not well saturated. Zhang\cite{Zhang_tog} produces some color in 1st, 3rd, and 6th image but overall desaturated color. Su\cite{Su} produces some color in 3rd, 4th, 6th and 8th images but desaturated in others and fully fail to generate color in 2nd and 7th images. Gain\cite{Gain2} produce some color in 4th, 6th and 8th images but overall desaturation. DD\cite{DD} produce some good colorization but not object wise plausible. Our proposed CCC++ method produces fully object wise plausible, saturated and vibrant colorization compared to the others and near to the ground truth.

\subsubsection{Quantitative Comnparison} \label{Quantitative Comnparison}
\begin{table*}[h]
    \centering
    \caption{Perceptual Image Patch Similarity and frethed image distance comparison of our proposed method with the baseline and SOTA methods using multiple datasets.}
    \label{lpips-fid}
    \begin{tabular}{c c c c c c c c c c c}
        \hline
        & \multicolumn{2}{c}  {ADE}  & \multicolumn{2}{c}  {COCO} & \multicolumn{2}{c}  {ImageNet}  & \multicolumn{2}{c}  {CelebA} & \multicolumn{2}{c}  {Oxford Flower}  \\
         &  LPIPS $\downarrow$ &  FID $\downarrow$ &  LPIPS $\downarrow$ &  FID $\downarrow$ &  LPIPS $\downarrow$ &  FID $\downarrow$ &  LPIPS $\downarrow$ &  FID $\downarrow$  &  LPIPS $\downarrow$ &  FID $\downarrow$ \\
        \hline
        DeOldify\cite{Deoldify} & 0.15 & 0.48 & 0.2631 & 0.5285 & 0.2216 & 2.06 & \textbf{0.13}  & 0.43 & 0.35 & 3.85\\
        \hline
        Iizuka\cite{Iizuka} & 0.16 & 1.05 & 0.1939 & 2.03 & 0.2218 & 2.29 & 0.16 & 0.45 & 0.31 & 3.57\\
        \hline
        Larsson\cite{Larsson} & 0.16 & 0.62 & 0.1965 & 0.8935 & 0.2561 & 2.59 & 0.14 & 0.37 & 0.34 & 2.42 \\
        \hline
        CIC\cite{Zhang_eccv} & 0.18 & 1.31 & 0.2889 & 2.54 & 0.2406 & 2.61 & 0.17 & 0.58 & 0.35 & 4.20\\
        \hline
        Zhang\cite{Zhang_tog} & \textbf{0.14} & 1.12 & 0.2538 & 2.12 & \textbf{0.2081} & 2.52 & \textbf{0.13} & 0.49 & 0.34 & 4.72\\
        \hline
        Su\cite{Su} & 0.21 & 1.24 & - & - & 0.2835 & 2.55 & 0.18 & 0.28 & 0.41 & 4.51\\
        \hline
        DD\cite{DD} & 0.16 & \textbf{0.30} & \textbf{0.1847} & \textbf{0.2049} & 0.2132  & \textbf{0.65} & 0.16 & \textbf{0.18} & 0.32 & 1.54\\
        \hline
        CCC++ & 0.15 & 0.82 & 0.2068 & 1.97 & 0.2137 & 1.77 & \textbf{0.13} & 0.41 & \textbf{0.29} & \textbf{1.49}\\
        \hline
    \end{tabular}
    \vspace{-2mm}
\end{table*}

\begin{table*}[h]
    \centering
    \caption{CNR, CCAR and TAR comparison of our proposed method with the baseline and SOTA methods using multiple datasets.}
    \label{cnr}
    \begin{tabular}{c c c c c c c c c c c c c c c c c c c}
    	\hline
    	& \multicolumn{3}{c}  {ADE}  & \multicolumn{3}{c}  {COCO} & \multicolumn{3}{c}  {ImageNet}  & \multicolumn{3}{c}  {CelebA} & \multicolumn{3}{c}  {Oxford Flower}  \\
        \hline
         &  CNR &  CCAR &  TAR &  CNR &  CCAR &  TAR &  CNR &  CCAR &  TAR &  CNR &  CCAR &  TAR &  CNR &  CCAR &  TAR  \\
        \hline
        DeOldify\cite{Deoldify} & 0.77 & 51.74 & 5.10 & 1.43 & 66.18 & 3.36 & 0.61  & 53.52 & 4.11 & 0.62 & 30.60 & 9.89 & 0.69 & 66.01 & 1.90 \\
        \hline
        Iizuka\cite{Iizuka} & 0.78 & 47.51 & 2.75 & 1.49 & 45.36 & 2.45 & 0.49 & 39.87 & 1.01  & 0.51 & 47.51 & 2.75 & 0.58 & 51.84 & 1.83 \\
        \hline
        Larsson\cite{Larsson} & 0.77 & 51.24 & 5.35 & 0.73 & 67.30 & 2.79 & 0.63 & 45.23 & 1.11 & 0.64 & 29.38 & 7.68 & 0.73 & 56.13 & 2.02 \\
        \hline
        CIC\cite{Zhang_eccv} & 0.81 & 52.21 & 2.96 & 1.57 & 52.82 & 2.28 & 0.58 & 48.29 & 2.48 & 0.86 & 42.62 & 4.98 & 0.67 & 63.54 & 1.50 \\
        \hline
        Zhang\cite{Zhang_tog} & 0.73 & 47.75 & 4.32 & 1.05 & 45.70 & 3.93 & 0.49 & 42.28 & 4.66 & 0.66 & 32.58 & \textbf{10.22} & 0.66 & 62.87 & 2.16 \\
        \hline
        Su\cite{Su} & 0.77 & 46.31 & 4.68 & - & - & - & 0.66 & 52.93 & 3.52 & 0.80 & 40.93  & 8.66 & 0.66 & 60.11  & 1.71 \\
        \hline
        DD\cite{DD} & 1.25 & 76.83 & 4.48 & 2.23 & \textbf{93.73} & 3.32 & 0.94 & 81.17 & 3.75 & 1.07 & 53.37 & 7.94 & 0.88 & 83.67 & 1.87 \\
        \hline
        CCC++ & \textbf{1.80} & \textbf{77.16} & \textbf{6.25} & \textbf{3.02} & 92.62 & \textbf{4.28} & \textbf{1.06} & \textbf{86.67} & \textbf{5.31}  & \textbf{1.12} & \textbf{67.59} & 9.44 & \textbf{0.96} & \textbf{85.63} & \textbf{3.15} \\
        \hline
    \end{tabular}
    \vspace{-2mm}
\end{table*}

Quantitatively evaluate the images shown in Fig. \ref{fig:comparison} by referring to TABLE \ref{tab:visual}. Through both visual and quantitative investigation, we have determined that Deoldify\cite{Deoldify} achieves the highest Structural Similarity Index (SSIM) compared to other methods, such as Iizuka\cite{Iizuka} has achieved the highest Mean Squared Error (MSE) and Peak Signal-to-Noise Ratio (PSNR) among all the competitors. This information is supported by the research conducted by Larsson et al. cite {Larsson} achieves the highest UIQI score, as reported by Zhang et al. in their ECCV paper, CIC\cite{Zhang_eccv} achieves the highest LPIPS score, as reported by Zhang et al. in the same paper. DD achieves the highest FID score, as reported in the paper referenced as \cite{DD}. CCC++ achieves the highest CNR, CCAR, and TAR scores. The CCC++ exhibits a visually more convincing blend of major and minor colors compared to the others. Hence, it is apparent that MSE, PSNR, SSIM, LPIPS, and FID metrics are not entirely appropriate for guaranteeing the detection of subtle hues. The most effective strategy to enhance the depiction of both main and minor colors in the generated images is to initially prioritize the optimization of CNR, CCAR, and TAR, followed by consistent adherence to conventional standards.
In TABLE \ref{mse-psnr}, we assess the performance of our proposed model compared to seven baseline models and state-of-the-art approaches on three datasets using regression criteria. The data presented in the table demonstrates the strong performance of our method across all datasets, particularly in the 'Oxford Flower' dataset, where it outperforms other methods. Given that `ADE' mostly demonstrates natural photos and `Celeba' specifically emphasizes human faces, it is common for `Celeba' to exhibit a narrower variety of color combinations than `ADE'. On the other hand, the `Oxford Flowers' dataset stands out for its wide range of flower species, each displaying a distinct and diversified color scheme. The presence of many colors and complexity in the `Oxford Flowers' dataset creates a more intricate and demanding colorization setting, showcasing our method's effectiveness in dealing with a diverse range of colors and complexities. 
We assess our proposed model against other approaches on three datasets using similarity measurement criteria, as shown in TABLE \ref{ssim-uiqi}. The table demonstrates the strong performance of our technique across all datasets while preserving the subtle color structure. Typically, it is more straightforward to attain a strong resemblance by disregarding insignificant color characteristics and concentrating on prominent ones. Nevertheless, our proposed strategy preserves a reasonable level of resemblance while preserving small color details.

\begin{figure}[!h]
	\centering
	\begin{subfigure}
		\centering
		{\rotatebox{90}{\small Gray}} 
		\includegraphics[width=.18\linewidth,height = 1.5cm]{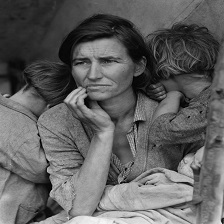}
		\includegraphics[width=.18\linewidth,height = 1.5cm]{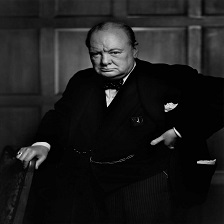}
		\includegraphics[width=.18\linewidth,height = 1.5cm]{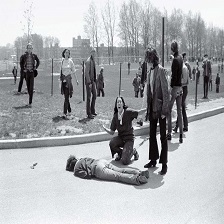}
		\includegraphics[width=.18\linewidth,height = 1.5cm]{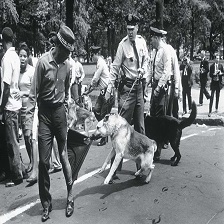}
		\includegraphics[width=.18\linewidth,height = 1.5cm]{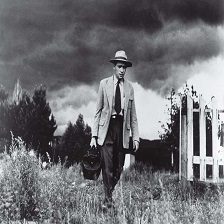}
	\end{subfigure}%
	\hfill
	\begin{subfigure}
		\centering
		{\rotatebox{90}{\small CCC++}} 
		\includegraphics[width=.18\linewidth,height = 1.5cm]{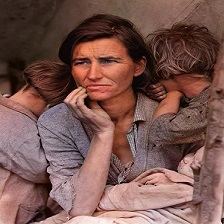}
		\includegraphics[width=.18\linewidth,height = 1.5cm]{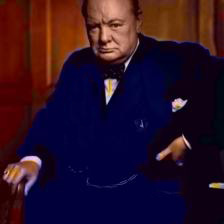}
		\includegraphics[width=.18\linewidth,height = 1.5cm]{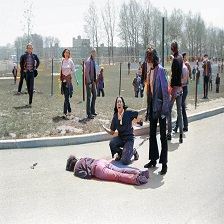}
		\includegraphics[width=.18\linewidth,height = 1.5cm]{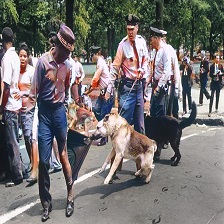}
		\includegraphics[width=.18\linewidth,height = 1.5cm]{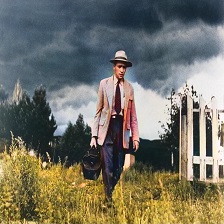}	
	\end{subfigure}
	{\hspace{5mm}(a) \hspace{10mm} (b) \hspace{10mm} (c) \hspace{12mm}(d) \hspace{12mm} (e)}

	\caption{Black and white historical images and their colorized version by our proposed CCC++ method. (a) Migrant Mother, 1936 (b) Winston Churchill, 1941 (c) Kent State Shootings, 1970 (d) Birmingham,Alabama, 1963 (e) Country Doctor, 1948.}
	\label{fig:historical}
	\vspace{-2mm}
\end{figure}

We assess our proposed model against other approaches on three datasets using the LPIPS and FID criteria, as shown in TABLE \ref{lpips-fid}. The data presented in the table demonstrates the strong performance of our method across all datasets, particularly in the 'Oxford Flower' dataset, where it outperforms other methods. The `Oxford flowers' dataset exhibits greater diversity than the `ADE' and `Celeba' datasets. 
In TABLE \ref{cnr}, we assess the effectiveness of our proposed model compared to other techniques on five datasets using the CNR, CCAR, and TAR criteria. The data presented in the table demonstrates that our approach surpasses all other methods across all datasets. The primary goal of our suggested approach is to guarantee the inclusion of both minor and major colors. Confirming minor colors enhances the diversity of color representations by including other colors alongside one or two dominant colors. 
\subsection{Ancient Black and White Photos Colorization} \label{Ancient colorization}
The presentation exhibits five historical photos with their colorized equivalents, created by our model, as seen in Fig. \ref{fig:historical}. This pair of images creates a captivating tale of visual eloquence. The colorization skill of our model beyond basic aesthetics, featuring a deep comprehension of historical contexts and subtleties. The model's capacity to accurately replicate the colors of the past is evident in the nostalgic sepia tones of historical images and the delicate blending of hues in landscapes from previous eras. This skillful use of various colors not only rejuvenates the historical visuals but also acts as a meaningful connection between different time periods, providing a modern viewpoint on filmed events from the past. The flawless incorporation of diverse shades serves as evidence of the subtle capabilities of our model, promoting a more profound and impact connection to the historical story inherent in each image.
\subsection{Ablation Study} \label{Ablation Study}
In this section, we rigorously conduct an ablation analysis to systematically dissect and assess the distinct contributions of pivotal components within our proposed model.
Illustrated in Fig. \ref{bin_size}, we offer a comprehensive comparison across various bin sizes, examining class points, trainable parameters, and average deviation. The figure reveals an intriguing trend wherein an increase in bin size corresponds to a decrease in class points and a concurrent rise in average deviation. While bin size 4 exhibits exceptionally high parameters, it also incurs elevated class points. Conversely, other bin sizes demonstrate a trade-off between parameters, class points, and average deviation. After meticulous evaluation, we strategically choose bin size 6 for our proposed model. Despite lower deviations observed in bin size 4, the associated surge in class points and parameters renders it sub-optimal. Conversely, although parameters and class points decrease in other bin sizes, a simultaneous increase in average deviation diminishes their viability. In Fig. \ref{fig:bin_size}, we present a visual representation featuring six distinct images across all bin sizes alongside their grayscale and ground truth counterparts. This visualization underscores the noteworthy correlation between lower bin sizes and heightened saturation, affirming the efficacy of our bin size selection in yielding plausible and vibrant colorization across diverse images.
In Fig. \ref{fig:hyperparameter}, we present the results for six different images across seven distinct minimum appearance value determining hyperparameter ($\mathcal{P}$) selection percentages, accompanied by their grayscale and ground truth counterparts. A discernible trend emerges, showcasing that lower percentage selections result in heightened saturation. Specifically, a minimum bin selection of 4\% produces a more aggressive and vivid color palette, while the 16\% selection yields a softer and less saturated colorization. Since low percentage make makes the minimum appearance value little, the minor start domination over minor. Likewise more percentage makes the minimum appearance of minor value large and the weights of the minor classes increase very little. Striking a balance between saturation and subtlety, we strategically choose a minimum bin selection percentage of 10\%. This choice is substantiated by the visual evidence in the figure, where the 10\% minimum bin selection consistently delivers colorization characterized by a harmonious blend of saturation, softness, and plausibility, bringing the colorized images closer to the ground truth representations.
In Fig. \ref{fig:harmonization}, we present a visual comparison featuring five images both with and without color harmonization, alongside their grayscale and ground truth counterparts. It is crucial to note a nuanced aspect here: during the introduction of minor colors into the images, occasional inconsistencies may arise. While such inconsistencies are not ubiquitous, their occasional occurrence necessitates a solution. The figure illustrates instances of color bleeding and inconsistencies in the model output, which are effectively addressed through the application of our color harmonization method. The visual transformation post-harmonization renders the images exceptionally plausible, eliminating inconsistencies and enhancing the overall aesthetic appeal. This nuanced improvement underscores the efficacy of our color harmonization technique in refining the output and ensuring a more coherent and visually pleasing result.


\section{Conclusion} \label{Conclusion}
Automatically colorizing grayscale images with objects of diverse colors and sizes presents a formidable challenge due to the intricate variations both within and between objects, compounded by the limited spatial coverage of principal elements. The learning process often grapples with favoring dominant features, thereby yielding biased models. To mitigate this, we introduce a weighted function designed to address feature imbalance, affording greater importance to minority features. In this paper, we put forth a set of formulas facilitating the conversion of color values into corresponding color classes and vice versa. We experiment on different bin size and propose 6 as optimal bin size value. To achieve ideal performance, we strategically optimize the class levels to 532 by experimenting on a large scale of diverse category images and establish a judicious trade-off between the weights assigned to major and minor classes, ensuring accurate class prediction for both. We propose a hyperparameter in the weightening formula to make a balance relation between the weights of the major and minor classes. Additionally, we introduce SAM-empowered object-selective color harmonization, a novel approach enhancing the stability of minor classes. Our contribution extends to proposing two innovative color picture assessment measures, namely Color Class Activation Ratio (CCAR) and True Activation Ratio (TAR), complementing our previously introduced Chromatic Number Ratio (CNR), to quantitatively evaluate color component richness.
Extensive evaluations on six datasets—Places, ADE, CelebA, COCO, Oxford 102 Flower, and ImageNet—against eight baseline and state-of-the-art models demonstrate that our proposed model consistently outperforms its predecessors. It excels across multiple criteria, including visualization, CNR, CCAR, and TAR while maintaining strong performance in regression metrics such as MSE and PSNR, similarity like including SSIM and LPIPS, as well as generative metrics like the UIQI and FID.




%
\vspace{-10mm}

\begin{IEEEbiography}[{\includegraphics[width=1in,height=1.1in,clip]{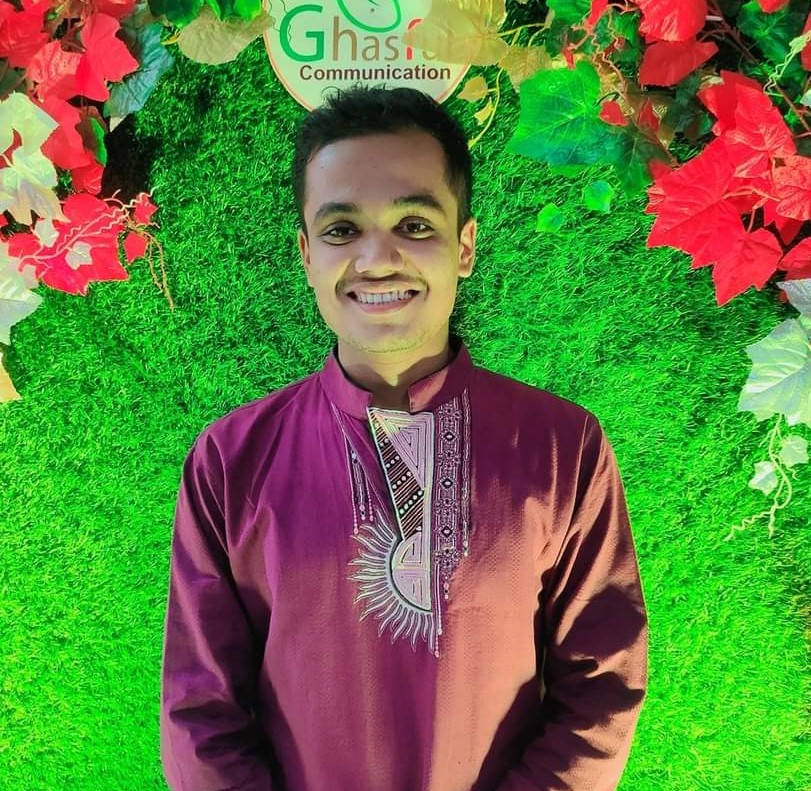}}]{Mrityunjoy Gain}
	received the B.S. degree
	in computer science from Khulna University,
	Bangladesh, in 2021, where he is currently pursuing the M.S. degree in computer science. His research interests includes continual learning, computer vision, pattern recognition etc.
\end{IEEEbiography}
\vspace{-10mm}
\begin{IEEEbiography}[{\includegraphics[width=1in,height=1.1in,clip]{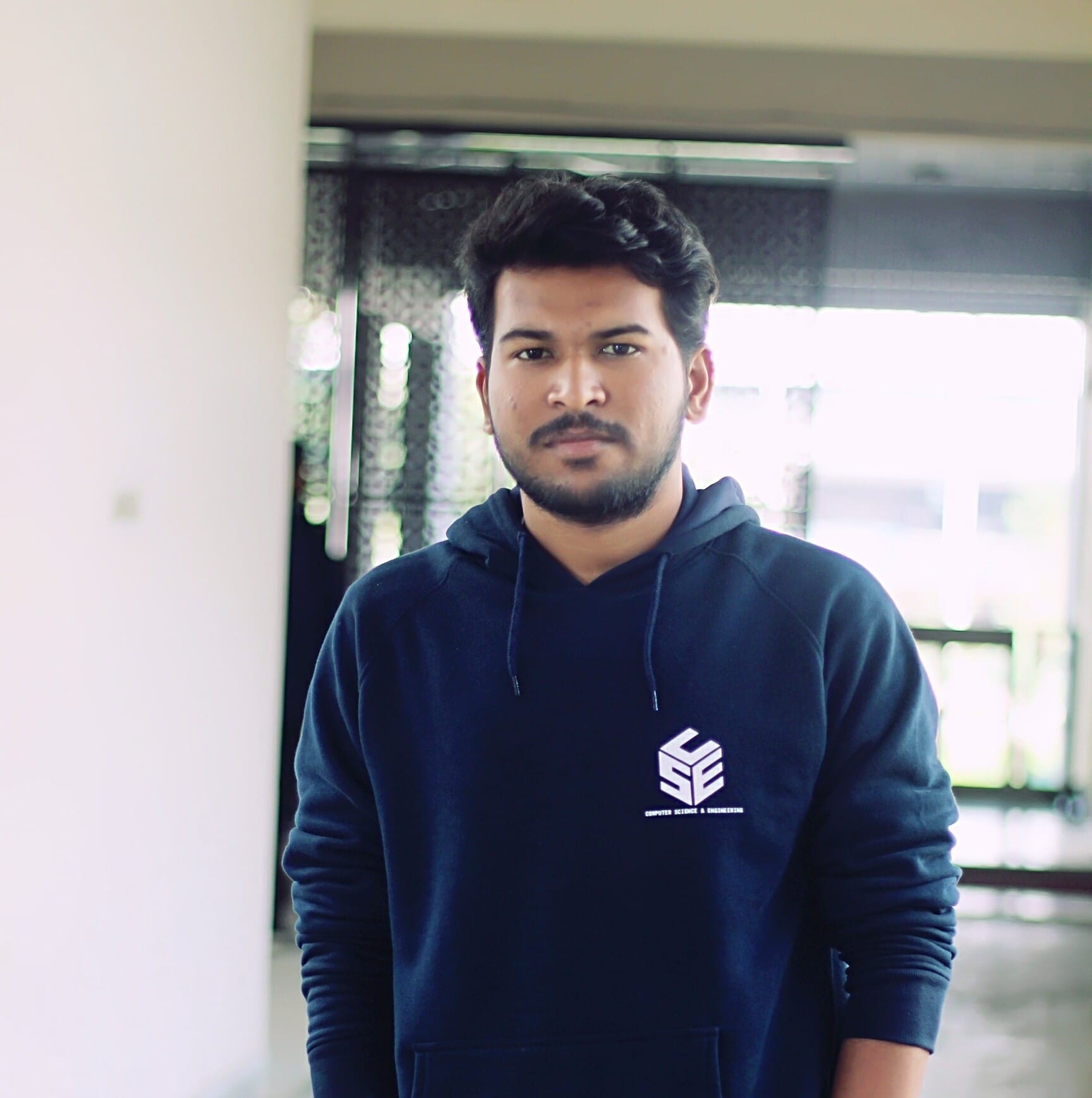}}]{Avi Deb Raha} received the B.S. degree in computer science from Khulna University, Bangladesh, in 2020. Currently he is a PhD student at the Department of Computer Science and Engineering at Kyung Hee University, South Korea. His research interests are currently focused on Deep Learning, Generative AI, Semantic Communication and Integrated Sensing and Communication. 
\end{IEEEbiography}	
\vspace{-10mm}
\begin{IEEEbiography}[{\includegraphics[width=1in,height=1.1in,clip]{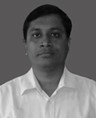}}]{Rameswar Debnath, (Member, IEEE)} received the bachelor’s degree (Hons.) in computer
	science and engineering from Khulna University,
	Bangladesh, in 1997, and the M.E. degree in
	communication and systems and the Ph.D. degree
	from The University of Electro-Communications,
	Tokyo, in 2002 and 2005, respectively, under the
	Japanese Government Scholarship.
	From 2008 to 2010, he was a Postdoctoral
	Researcher under the JSPS Fellowship with the
	Department of Informatics, The University of Electro-Communications; the
	Neuroscience Research Institute; and the National Institute of Advanced
	Industrial Science and Technology, Tsukuba. From 2012 to 2015, he was
	the Head of the Computer Science and Engineering Discipline, Khulna
	University, where he is currently a Professor. His research
	interests include image data analysis, deep learning, bioinformatics, support
	vector machine, artificial neural networks, statistical pattern recognition,
	and medical image processing. He was the Organizing Chair of the
	16th International Conference on Computer and Information Technology
	(ICCIT), in 2014
\end{IEEEbiography}

%
\vfill


\begin{thebibliography}{1}
\bibliographystyle{IEEEtran}

\bibitem{Huang}
Y. C. Huang et al., ``An adaptive edge detection based colorization algorithm and its applications", in \textit{Proc. ACM international conference on Multimedia}, pp. 351-354, 2005.

\bibitem{Levin}
A. Levin et al., “Colorization using optimization,” in \textit{ACM SIGGRAPH 2004 Papers}, ACM, pp.689–694, 2004.

\bibitem{Yatziv}
L. Yatziv and G. Sapiro, “Fast image and video colorization using chrominance blending,” \textit{IEEE Transactions on Image Processing}, vol. 15, no. 5, pp. 1120–1129, 2006.

\bibitem{Qu}
Y. Qu, T.T Wong, and P.A. Heng, ``Manga colorization", \textit{ACM Transactions on Graphics (ToG)}, vol. 25, no. 3, pp.1214-1220, 2006.

\bibitem{Luan}
Q. Luan, F. Wen, D. Cohen-Or, L. Liang, Y.-Q. Xu, and H.-Y. Shum, ``Natural image colorization," In \textit{Proc. 18th Eurographics conference on Rendering Techniques}, pp. 309-320. 2007.

\bibitem{Welsh}
T. Welsh, M. Ashikhmin and K. Mueller, “Transferring color to greyscale images,” in Proc. \textit{29th Annual Conference on Computer Graphics and Interactive Techniques}, Texas, San Antonio, USA, pp. 277–280, 2002.

\bibitem{Ironi}
R. Ironi, D. Cohen-Or, D. Lischinski, ``Colorization by Example." \textit{Rendering techniques 29}, pp. 201-210, 2005.

\bibitem{Tai}
Y.-W. Tai, J. Jia and C.-K. Tang, ``Local color transfer via probabilistic segmentation by expectation-maximization," in \textit{CVPR}, pp. 747-754 vol. 1, 2005.

\bibitem{Chia}
A. Y.-S. Chia et al., ``Semantic colorization with internet images." \textit{ACM Transactions on Graphics (ToG)} vol. 30, no. 6 pp. 1-8, 2011.
  
\bibitem{Liu}
X. Liu et al., ``Intrinsic colorization," In \textit{ACM SIGGRAPH Asia 2008 papers}, pp. 1-9. 2008.

\bibitem{Sousa}
U. Sousa, R. Kabirzadeh and P. Blaes, ``Automatic colorization of grayscale images,” Department of Electrical Engineering, Stanford University, 2013.

\bibitem{He}
M. He et al., ``Deep exemplar-based colorization." \textit{ACM Transactions on Graphics (TOG)} vol. 37, no. 4, pp. 1-16, 2018.

\bibitem{Zhang_tog}
R. Zhang et al, “Real-time user-guided image colorization with learned deep priors,” \textit{ACM Transactions on Graphics (TOG)}, vol. 36, no. 4, pp. 1- 11, 2017.

\bibitem{Charpiat}
G. Charpiat et al., ``Automatic image colorization via multimodal predictions." In \textit{ECCV}, pp. 126-139, 2008.

\bibitem{Gupta}
R. K. Gupta et al., ``Image colorization using similar images." In \textit{Proc. of the 20th ACM international conference on Multimedia}, pp. 369-378. 2012.

\bibitem{Bugeau}
A. Bugeau et al., ``Variational Exemplar-Based Image Colorization," in \textit{IEEE Transactions on Image Processing}, vol. 23, no. 1, pp. 298-307, Jan. 2014.

\bibitem{Wu1}
D. Wu et al., ``Fine‐grained semantic ethnic costume high‐resolution image colorization with conditional GAN," \textit{International Journal of Intelligent Systems}, vol. 37, no. 5, pp. 2952-2968. 2022.

\bibitem{Wu2}
Y. Wu et al., ``Towards vivid and diverse image colorization with generative color prior,” in \textit{ICCV}, pp. 14377-14386, 2021.

\bibitem{Wu3}
M. Wu et al., ``Remote sensing image colorization using symmetrical multi-scale DCGAN in YUV color space,” \textit{The Visual Computer}, vol. 37, no. 7, pp. 1707-1729, 2021.

\bibitem{Guo}
H. Guo et al., ``Bilateral res-unet for image colorization with limited data via GANs," in \textit{ICTAI}, pp. 729-735, 2021.

\bibitem{Bahng}
H. Bahng et al., ``Coloring with words: guiding image colorization through text-based palette generation,” in \textit{ECCV} pp. 431-447, 2018.

\bibitem{Liang}
Y. Liang et al., ``Unpaired medical image colorization using generative adversarial network,” \textit{Multimedia Tools and Applications}, vol. 81, no. 19, pp. 26669-26683, 2022.

\bibitem{Zang}
S. Zang et al.,``Texture-aware gray-scale image colorization using a bistream generative adversarial network with multi scale attention structure," \textit{Engineering Applications of Artificial Intelligence}, vol. 122, pp. 106094, 2023.

\bibitem{Larsson}
G. Larsson et al., ``Learning representations for automatic colorization,” in \textit{ECCV}, pp. 577-593, 2016.

\bibitem{Gain1}
M. Gain et al., ``An Improved Encoder-Decoder CNN with Region-Based Filtering for Vibrant Colorization,”\textit{ Computer Systems Science and Engineering}, vol. 46, no. 1, pp. 1059-1077, 2023.

\bibitem{Gain2}
M. Gain and R. Debnath, ``A Novel Unbiased Deep Learning Approach (DL-Net) in Feature Space for Converting Gray to Color Image," in IEEE Access, vol. 11, pp. 78918-78933, 2023.

\bibitem{An}
J. An et al., ``Image colorization with convolutional neural networks," in \textit{CISP-BMEI}, pp. 1-4, 2019.


\bibitem{Iizuka}
S. Iizuka et al., ``Let there be color! joint end-to-end learning of global and local image priors for automatic image colorization with simultaneous classification,” \textit{ACM Transactions on Graphics(ToG)}, vol. 35, no. 4, pp. 1–11, 2016.

\bibitem{Su}
J. W. Su et al., ``Instance-aware image colorization,” in \textit{CVPR}, pp. 7968-7977, 2020

\bibitem{Dai}
J. Dai et al., ``Local pyramid attention and spatial semantic modulation for automatic image colorization,” in \textit{CCF Conference on Big Data}, pp. 165-181, 2022.

\bibitem{Dahl}
R. Dahl, ``Automatic colorization,” 2016. [Online]. Available: https://tinyclouds.org/colorize.

\bibitem{Baldassarre}
F. Baldassarre et al., ``Deep koalarization: image colorization using cnns and inception-resnet-v2,” \textit{arXiv preprint arXiv:1712.03400}, 2017.

\bibitem{Zhang_eccv}
R. Zhang et al., ``Colorful image colorization,” in \textit{ECCV}, pp. 649–666, 2016.

\bibitem{Xia}
L. Xia et al., ``A Weakly Supervised Method With Colorization for Nuclei Segmentation Using Point Annotations," in \textit{IEEE Transactions on Instrumentation and Measurement}, vol. 72, pp. 1-11, 2023.

\bibitem{Hesham}
M. Hesham et al., "Image colorization using Scaled-YOLOv4 detector," \textit{International Journal of Intelligent Computing and Information Sciences}, vol. 21, no. 3, pp. 107-118, 2021.

\bibitem{Ozbulak}
G. Ozbulak, ``Image colorization by capsule networks,” in \textit{CVPR Workshops}, pp. 0-0, 2019.

\bibitem{Kong}
G. Kong et al., ``Adversarial edge-aware image colorization with semantic segmentation,” \textit{IEEE Access}, vol. 9, pp. 28194-28203, 2021.

\bibitem{Treneska}
S, Treneska et al., ``GAN-based image colorization for self-supervised visual feature learning,” \textit{Sensors}, vol. 22, no. 4, pp. 1599, 2022.

\bibitem{Xu}
M. Xu and Y. Ding, "Fully automatic image colorization based on semantic segmentation technology," \textit{PloS one}, vol. 16, no. 11, pp. 1- 25, 2021.

\bibitem{icolorit}
J. Yun et al., ``iColoriT: Towards Propagating Local Hints to the Right Region in Interactive Colorization by Leveraging Vision Transformer," in \textit{CVPR}, pp. 1787-1796, 2023.

\bibitem{DD}
X. Kang et al., ``DDColor: Towards Photo-Realistic Image Colorization via Dual Decoders," in \textit{ICCV}, pp. 328-338, 2023.

\bibitem{codebook}
H. Tang et al., "Two-stage image colorization via color codebook," \textit{Expert Systems with Applications}, vol. 250, p. 123943, 2024.

\bibitem{Lee}
G. Lee et al., ``Real-Time User-Guided Adaptive Colorization With Vision Transformer," in \textit{WACV}, pp. 484-493, 2024.

\bibitem{lcad}
S. Weng et al., ``L-CAD: Language-based  Colorization with Any-level Descriptions using Diffusion Priors,'' in  \emph{NeurIPS}, 2024.

\bibitem{diffusart}
H. Carrillo et al., ``Diffusart: Enhancing line art colorization with conditional diffusion models,'' in \emph{CVPR}, pp. 3485-3489, 2023.

\bibitem{colorcontrol}
Z. Liang et al., ``Control Color: Multimodal Diffusion-based Interactive Image Colorization,'' \emph{arXiv preprint arXiv:2402.10855}, 2024.

\bibitem{Gain_facemask}
M. Gain et al., ``Transfer Learning Based Face Mask Detection Using Deep Neural Networks", in \textit{Korean Computer Congress(KCC)},2023.

\bibitem{Raha_access}
A. D. Raha et al., ``Attention to Monkeypox: An Interpretable Monkeypox Detection Technique Using Attention Mechanism," in \textit{IEEE Access}, vol. 12, pp. 51942-51965, 2024.

\bibitem{Raha_advancing}
A. D. Raha et al., ``Advancing Ultra-Reliable 6G: Transformer and Semantic Localization Empowered Robust Beamforming in Millimeter-Wave Communications," arXiv preprint arXiv:2406.02000 (2024).

\bibitem{Gain_iceeict}
M. Gain et al., ``LEO Satellite Oriented Wildfire Detection Model Using Deep Neural Networks: A Transfer Learning Based Approach," in \textit{ICEEICT} pp. 214-219, 2024.

\bibitem{DWBL_12}
K. Ruwani et al., ``Dynamically Weighted Balanced Loss: Class Imbalanced Learning and Confidence Calibration of Deep Neural Networks,” In \textit{IEEE Trans. Neural Networks and Learning Systems}, vol. 33, no. 7, pp. 2940-2951, 2021.

\bibitem{Hendrycks_13}
D. Hendrycks and K. Gimpel, ``A baseline for detecting misclassified and out-of-distribution examples in neural networks,” \textit{ICLR}, 2017.

\bibitem{Wallace_14}
B. C. Wallace and I. J. Dahabreh, ``Improving class probability estimates for imbalanced data,” \textit{Knowledge and information systems}, vol. 41, no. 1, pp. 33–52, 2014.

\bibitem{ccc}
M. Gain, A. D. Raha, R. Debnath, ``CCC: Color Classified Colorization," arXiv preprint arXiv:2403.01476 (2023).

\bibitem{Place}
B. Zhou et al., ``Places: a 10 million image database for scene recognition," in \textit{TPAMI}, vol. 40, no. 6, pp. 1452-1464, 2018.

\bibitem{Reinhard}
E. Reinhard et al., ``Color transfer between images," \textit{IEEE Computer graphics and applications}, vol. 21, no. 5 pp. 34–41, 2001.

\bibitem{VGG}
K. Simonyan and A. Zisserman, ``Very deep convolutional networks for large-scale image recognition," \textit{arXiv preprint arXiv:1409.1556}, 2014.

\bibitem{Hwang}
J. Hwang and Y. Zhou, “Image colorization with deep convolutional neural networks,” in Stanford University, Tech. Rep., vol. 219, Stanford, CA 94305, United States, pp. 1-7, 2016.

\bibitem{Nguyen-Quynh}
T. -T. Nguyen-Quynh et al., “Image colorization using the global scene-context style and pixel-wise semantic segmentation,”\textit{ IEEE Access}, vol. 8, pp. 214098-214114, 2020.


\bibitem{Deoldify}
J. Antic, ``A deep learning based project for colorizing and restoring old images and video!,” 2018. [Online]. Available: https://github.com/jantic/DeOldify.


\bibitem{CIE}
A. R. Robertson, ``The CIE 1976 color-difference formulae”, \textit{Color Research \& Application}, vol. 2, no. 1, pp.7–11, 1977.

\bibitem{DenseNet}
G. Huang et al., ``Densely Connected Convolutional Networks," in \textit{CVPR}, 2017, pp. 2261-2269, doi: 10.1109/CVPR.2017.243.

\bibitem{SAM}
A. Kirillov et al., ``Segment Anything". In \emph{ICCV}, pages 4015-4026, October 2023.

\bibitem{raha_arxiv}
A. D. Raha et al.. ``Generative AI-driven Semantic Communication Framework for NextG Wireless Network." arXiv preprint arXiv:2310.09021 (2023).

\bibitem{Imagenet}
J. Deng et al., ``ImageNet: A Large-Scale Hierarchical Image Database",  \textit{IEEE Computer Vision and Pattern Recognition (CVPR)}, 2009.

\bibitem{Oxford_flower}
M. Nilsback et al., ``Automated Flower Classification over a Large Number of Classes", In \textit{Proc. Indian Conference on Computer Vision, Graphics and Image Processing}, 2008.

\bibitem{Celeba}
Z. Liu et al., ``Large-scale CelebFaces Attributes (CelebA) Dataset", \textit{In ICCV}, 2015.

\bibitem{COCO}
T.-Y. Lin et al., ``Microsoft coco: Common objects in context", In \textit{ECCV}, pp. 740-755, 2014. 

\bibitem{Pytorch}
A. Paszke et al., ``Pytorch: an imperative style, high-performance deep learning library,” \textit{NeurIPS}, vol. 32, 2019.

\bibitem{MSE}
Z. Wang and A.C. Bovik, ``Mean squared error: Love it or leave it? A new look at signal fidelity measures,” \textit{IEEE Signal Processing Magazine}, vol. 26, no. 1, pp.98-117, 2009.

\bibitem{Psnr_ssim}
A. Hore´ and D. Ziou, ``Image quality metrics: PSNR vs. SSIM,” in \textit{ICPR}, pp. 2366–2369, 2010.

\bibitem{LPIPS}
R. Zhang et al., ``The unreasonable effectiveness of deep features as a perceptual metric", in \textit{CVPR}, 2018.

\bibitem{UIQI}
Z. Wang and A. C. Bovik, "A universal image quality index," in \textit{IEEE Signal Processing Letters}, vol. 9, no. 3, pp. 81-84, 2002.

\bibitem{FID}
M. Heusel et al., ``GANs trained by a two time-scale update rule converge to a local nash equilibrium", in Proc. \textit{NeurIPS}, vol. 30, 2017.

\end{thebibliography}
\end{document}